\documentclass[12pt]{article}
\usepackage{amsfonts,amssymb}
\usepackage{hyperref}
\usepackage{titlesec}
\usepackage{titletoc}
\usepackage{amsmath}
\numberwithin{equation}{section}
\usepackage{listings}
\usepackage{verbatim}
\usepackage{graphicx}
\usepackage{geometry}
\usepackage{mathrsfs}
\usepackage{amsthm}
\usepackage{bbm}
\usepackage{bm}
\usepackage{subcaption}
\newtheorem{assumption}{Assumption}
\newtheorem{definition}{Definition}
\usepackage{xcolor, tikz}
\usetikzlibrary{positioning}
\usetikzlibrary{matrix}
\newtheorem{theorem}{Theorem}
\newtheorem{pro}{Proposition}
\newtheorem{lem}{Lemma}
\newtheorem{cor}{Corollary}
\newtheorem{rem}{Remark}
\theoremstyle{definition}
\newtheorem{exmp}{Example} 

\DeclareMathOperator*{\sign}{\mathrm{sign}}
\DeclareMathOperator*{\uni}{\mathrm{uni}}

\usepackage[autostyle=false, style=english]{csquotes}
\MakeOuterQuote{"}

\usepackage{cite}
\usepackage{bigfoot}

\DeclareNewFootnote{AAffil}[arabic]
\DeclareNewFootnote{ANote}[fnsymbol]

\usepackage{etoolbox}
\makeatletter
\patchcmd\maketitle{\def\@makefnmark{\rlap{\@textsuperscript{\normalfont\@thefnmark}}}}{}{}{}
\makeatother
\makeatletter
\def\thanksAAffil#1{
  \footnotemarkAAffil\protected@xdef\@thanks{\@thanks%
        \protect\footnotetextAAffil[\the \c@footnoteAAffil]{#1}}%
}
\def\thanksANote#1{%
  \footnotemarkANote%
  \protected@xdef\@thanks{\@thanks%
        \protect\footnotetextANote[\the \c@footnoteANote]{#1}}%
}
\makeatother
\title{High Dimensional Statistical Estimation under Uniformly Dithered One-bit Quantization\thanks{Junren Chen and Michael K. Ng were supported in part by Hong Kong Research Grant Council GRF 12300218, 12300519, 17201020,
17300021, C1013-21GF, C7004-21GF and Joint NSFC-RGC N-HKU76921. Di Wang and Cheng-Long Wang were supported in part by the baseline funding BAS/1/1689-01-01, funding from the CRG grand URF/1/4663-01-01, FCC/1/1976-49-01 from CBRC and funding from the AI Initiative REI/1/4811-10-01 of King Abdullah University of Science and Technology (KAUST).}}

\author{
Junren Chen%
\thanksAAffil{Department of 
		Mathematics, The University of Hong Kong.}
		$^{,}$\thanksANote{Corresponding authors:  chenjr58@connect.hku.hk, mng@maths.hku.hk}
		,
		Cheng-Long Wang%
		\thanksAAffil{Division of CEMSE, King Abdullah University of Science and Technology (KAUST).}
		,
		Michael K. Ng\footnotemarkAAffil[1]
		$^{,}$\footnotemarkANote[1]
		,
		Di Wang\footnotemarkAAffil[2]
		$^{,}$\thanksAAffil{Computational Bioscience Research Center (CBRC), King Abdullah University of Science and Technology (KAUST).}
		}
\date{\today}
\begin{document}

\maketitle
\begin{abstract}
In this paper, we propose a uniformly dithered 1-bit quantization scheme for high-dimensional statistical estimation. The scheme contains truncation, dithering, and quantization as typical steps. As canonical examples,  the quantization scheme is applied to the estimation problems of sparse covariance matrix estimation, sparse linear regression (i.e., compressed sensing), and matrix completion. We study both sub-Gaussian and heavy-tailed regimes, where the underlying distribution of heavy-tailed data  is assumed to have bounded moments of some order. We propose new estimators based on   1-bit quantized data. In sub-Gaussian regime, our estimators achieve minimax rates  up to   logarithmic factors,  indicating that our quantization scheme  costs very little. In heavy-tailed regime, while the rates of our estimators become essentially slower, these  results are either  the first  ones in an 1-bit quantized and heavy-tailed setting, or already improve on existing comparable results from some respect. Under the observations in our setting, the rates are almost tight in compressed sensing and matrix completion. Our 1-bit compressed sensing results feature general sensing vector that is sub-Gaussian or even heavy-tailed. We also first investigate a novel setting where both the covariate and response are quantized. In addition, our approach to 1-bit matrix completion  does not rely on likelihood and  represent the first method robust to pre-quantization noise with unknown distribution. Experimental results on synthetic data are presented to support our theoretical analysis.
\vspace{15mm}
	\end{abstract}

\section{Introduction}
\label{section1}
1-bit quantization of signals or data recently has received much attention in both signal processing and machine learning communities. In some signal processing problems, power consumption, manufacturing cost and chip area of analog-to-digital devices grow exponentially with their resolution \cite{kipnis2018fundamental}. Thus, it is impractical and infeasible to use high-precision data or signals. Alternatively, it was proposed to use low-resolution, specifically 1-bit quantization, see for instance 
\cite{mo2018limited,de2018reconsidering,choi2016near,bar2002doa,roth2015covariance,khobahi2019deep}. Note that, generally speaking, the quantization itself that maps an analog signal into digital representation of a finite dictionary is an inevitable process in digital signal processing \cite{gray1998quantization,gray1993dithered}. Besides, in many distributed machine learning or federated learning scenarios, multiple parties transmit information among themselves. The communication cost can be prohibitive for distributed algorithms where each party only has a low-power and low-bandwidth device such as a mobile device \cite{konevcny2016federated}. To address the bottleneck of communication cost, recent works have studied how to send a small number or even one bit per entry for such distributed machine learning 
applications \cite{bai2021gradient,seide20141,bernstein2018signsgd,vargaftik2021drive}. 

Because of the pratical interest of 1-bit quantization in many applications, recent years have witnessed increasing literature on high-dimensional statistical estimation from merely binary (1-bit) data, which we sometimes refer to as 1-bit estimation. Existing works tried to understand the interplay between recovery procedures and 1-bit quantization in some prototypical estimation problems, including compressed sensing\footnote{This is also referred to as sparse linear regression in statistics. In this work we will adopt more statistical conventions --- we term sensing vector  and (compressive) measurement  as covariate and response, respectively.} (e.g., \cite{thrampoulidis2020generalized,dirksen2019quantized,dirksen2021non,plan2012robust,plan2016generalized,jacques2013robust})  and matrix completion (e.g., \cite{davenport20141,cai2013max,bhaskar2016probabilistic,klopp2015adaptive,lafond2014probabilistic}). The main goal in this area is to design quantization scheme for observed data, and of course, one hopes that accurate estimate can be obtained from the quantized data produced by the scheme. Very recently, covariance matrix estimation was also studied under 1-bit quantization by Dirksen et al. \cite{dirksen2022covariance}; they proposed to collect 2 bits per entry for each sample by using a dithered 1-bit quantizer, and also, they developed a covariance matrix estimator that enjoys near optimal operator norm error.

We believe, however, these theoretical results are still highly insufficient and the current understanding on 1-bit estimation remains incomplete. For instance, almost all existing results heavily rely on sub-Gaussianity of the underlying distribution. While many modern datasets exhibit heavy-tailed behaviour, the 1-bit quantization of heavy-tailed data is yet to investigate. Besides, a limitation of prior results for 1-bit matrix completion is that they cannot tolerate unknown pre-quantization noise, as they require construction of likelihood. We also point out that the recent results for 1-bit covariance estimation in \cite{dirksen2021non} are restricted to the low-dimensional regime. 


The main goal of this paper is to promote the   understanding on 1-bit estimation. Specifically, we study three fundamental high-dimensional statistical estimation problems based on data that are   quantized to one bit. The quantization scheme include the typical steps of truncation, dithering, and quantization (note that truncation is for heavy-tailed data only), see Section 1.2 for detailed discussions.
We present extensive theoretical results on sparse covariance matrix estimation, sparse linear regression, and low-rank matrix completion, under both sub-Gaussian data and heavy-tailed data. Here, the underlying distribution of heavy-tailed data is only assumed to have bounded moments of some order, as opposed to the conventional sub-Gaussian assumption. Our estimators in sub-Gaussian regime have remarkable statistical properties, i.e., they   achieve near  minimax rates (up to some logarithmic factors). 
In the heavy-tailed regime, our estimators can still  deliver a faithful estimation under a high-dimensional scaling; while  the error rates are essentially slower than the minimax ones because of a bias-and-variance trade-off. 
   However, to our best knowledge, these are the first high-dimensional statistical results under such two-fold predicament, i.e., heavy-tailed distribution that breaks the robustness, and 1-bit quantization that loses data information. For compressed sensing and matrix completion, we prove that the rates are nearly tight if the data are quantized under the proposed scheme and selected parameters.
   Here we summarize our key results and contributions as follows (For simplicity we only consider parameters $n$, $d$, $s$ (or $r$), $q$ and omit the others).

\begin{itemize}
    \item In Section \ref{section2}, 
  for some zero-mean $d$-dimensional random vector $X$, we study the problem of estimating its covariance matrix $\bm{\Sigma^*}= \mathbbm{E}\big(XX^T\big) = [\sigma^*_{ij}]$, where  $\bm{\Sigma^*}$ has the approximate column-wise sparsity structure, i.e., $\sup_{j\in [d]}\sum_{i=1}^d |\sigma^*_{ij}|^q \leq s$ for some $0\leq q<1$ and $s>0$. Denote the full data that are i.i.d. copies of $X$ by $X_1,...,X_n$. For sub-Gaussian $X$, we i.i.d. sample the dithering noise vector $\{\Gamma_{k1},\Gamma_{k2}:k\in [n]\}$  that are uniformly distributed on $[-\gamma, \gamma]^d$, and then dither and quantize each $X_k$ to binary data $\sign(X_k + \Gamma_{k1}),\ \sign(X_k + \Gamma_{k2})$. 
  Based on these binary data, we propose a thresholding estimator $\bm{\widehat{\Sigma}}$, see (\ref{2.4}) and (\ref{2.19}). Although only two bits are collected per entry, we show a near optimal minimax rate 
    $$
    \|\bm{\widehat{\Sigma}- \Sigma^*}\|_{\mathrm{op}} \lesssim s\log n \Big(\frac{\log d}{n}\Big)^{(1-q)/2}.
    $$
    For heavy-tailed $X$   assumed to have bounded fourth moment, we first element-wisely truncate  the full sample $X_k$ to be   $\widetilde{X}_k:= \sign(X_k)\min\{|X_k|,\eta\}$ (element-wise operation). Then similar to sub-Gaussian data, we deal with $\widetilde{X}_k$ by dithering and quantization. Our estimator possesses an estimation error bound for operator norm error 
    $$
    \|\bm{\widehat{\Sigma}- \Sigma^*}\|_{\mathrm{op}} \lesssim s \Big(\frac{\log d}{n}\Big)^{(1-q)/4}.
    $$
    
    \item In Section \ref{section3}, we study   
    sparse linear regression   $Y_k= X_k^T\Theta^* +\epsilon_k,~k\in[n]$ where the desired signal $\Theta^*=[\theta^*_i]\in \mathbb{R}^d$ satisfies $\sum_{i=1}^d |\theta^*_i|^q \leq s$ for some $0\leq q<1$ and $s>0$, the covariate $X_k$ and the additive noise $\epsilon_k$ can be either  sub-Gaussian or heavy-tailed. Given the full data $\{(X_k,Y_k):k\in[n]\}$, we first study a novel setting where both $X_k$ and $Y_k$ are quantized to binary data\footnote{The novelty here is that the covariate $X_k$ is quantized, while all prior works on quantized compressed sensing only considered the quantization of $Y_k$ (we refer to the survey \cite{dirksen2019quantized}).}. The covariate $X_k$ is quantized by exactly the same method as Section \ref{section2}. For sub-Gaussian $X_k$ and $\epsilon_k$, the response $Y_k$ is quantized to be $\mathrm{sign}(Y_k+\Lambda_k)$ with $\Lambda_k$ uniformly distributed on $[-\gamma,\gamma]$. When $X_k$ and $\epsilon_k$ are heavy-tailed (with bounded fourth moment), we truncate $Y_k$ to be $\widetilde{Y}_k$ and then similarly apply the dithered quantization to $\widetilde{Y}_k$. The estimation relies on the 1-bit sparse covariance matrix estimator $\bm{\widehat{\Sigma}}$ developed in Section \ref{section2}. To deal with the lack of positive semi-definiteness, we assume $\bm{\Sigma}_{XX}=\mathbbm{E}X_kX_k^T$ has column-wise sparsity, which accommodates the conventional isotropic condition (i.e., $\bm{\Sigma}_{XX} = \bm{I_d}$) used in compressed sensing. We formulate the recovery as a convex programming problem with objective function combining a generalized quadratic loss and an $\ell_1$ regularizer, see (\ref{3.18}). In sub-Gaussian case, we show our estimator $\widehat{\Theta}$ could achieve a near optimal minimax rate of 
     $$
     \|\widehat{\Theta}-\Theta^*\|_2 \lesssim \sqrt{s}\Big(\log n \sqrt{\frac{\log d}{n}}\Big)^{1-q/2}.
     $$
    In heavy-tailed case, our estimator possesses the error rate
    $$
    \|\widehat{\Theta}-\Theta^*\|_2 \lesssim \sqrt{s}\Big(\frac{\log d}{n}\Big)^{(1-q/2)/4}.
    $$  
    Besides the first results for this new setting, we also revisit the canonical 1-bit compressed sensing   problem where we quantize $Y_k$ in a same manner but have full knowledge of $X_k$. We estimate $\Theta^*$ via analogous convex programming problems, see (\ref{add72_5}) and (\ref{add72_8}). In sub-Gaussian regime, our estimator achieves a near optimal minimax rate 
     $$
     \|\widehat{\Theta}-\Theta^*\|_2 \lesssim \sqrt{s}\Big( \sqrt{\frac{\log d \log n }{n}}\Big)^{1-q/2}.
     $$
    In heavy-tailed regime, our estimator   still enjoys an error bound 
      $$
    \|\widehat{\Theta}-\Theta^*\|_2 \lesssim \sqrt{s}\Big(\frac{\log d}{n}\Big)^{(1-q/2)/3},
    $$ 
    which is almost tight for the specific estimation problem where the 1-bit observation is produced by our scheme with the specified parameters (Theorem \ref{lower2}). As it turns out, these two results embrace some  improvements on existing ones (e.g., recovery via convex programming, faster error rate), see a detailed comparison in Appendix \ref{appendxD}.

    \item In Section \ref{section4}, we  study  the problem of low-rank matrix completion   $Y_k=\left<\bm{X_k},\bm{\Theta^*} \right> + \epsilon_k$, where the desired $d\times d$ matrix 
    $\bm{\Theta^*} $ with singular values $\sigma_1(\bm{\Theta^*}) \geq ... \geq \sigma_d(\bm{\Theta^*})$ is (approximately) low-rank $\sum_{k=1}^d \sigma_k(\bm{\Theta^*})^q \leq r$ for some $0\leq q<1$ and $r>0$. The covariate $\bm{X_k} $ is uniformly distributed on $\{e_ie_j^T:i,j\in [d]\}$ where $e_i$ is the $i$-th column of the $\bm{I_d}$, $\epsilon_k$ is sub-Gaussian or heavy-tailed noise. Given the full data   $\{(\bm{X_k},Y_k)\}$, we quantize $Y_k$  to one bit by   the same process as 1-bit compressed sensing in Section \ref{section3}. Our estimator $\bm{\widehat{\Theta}}$ is given by minimizing an objective functions constituted of a generalized quadratic loss and a nuclear norm penalty, see (\ref{4.6}). If $\epsilon_k$ is sub-Gaussian, we show that $\bm{\widehat{\Theta}}$   achieves a near optimal minimax rate 
    $$
   \frac{ \|\bm{\widehat{\Theta}- \Theta^*}\|^2_{\mathrm{F}} }{d^2} \lesssim rd^{-q}\Big(\log n \frac{d \log d}{n}\Big)^{1-q/2},
    $$
   If $\epsilon_k$ is heavy-tailed with bounded second moment, we show the recovery guarantee
    $$ \frac{ \|\bm{\widehat{\Theta}- \Theta^*}\|^2_{\mathrm{F}} }{d^2}  \lesssim rd^{-q} \Big(\frac{d\log d}{n}\Big)^{1/2-q/4},
    $$
    which is  almost tight if one only has access to the 1-bit observation produced by our quantization scheme with the chosen parameters (Theorem \ref{lower1}). We emphasize that our approach is totally different from the existing method for 1-bit matrix completion, i.e., based on maximizing a likelihood function. Our essential advantage is that our method can handle unknown pre-quantization random noise that can even be heavy-tailed. See more discussions in Appendix \ref{appendxD}. 
\end{itemize}


   The rest of the paper is structured as follows. In the remainder of Section \ref{section1} we introduce the notation, propose the 1-bit quantization scheme, and provide some intuitions for our results (i.e., near optimality in sub-Gaussian case, and the rate degradation in heavy-tailed case). Our main results for three estimation problems are presented in Sections \ref{section2}-\ref{section4}; In Section \ref{proofsketch} we provide an overview of the proofs and the main techniques; In Section \ref{section5}, we present experimental results to corroborate our theories; We finally give some concluding remarks in Section \ref{section6}. The complete proofs are provided in Appendices \ref{appendixA}-\ref{appendixC}. Also, a comparison between this work and the literature is given in Appendix \ref{appendxD} (review of related work is provided there to avoid a lengthy beginning), while details of the numerical simulations are deferred to Appendix \ref{appendixE}. 

\subsection{Notations and Preliminaries}

As general principles, lowercase letters (e.g., $s$, $r$) represent scalars, capital letters (e.g., $X$, $Y$) represent vectors, and capital bold letters (e.g., $\bm{X, \Theta}$) represent matrices. Some exceptions are that we use capital letter $Y,Y_k$ to denote the responses, $\Lambda,\Lambda_k$ to denote the dithering noise for $Y,Y_k$, and $X_{k,i}$ for the i-th entry of $X_k$. Notations marked by $*$ denote the desired underlying signals, e.g., $\bm{\Sigma^*}, \Theta^*, \bm{\Theta^*}$, while those with a hat denote our estimators, e.g., $\bm{\widehat{\Sigma}},\widehat{\Theta},\bm{\widehat{\Theta}}$.

We first introduce different vector or matrix norms. Let $[N]=\{1,2,...,N\}$. For a vector $X = [x_i]\in \mathbb{R}^d$, the $\ell_1$ norm, $\ell_2$ norm and max norm are given by $\|X\|_1= \sum_{i=1}^d|x_i|$, $\|X\|_{2} = (\sum_{i=1}^d|x_i|^2)^{1/2}$, $\|X\|_{\max} = \max_{i\in[d]}|x_i|$, respectively. Note that we also use $\|X\|_0$ to denote the number of non-zero entries in $X$. For a matrix $\bm{X}=[x_{ij}] \in \mathbb{R}^{d\times d}$, the operator norm, Frobenius norm and max norm are defined as $\|\bm{X}\|_{\mathrm{op}}=\sup_{\|V\|_2=1}\|\bm{X}V\|_2$, $\|\bm{X}\|_{\mathrm{F}}=(\sum_{i=1}^d\sum_{j=1}^dx^2_{ij})^{1/2}$, $\|\bm{X}\|_{\max}=\max_{1\leq i,j\leq d}|x_{ij}|$. Assume the singular values are $\sigma_1(\bm{X})\geq \sigma_2(\bm{X})\geq ...\geq \sigma_d(\bm{X})$, then  the nuclear norm $\|\bm{X}\|_{\mathrm{nu}} = \sum_{i=1}^d \sigma_i(\bm{X})$ serves as the counterpart of the $\ell_1$ norm of vectors. Given $\bm{A} = [\alpha_1,...,\alpha_d] \in \mathbb{R}^{d\times d}$, we use $\mathrm{vec}(\cdot)$ to vectorize $\bm{A}$ in a column-wise manner, i.e., $\mathrm{vec}(\bm{A}) = [\alpha_1^T,\alpha_2^T,...,\alpha_d^T]^T,$ while the inverse of $\mathrm{vec}(\cdot)$ is denoted by $\mathrm{mat}(\cdot)$. Assume $\bm{B}\in \mathbb{R}^{d\times d}$, then the inner product in $\mathbb{R}^{d\times d}$ is defined as 
$\left<\bm{A,B}\right> = \mathrm{Tr}(\bm{A}^T\bm{B}) = \mathrm{vec}(\bm{A})^T\mathrm{vec}(\bm{B}).$

Throughout the paper, we use $n$ to denote the number of samples in data, while $d$ the ambient dimension of the problem. Expectation and probability are denoted by $\mathbbm{E}(\cdot)$, $\mathbb{P}(\cdot)$ respectively. For a specific event $E$, $\mathbbm{1}(E)$ stands for the corresponding indicator function, i.e., $\mathbbm{1}(E)=1$ if $E$ happens, $\mathbbm{1}(E)=0$ otherwise. We work with quite a lot of parameters arising in several signal processing steps. To avoid confusion of constants, we use $\{D_1,D_2,D_3,...\}$ to denote constants whose values may vary from line to line, while $\{C_1,C_2,C_3,...\}$ would only be used once to set a specific parameter, see (\ref{2.6}), (\ref{2.9}) for example.

We adopt standard asymptotic  notations that omits absolute constants. Specifically, we use $B_1\lesssim B_2$ or $B_1 = O(B_2)$ to abbreviate the fact that $B_1 \leq CB_2$ for some absolute constant $C$. Similarly, we write $B_1\gtrsim B_2$ or alternatively $B_1=\Omega(B_2)$ if $B_1 \geq CB_2$ for some $C>0$. If both $B_1 =O(B_2)$ and $B_1 = \Omega(B_2)$ hold, i.e., $B_1$ equals $B_2$ up to constants, we write $B_1 \asymp B_2$.

The function $\sign(\cdot)$ extracts the sign of a real number $x$, i.e., $\sign(x)=1$ if $x\geq 0$, $\sign(x)=-1$ if $x<0$. Hard thresholding operator with threshold $\zeta$ is defined by $\mathcal{T}_{\zeta}(x) = x\mathbbm{1}(|x|\geq \zeta)$. Both $\mathrm{sign}(\cdot)$ and $\mathcal{T}_{\zeta}(\cdot)$ operate on vectors or matrices element-wisely.

To broaden the range of our readers, we give some preliminaries on sub-Gaussian random variable or concentration inequality as follows. 
\begin{definition}
Given a real random variable $X\in\mathbb{R}$, its sub-Gaussian norm $\|X\|_{\psi_2}$, sub-exponential norm $\|X\|_{\psi_1}$  are defined as \begin{equation}
        \|X\|_{\psi_2} = \inf \Big\{t>0:\mathbbm{E} \exp \Big(\frac{X^2}{t^2}\Big) \leq 2\Big\}~~,~~\|X\|_{\psi_1} = \inf \Big\{t>0:\mathbbm{E} \exp \Big(\frac{|X|}{t}\Big) \leq 2\Big\}.
        \label{n1.1}
    \end{equation}
   $X$ is said to be sub-Gaussian if $\|X\|_{\psi_2}\leq\infty$.
\end{definition}
 
\begin{definition}
Given a real random vector $X\in \mathbb{R}^d$, the sub-Gaussian norm is defined as $\|X\|_{\psi_2} = \sup_{\|V\|_2 = 1} \|V^TX\|_{\psi_2}$. $X$ is said to be sub-Gaussian if $\|X\|_{\psi_2} \leq \infty$.
\end{definition} 
For $X,Y\in\mathbb{R}$ we note a useful relation (see \cite[Lemma 2.7.7]{vershynin2018high})
\begin{equation}
    \label{n1.2}
    \|XY\|_{\psi_2} \leq \|X\|_{\psi_1}\|Y\|_{\psi_1}.
\end{equation}

Sub-Gaussian variable $X$ has properties similar to the Gaussian one, such as light probability tail and bounded moment constraint. 

\begin{pro}
\label{pro1}
{\rm (Proposition 2.5.2, \cite{vershynin2018high}){\bf \sffamily.}} Assume random variable $X$ is sub-Gaussian, then for absolute constants $D_1,D_2$ we have:

\vspace{1mm}

\noindent{\rm (a)} For any $t>0$, $\mathbbm{P}\big(|X|\geq t\big)\leq 2\exp \big(- \frac{D_1t^2}{\|X\|_{\psi_2}^2}\big).$

\noindent{\rm (b)} For any $p\geq 1$, $\big(\mathbbm{E}|X|^p\big)^{1/p} \leq D_2\|X\|_{\psi_2}\sqrt{p}$. 
\end{pro}
  
 \begin{pro}
 \label{pro2}
 {\rm(Proposition 2.6.1, \cite{vershynin2018high}){\bf \sffamily.}}  Let $X_1,...,X_N$ be independent, zero-mean, sub-Gaussian random variables, then for some absolute constant $D_1$ we have $\big\|\sum_{k=1}^NX_k\big\|_{\psi_2}^2\leq D_1 \sum_{k=1}^N \|X_k\|_{\psi_2}^2$. 
 \end{pro}
  For concentration results, we only introduce Hoeffding's inequality and Bernstein's inequality. Several other concentration inequalities (e.g., Matrix Bernstein's inequality) would be properly referred to the sources when they are invoked in the proof.
\begin{pro}
\label{pro3}{\rm(Hoeffding's inequality, \cite[Theorem 1.9]{rigollet2015high}){\bf \sffamily.}} Let $X_1,...,X_n$ be independent, bounded random variables satisfying $X_i\in[a_i,b_i]$, then for any $t>0$ it holds that
\begin{equation}
\mathbbm{P}\Big(\Big|\frac{1}{n}\sum_{k=1}^n(X_k-\mathbbm{E}X_k)\Big|\geq t\Big)\leq 2\exp\Big(-\frac{2n^2t^2}{\sum_{i=1}^n(b_i-a_i)^2}\Big).
    \label{1.3}
\end{equation}
\end{pro}
\begin{pro}
\label{pro4}
{\rm(Bernstein's inequality, \cite[ Theorem 2.8.1]{vershynin2018high}){\bf \sffamily.}} Let $X_1,...,X_N$ be independent random variables, then for any $t>0$ and for some absolute constant $D_1$ we have 
\begin{equation}
    \label{1.4}
    \mathbbm{P}\Big(\big|\sum_{k=1}^N (X_k - \mathbbm{E}X_k)\big|\geq t\Big) \leq 2\exp\Big(-D_1\min\big\{\frac{t^2}{\sum_{k=1}^N\|X_k\|_{\psi_1}^2}, \frac{t}{\max_{k\in [N]} \|X_k\|_{\psi_1}}\big\}\Big). 
\end{equation}
\end{pro}

Although sub-Gaussian data has exciting statistical properties like similar tail bounds as Gaussian distribution, data in some real problems may have much heavier tail, to name a few, data in economics and finance \cite{ibragimov2015heavy}, biomedical data \cite{biswas2007statistical,woolson2011statistical}, noise in signal processing \cite{swami2002some,wang2021robust}, and even signal itself \cite{kruczek2020detect,aysal2006second,li2012heavy}. Therefore, we will also consider the 1-bit quantization of heavy-tailed data. We use bounded moment of some order to capture the heavy-tailedness, i.e., $\mathbbm{E}|X|^{l}\leq M$ for some $l>0$. Note that this is a widely used definition \cite{fan2021shrinkage,fan2021robust,sun2020adaptive,zhu2021taming,wang2021robust,ke2019user,hu2022high}.

\subsection{1-bit Quantization Scheme}
Truncation, dithering and quantization are three typical signal processing steps in our work.  We  {summarize}   our 1-bit quantization scheme as follows:
\begin{enumerate}
    \item  \textbf{Truncation.} The truncation step will only be used to heavy-tailed data. Specifically, we first specify a threshold $\eta>0$, then the truncation step shrinks a scalar $x$ to be $\mathrm{sign}(x)\min\{|x|,\eta\}$, and hence $x$ with magnitude smaller than $\eta$   remains unchanged in truncation. Vectors are truncated element-wisely. Notations marked by tilde are used exclusively to denote truncated data, for example, $\widetilde{X}_{k}$ and $\widetilde{Y}_k$.
    \item \textbf{Dithering.} The dithering step is applied to all the data that we plan to quantize to 1 bit. For $E\subset\mathbb{R}^m$, we use $X\sim \mathrm{uni}(E)$ to state that $X$ obeys uniform distribution on $E$. In sub-Gaussian case we dither the covariate $X_k$ and response $Y_k$ by uniformly distributed noise. Note that   we need to sample two bits per entry for $X_k$ (the covariate in sparse linear regression). Thus, we draw $\Gamma_{k1},\Gamma_{k2}\sim \mathrm{uni}([-\gamma,\gamma]^d)$ and dither $X_k$ to be $X_k+\Gamma_{k1}$, $X_k+\Gamma_{k2}$. We only need 1-bit information for each response $Y_k$, so we sample $\Lambda_k \sim \uni([-\gamma,\gamma])$ and obtain the dithered response $Y_k+\Lambda_k$. In heavy-tailed case $X_k$ and $Y_k$ are substituted with the truncated data $\widetilde{X}_k$ and $\widetilde{Y}_k$.
    
    \item \textbf{Quantization.} In quantization step we simply apply $\sign(\cdot)$ to the dithered data, and notations marked by a dot (e.g., $\dot{Y}_k,\dot{X}_{k1},\dot{X}_{k2}$) exclusively represent the 1-bit quantized data. More precisely, we have $\dot{Y}_k = \mathrm{sign}(Y_k+\Lambda_k)$, $\dot{X}_{kj}= \mathrm{sign}(X_{k}+\Gamma_{kj})$, $j=1,2$ for sub-Gaussian $X_k$, $Y_k$, and $\dot{Y}_k =\mathrm{sign}(\widetilde{Y}_k+\Lambda_k)$, $\dot{X}_{kj}  =\mathrm{sign}(\widetilde{X}_{k}+\Gamma_{kj}) $, $j=1,2$ for heavy-tailed $X$ and $Y$.
\end{enumerate}

\subsection{Intuition and Heuristic}
\label{intuition}
Here we illustrate the intuition of our results before proceeding to details. Readers familiar with the 1-bit quantization with uniform dithering noise can directly skip this part.

Specifically, we will  heuristically analyse a multi-bit matrix completion setting to illustrate the   the reason why  our estimators could achieve near optimal minimax rates  in sub-Gaussian regime. In fact, the idea of the whole paper stems from two simple observations, which are given in the following two lemmas. We mention that Corollary \ref{corollary1}   motivates \cite{dirksen2022covariance} to estimate $\mathbbm{E}(XY)$ and hence an unstructured covariance matrix via binary data, while Lemma \ref{lemma1} is its more elementary version and enlightens the estimators in our work. For instance, while full observations are not available, our loss function in matrix completion is constructed by  substituting the full data $Y_k$ in the empirical $\ell_2$ loss with the 1-bit surrogate $\gamma\cdot \dot{Y}_k$ (see (\ref{4.6})). This idea comes from Lemma \ref{lemma1}. 

\begin{lem}
Let $X,\Lambda$ be two independent random variables satisfying $|X|\leq B$, $\Lambda\sim \uni\big([-\gamma,\gamma]\big)$ where $\gamma\geq B$, then we have $\mathbbm{E}\big[\gamma \cdot \mathrm{sign}(X + \Lambda)\big] = \mathbbm{E}X$.
\label{lemma1}
\end{lem}
\begin{cor}{\rm(Lemma 16 in \cite{dirksen2022covariance}){\bf \sffamily.}} Let $X,Y$ be bounded random variables satisfying $|X|\leq B$, $|Y|\leq B$, $\Lambda_1,\Lambda_2$ are i.i.d. uniformly distributed on $[-\gamma,\gamma]$, $\gamma \geq B$, and $\Lambda_1,\Lambda_2$ are independent of X, Y. Then we have $\mathbbm{E}\big[\gamma^2\cdot\mathrm{sign}(X+\Lambda_1)\cdot\mathrm{sign}(Y+\Lambda_2)\big] = \mathbbm{E}XY$. 
\label{corollary1}
\end{cor}

Next, 
by informal arguments, we heuristically compare 
full-data-based matrix completion  and quantized matrix completion where one can sample finitely many bits from each $Y_k$ (we refer it as multi-bit matrix completion). This comparison can provide some insights of why our estimators can achieve a near optimal  minimax rate in sub-Gaussian regime.

We consider a   full-data sample of size $n$ from matrix completion  (\ref{4.1}) and denote it by $$\mathcal{D}_{\mathrm{full}} = \Big\{(\bm{X_1},Y_1),...,(\bm{X_n},Y_n)\Big\}.$$ 
For some positive integer $f(n)  $ we i.i.d. draw $\{\Lambda_{kj}:j\in [f(n)]\}$ from $\mathrm{uni}\big([-\gamma,\gamma]\big)$, and sample $f(n)$ bits from each $Y_k$ by the proposed dithered quantization, that is, $\{\dot{Y}_{kj} := \mathrm{sign}(Y_k +\Lambda_{kj} ):j\in [f(n)]\}$. This quantization process yields the  sample containing $n\cdot f(n)$ binary observations  $$\mathcal{D}_{\mathrm{mult}}= \Big\{(\bm{X_k},\dot{Y}_{kj}):k\in [n],~j\in [f(n)]\Big\}.$$ Interestingly, from $\mathcal{D}_{\mathrm{mult}}$ one can build a dataset with size $n$ as $$\mathcal{D}_{\mathrm{appr}}= \left\{ (\bm{X_k},Y_{k,\mathrm{appr}}):Y_{k,\mathrm{appr}}  = \frac{1}{f(n)}\sum_{j\in [f(n)]} \gamma \cdot \dot{Y}_{kj},~k\in [n]\right\}.$$
We aim to reveal that the above three samples are comparably informative for the estimation.

 For simplicity we assume $\|\epsilon_k\|_{\psi_2}=O(1)$, $\|Y_k\|_{\psi_2} = O(1)$, then with probability at least $1-O(n^{-\Omega(1)})$ we have $\max_k |Y_k| = O(\sqrt{\log n})$ (see \cite[Theorem 1.14]{rigollet2015high}). Thus, we can choose  $\gamma =  \mathrm{Poly}(\log n)$\footnote{Here $\mathrm{Poly}(\log n)$   denotes any term $T$ satisfying $T = O([\log n]^m)$ for some positive integer $m$.} to guarantee $\gamma >\max_k |Y_k|$ with high probability. We proceed the analysis on this event. Define $\epsilon_{k,\mathrm{appr}} := Y_{k,\mathrm{appr}}- Y_k$,   equivalently we can write
\begin{equation}
    \label{heu}
    Y_{k,\mathrm{appr}} = Y_k + \epsilon_{k,\mathrm{appr}} = \big<\bm{X_k},\bm{\Theta^*}\big> + \epsilon_k + \epsilon_{k,\mathrm{appr}}.
\end{equation}
For $\epsilon_{k,\mathrm{appr}}$, Lemma \ref{lemma1} gives $\mathbbm{E}_{\Lambda_{kj}}(\gamma \cdot \dot{Y}_{kj}) = Y_k$ and hence $\mathbbm{E}\epsilon_{k,\mathrm{appr}}= 0$. Moreover,   conditioned on $Y_k$,   $\epsilon_{k,\mathrm{appr}}$ is the mean of $f(n)$ zero-mean, independent random variables lying in $[-\gamma - Y_k, \gamma - Y_k]$. Thus, Proposition \ref{pro2} and Hoeffding's Lemma (see Lemma 1.8, \cite{rigollet2015high}) give \begin{equation}
\label{712_1}
    \|\epsilon_{k,\mathrm{appr}}\|_{\psi_2} =O\Big( \frac{\gamma}{\sqrt{f(n)}}\Big) .
\end{equation}
Therefore, $\|\epsilon_{k,\mathrm{appr}}\|_{\psi_2}= O\big(1\big)$ as long as $f(n)$ dominates $\gamma^2$, while   $f(n)=\mathrm{Poly}(\log n)$ would suffice due to $\gamma = \mathrm{Poly}(\log n)$. In conclusion, $\mathcal{D}_{\mathrm{mult}}$  containing $n\cdot \mathrm{Poly}(\log n)$ binary data can generate the sample  $\mathcal{D}_{\mathrm{appr}}$  of size $n$, where each $Y_{k,\mathrm{appr}}$ can be viewed as a full observation from (\ref{heu}). Moreover, since $\|\epsilon_k\|_{\psi_2} = \|\epsilon_{k,\mathrm{appr}}\|_{\psi_2} = O(1)$,  (\ref{heu}) is almost equivalent to the original model (\ref{4.1}). This reveals $\mathcal{D}_{\mathrm{appr}}$, and hence $\mathcal{D}_{\mathrm{appr}}$ with $n\cdot\mathrm{Poly}(\log n)$ binary observations, are comparable to $\mathcal{D}_{\mathrm{full}}$ with $n$ full observations. Furthermore, this indicates the inessential logarithmic degradation of recovery error after 1-bit quantization.

Note that similar heuristics can be found in sub-Gaussian regime of (sparse) covariance matrix estimation and sparse linear regression. 
Of course, such multi-bit heuristic deviates from the 1-bit setting where we collect only 1 bit from each $Y_k$ (see the following graphical illustration). But since $f(n): = \mathrm{Poly}(\log n)$ is
negligible compared with $n$, one may tend to believe $\mathcal{D}_{\mathrm{mult}}$ and    $\mathcal{D}_{\mathrm{1bit}} = \{\dot{Y}_{k}:=\mathrm{sign}(Y_k+\Lambda_k):k\in [n\cdot f(n)]\}$  are   comparable. From this perspective,  the near-optimal rates in sub-Gaussian regime are merely matter of courses.

\begin{tikzpicture}[>=stealth,->,shorten >=1pt,looseness=.5,auto]
\node[] at (-3.2,-4) {(A heuristic multi-bit setting)};
\node[] at (4.3,-4) {(The 1-bit setting)};
\node[] at (-0.3,0.15) {$\vdots$};
\node[] at (-0.3,-0.4) {$\vdots$};
\node[] at (-6.5,0.15) {$\vdots$};
\node[] at (-6.5,-0.4) {$\vdots$};
\matrix [matrix of math nodes,
column sep={3.1cm,between origins},
row sep={0.8cm,between origins},
nodes={}]
{
& |(A)| ~\dot{Y}_{11} & &     |(O1)| \stackrel{~}{Y_1} &  |(Q1)|\dot{Y_1}\\
  & |(E)| ~\dot{Y}_{12} & &     |(O2)| \stackrel{~}{Y_2} &  |(Q2)|\dot{Y}_2\\
|(B)| Y_1& |(C)| \vdots & |(F)| Y_{1,\mathrm{appr}}&  \vdots &  \vdots\\
 & |(D)| \dot{Y}_{1f(n)} & &  |(Of)| \stackrel{~}{Y_{f(n)}} & |(Qf)| \dot{Y}_{f(n)}\\
 \vspace{13mm}
 & |(An)| ~\dot{Y}_{n1}  & &    |(On-1)| \stackrel{~}{Y_{(n-1)f(n)+1}} &  |(Qn-1)|\dot{Y}_{(n-1)f(n)+1}\\
  & |(En)| ~\dot{Y}_{n2} & &    |(On-12)| \stackrel{~}{Y_{(n-1)f(n)+2}} & |(Qn-12)|\dot{Y}_{(n-1)f(n)+2}\\
 |(Bn)| Y_n& |(C)| \vdots &|(Fn)| Y_{n,\mathrm{appr}}&\vdots & \vdots\\
 & |(Dn)| \dot{Y}_{nf(n)} & & |(Onf)| \stackrel{~}{Y_{nf(n)}} &  |(Qnf)|\dot{Y}_{nf(n)}\\
 (\mathcal{D}_{\mathrm{full}}) &  (\mathcal{D}_{\mathrm{mult}} )&  (\mathcal{D}_{\mathrm{appr}}) &  &  (\mathcal{D}_{\mathrm{1bit}}) \\
};
\begin{scope}[every node/.style={font=\small\itshape}]
\draw (O1) --   (Q1);
\draw (O2) --   (Q2);
\draw (Of) --   (Qf);
\draw (On-1) --   (Qn-1);
\draw (On-12) --   (Qn-12);
\draw (Onf) --   (Qnf);

\draw (B) --   (A);
\draw (B) --   (D);
\draw (B) --   (E);
\draw (A) --   (F);
\draw (D) --   (F);
\draw (E) --   (F);
 
\draw (An) --   (Fn);
\draw (Dn) --   (Fn);
\draw (En) --   (Fn);
\draw (Bn) --   (An);
\draw (Bn) --   (Dn);
\draw (Bn) --   (En);
\end{scope}
\end{tikzpicture}

However, in heavy-tailed regime the story becomes totally different. Specifically, $\gamma = \mathrm{Poly}(\log n)$ will no longer guarantee $\gamma > \max_k|Y_k|$ with high probability. When this vital condition fails, the dithering becomes invalid for responses with absolute value larger than $\gamma$. Indeed, for these measurements the proposed dithered quantization reduces to a direct collection of the  sign, while under such direct quantization we even lose the well-posedness of the problem (e.g., matrix completion, see \cite{davenport20141}) or the possibility of full  signal reconstruction (e.g., 1-bit compressed sensing, see \cite{plan2013one}).

To resolve the issue, we truncate the heavy-tailed data according to some threshold $\eta$, which produces data bounded by $\eta$. Then we can treat the truncated data as sub-Gaussian data and   use dithering noise drawn from $\mathrm{uni}\big([-\gamma,\gamma]\big)$ with $\gamma > \eta$. It is not hard to see that $\eta$ represents   the data bias   introduced in truncation. More precisely, smaller $\eta$ corresponds to larger bias. Because of Hoeffding's Lemma, $\gamma$ is positively related to data variance.   Definitely, for estimation or signal recovery we prefer data with small bias (i.e., big $\eta$) and small variance (i.e., small $\gamma$). But, note that we also need $\gamma > \eta$ to enforce the effectiveness of dithering. Thus, a trade-off between bias and variance is needed. We comment that, making an optimal balance between bias and variance leads to our error rates in heavy-tailed regime.  See Example 1 in  Section \ref{proofsketch} for instance.

\section{Sparse Covariance Matrix Estimation}
\label{section2}
We start from the problem of estimating a sparse covariance matrix. Let $X\in \mathbb{R}^d$ be a random vector with zero mean, the i.i.d. realizations $X_k$ are quantized to 1-bit data $(\dot{X}_{k1},\dot{X}_{k2})$, and we aim to estimate the underlying covariance matrix $\bm{\Sigma^*} = \mathbbm{E}XX^T$ based on the quantized data.

We first ideally assume the underlying $d$-dimensional random vector $X$ has entries bounded by $\gamma$, then  Corollary \ref{corollary1} delivers that $\mathbbm{E}\big[\gamma^2\cdot \dot{X}_{k1}\dot{X}_{k2}^T\big] = \mathbbm{E}XX^T = \Sigma^*$, which is just the desired covariance matrix. Besides, the concentration of $\gamma^2 \cdot\dot{X}_{k1}\dot{X}_{k2}^T $ should be fast due to boundedness, see Hoeffding's inequality in Proposition \ref{pro3}. Combining the two observations, \cite{dirksen2022covariance} proposed a covariance matrix estimator as an empirical version of $\mathbbm{E}\big[\gamma^2\cdot \dot{X}_{k1}\dot{X}_{k2}^T\big]$, followed by symmetrization:
\begin{equation}
    \bm{\breve{\Sigma}} = \frac{\gamma^2}{2n}\sum_{k=1}^n \Big[\dot{X}_{k1}\dot{X}_{k2}^T+\dot{X}_{k2}\dot{X}_{k1}^T \Big].
    \label{2.1}
\end{equation}
For sub-Gaussian $X_k$, this estimator achieves a near minimax rate (compared with full data setting in \cite{cai2010optimal})
\begin{equation}
    \|\bm{\breve{\Sigma}} - \bm{\Sigma^*}\|_{\mathrm{op}} \lesssim \log n \sqrt{\frac{d\log d}{n}}.
\label{2.2}
\end{equation}  
Here, we point out that sampling two bits (rather than one bit) per entry is merely for estimating the diagonal entries of $\bm{\Sigma^*}$, since the 1-bit version of (\ref{2.1}), $$\bm{\breve{\Sigma}}_{1\mathrm{bit}} = \frac{\gamma^2}{n}\sum_{k=1}^n \dot{X}_{k1}\dot{X}_{k1}^T,$$always gives $\gamma^2$ in the diagonal and hence fails to recover the diagonal of the covariance matrix. 


It is evident that (\ref{2.2}) requires at least $n\gtrsim d$ to provide a non-trivial error bound. Actually it has been reported that even the sample covariance matrix $\sum_{k}X_kX_k^T/n$ has extremely poor performance under high dimensional scaling where $d\geq n$ \cite{johnstone2001distribution}, not to mention (\ref{2.1}). On the other hand, high-dimensional databases are undoubtedly becoming ubiquitous in genomics, biomedical, imaging, tomography, finance and so forth, while covariance matrix plays a fundamental role in the analysis of these databases.

To address the high-dimensional issue, extra structures are necessary to reduce the intrinsic problem dimensionality. For covariance matrix we usually have sparsity as prior knowledge, especially in the situation where dependencies among different features are weak, for instance, the Genomics data \cite{efron2012large}, functional data drawn from underlying curves \cite{ramsey2005functional}. A precise formulation of the sparse structure  is provided in Assumption \ref{assumption1}.
\begin{assumption}
{\rm(Approximate column-wise sparsity){\bf \sffamily.}} For a specific $0\leq q<1$, the columns of covariance matrix $\bm{\Sigma^*}=[\sigma^*_{ij}]$ are approximately sparse in the sense that 
\begin{equation}
    \sup_{j\in[d]}\sum_{i=1}^d |\sigma^*_{ij}|^q \leq s
    \label{2.3}
\end{equation}
\label{assumption1}
\end{assumption}

In literature there are two mainstreams to incorporate sparsity into covariance matrix estimation, namely penalized likelihood method  \cite{bien2011sparse,rutimann2009high} and \textcolor{black}{a thresholding method} \cite{bickel2008covariance,cai2013optimal,el2008operator,cai2011adaptive,cai2012minimax}. Thresholding method refers to the direct regularizer that element-wisely hard thresholding the sample covariance matrix, i.e., $\mathcal{T}_{\zeta}(\sum_{k=1}^nX_kX_k^T/n)$, which promotes sparsity intuitively. With suitable threshold $\zeta$, Cai and Zhou \cite{cai2012optimal} showed $\mathcal{T}_{\zeta}(\sum_{k=1}^nX_kX_k^T/n)$
could   achieve  minimax  rate under operator norm over the class of column-wisely sparse covariance matrices (Assumption \ref{assumption1}). Motivated by previous work, we propose to hard thresholding $\bm{\breve{\Sigma}}$ in (\ref{2.1}) to obtain a high-dimensional estimator $\bm{\widehat{\Sigma}} = [\widehat{\sigma}_{ij}]$ given by 
\begin{equation}
\bm{\widehat{\Sigma} }= \mathcal{T}_{\zeta}\bm{\breve{\Sigma} }.
    \label{2.4}
\end{equation}
The statistical rates of $\bm{\widehat{\Sigma} }$ under both max norm and operator norm are established in what follows. 

\subsection{Sub-Gaussian Data}
Assume $X_k = [X_{k,1},X_{k,2},...,X_{k,d}]$ are i.i.d. sampled from a random vector $X\in \mathbb{R}^d$ with zero-mean sub-Gaussian components. In particular, we assume \begin{equation}
  \mathbbm{E}X_k = 0,~ \|X_{k,i}\|_{\psi_2}\leq \sigma, \ \forall i \in [d].
    \label{2.5}
\end{equation}  
From (\ref{2.4}),   $\bm{\breve{\Sigma}}= [\breve{\sigma}_{ij}]$ serves as an intermediate estimator to construct $\bm{\widehat{\Sigma}}$, hence we first provide an element-wise error bound of $\bm{\breve{\Sigma}}$ in Theorem \ref{theorem1}.

\begin{theorem}
Assume (\ref{2.5}) holds. For specific $\delta \geq 1$ we assume $n>2 \delta \log d$. For some sufficiently large constant $C_1$ we set the dithering scale $\gamma$ as \begin{equation}
\gamma = C_1\sigma \sqrt{\log \Big(\frac{n}{2\delta \log d}\Big)}
    \label{2.6}
\end{equation}
and assume $\gamma > \sigma$. Then for $\bm{\breve{\Sigma}} = [\breve{\sigma}_{ij}]$ we have
\begin{equation}
\mathbbm{P}\Big(|\breve{\sigma}_{ij}- \sigma^*_{ij}| \lesssim \sigma^2 \log n \sqrt{\frac{\delta \log d}{n}}\Big) \geq 1-2d^{-\delta}
    \label{2.7}
\end{equation}
for $i,j\in[d]$. Moreover, we have the error bound for max norm 
\begin{equation}
\mathbbm{P}\Big(\|\bm{\breve{\Sigma}- \Sigma^*}\|_{\max}\lesssim \sigma^2 \log n \sqrt{\frac{\delta \log d}{n}}\Big) \geq 1-2d^{2-\delta}.
    \label{2.8}
\end{equation}
\label{theorem1}
\end{theorem}

Recall that our estimator is obtained by hard thresholding $\bm{\breve{\Sigma}}$. The next Theorem shows that with suitable threshold $\zeta$, the hard thresholding even brings a tighter statistical bound for element-wise error. 

\begin{theorem}
Assume (\ref{2.5}) holds, $\delta \geq 1$ is the same as Theorem \ref{theorem1}, and the dithering scale $\gamma$ is given as (\ref{2.6}) with some $C_1$. Then we choose the threshold $\zeta$ by 
\begin{equation}
\zeta = C_2\sigma^2\log n \sqrt{\frac{\delta \log d}{n}},
    \label{2.9}
\end{equation}
where $C_2$ is a sufficiently large constant. Then for any $i,j\in [d]$ we have 
\begin{equation}
\mathbbm{P}\Big(|\widehat{\sigma}_{ij}- \sigma^*_{ij}| \lesssim \min\Big\{|\sigma^*_{ij}|,\sigma^2 \log n \sqrt{\frac{\delta \log d}{n}}\Big\}\Big)\geq 1-2d^{-\delta}.
    \label{2.10}
\end{equation}
\label{theorem2}
\end{theorem}

By combining (\ref{2.10}) and Assumption \ref{assumption1}, we are in a position to establish the rate of $\bm{\widehat{\Sigma}}$ under operator norm. Specifically, we prove that our 1-bit estimator achieves a rate $O\big(s((\log n)^2\frac{\log d}{n})^{(1-q)/2}\big)$, which almost matches the minimax rate $O\big(s\left(\frac{\log d}{n}\right)^{(1-q)/2}\big)$ proved in   \cite[Theorem 2]{cai2012optimal}. Note that the estimator based on full data in \cite{cai2012optimal} achieves the minimax rate. From this perspective, the 1-bit quantization only introduces minor information loss to the learning process, i.e., a logarithmic factor. Thus, by using our method, one can embrace the privileges of 1-bit data and accurate covariance matrix estimation simultaneously. 

\begin{theorem}
Assume Assumption \ref{assumption1}, (\ref{2.5}) hold, $\delta $ is the same as Theorem \ref{theorem1}, \ref{theorem2} (set $\delta \geq 4$), and the dithering scale $\gamma$, the threshold $\zeta$ are respectively given by (\ref{2.6}), (\ref{2.9}) with some $C_1,C_2$. Besides, assume $\delta \log d(\log n)^2/n$ is sufficiently small. Let $p=\delta/4$, we have 
\begin{equation}
\Big(\mathbbm{E}\|\bm{\widehat{\Sigma}- \Sigma^*}\|_{\mathrm{op}}^p\Big)^{1/p}  \lesssim s\Big(\sigma^2\log n \sqrt{\frac{\delta\log d}{n}}\Big)^{1-q}.
    \label{2.12}
\end{equation}
Moreover,  the probability tail of operator norm error is bounded as
\begin{equation}
\mathbbm{P} \Big(\bm{\|\widehat{\Sigma}- \Sigma^*}\|_{\mathrm{op}} \lesssim  s\Big[\sigma^2\log n \sqrt{\frac{\delta\log d}{n}}\Big]^{1-q} \Big)\geq 1-\exp(-\delta).
    \label{2.13}
\end{equation}
\label{theorem3}
\end{theorem}
\begin{rem}
\label{remark1}
We point out that the proof of Theorem \ref{theorem3} may be of independent technical interest, especially the probabilistic inequality (\ref{2.13}) that seems quite new in the literature. In fact, only the upper bound for the second moment (i.e., $p$=2 and $\delta = 8$ in (\ref{2.12})) is obtained in literature (e.g., \cite[Theorem 3]{cai2012optimal}), and by Markov inequality this can only give a probability term $1- \frac{1}{\delta^{1-q}}$ in (\ref{2.13}). Here, by contrast, we derive a much better probabilistic term $1 - \exp(-\delta)$. The key idea is to adaptively bound the $\Omega(\delta)$-th moment rather than a specific second moment, which gives (\ref{2.12}). It is straightfoward to apply this method to the traditional full-data thrsholding estimator and gain some improvement on prior results. 
\end{rem}

To guarantee positive semi-definiteness, we introduce a trick developed in literature. Let the eigenvalue decomposition of $\bm{\widehat{\Sigma}}$ be $\sum_{i=1}^d \lambda_{i}(\bm{\widehat{\Sigma}})v_iv_i^T$, we remove the components corresponding to negative eigenvalues and obtain the positive part $\bm{\widehat{\Sigma}^+} = \sum_{i=1}^d \max(\lambda_{i}(\bm{\widehat{\Sigma}}),0)v_iv_i^T$. It is not hard to show that $\|\bm{\widehat{\Sigma}^+-\Sigma^*}\|_{\mathrm{op}}\leq 2\|\bm{\widehat{\Sigma}-\Sigma^*}\|_{\mathrm{op}}$. Thus, $\bm{\widehat{\Sigma}^+}$ retains the operator norm rate of $\bm{\widehat{\Sigma}}$. However, removing the negative components may destroy the element-wise error or the sparse pattern of $\bm{\widehat{\Sigma}}$, see \cite{pourahmadi2013high}.

Besides, it is worth noting that we present Theorem \ref{theorem3} under operator norm by convention, but both (\ref{2.12}) and (\ref{2.13}) are applicable to the larger norm $\|\bm{X}\|_{1,\infty}  = \sup_{j}\sum_i |x_{ij}|$, see   an initial step in the proof (\ref{74add_1}). 

\subsection{Heavy-tailed Data}
Let $X_k= [X_{k,1},...,X_{k,d}]^T$ be i.i.d. drawn from the random vector $X\in \mathbb{R}^d$, in this part we consider zero-mean, heavy-tailed $X$ assumed to have bounded fourth moments
\begin{equation}
\mathbbm{E}X_k = 0, \mathbbm{E}|X_{k,i}|^4\leq M,\ ~\forall ~i\in [d].
    \label{2.14}
\end{equation}
 Note that this offers great relaxation compared to sub-Gaussian random variable   and encompasses more distributions such as t-distribution, log-normal distribution.

Compared with the light tail in   Proposition \ref{pro1}(a), $X$ satisfying (\ref{2.14}) can have a much heavier tail, and so overlarge data appear more frequently. This is problematic because our dithering noise has finite scale $\gamma$, hence the dithering is invalid for data with magnitude larger than $\gamma$. More precisely, this issue can be formulated as
$$\mathrm{sign}(X_{k,i}+ \Gamma_{kj,i}) = \mathrm{sign}(X_{k,i}),\ \ \  \mathrm{if}\ |X_{k,i}|>\gamma.$$
Therefore, for those entries larger than $\gamma$, our signal processing reduces to a direct quantization without dithering noise, which is known to introduce great loss of information.

To deal with the issue, we first truncate the data larger than a specified threshold $\eta$ and obtain the truncated data $\widetilde{X}_k$ bounded by $\eta$, which is of the spirit to introduce some biases for variance reduction. Now that the truncated data are bounded, we similarly dither them by uniform noise, and then quantize to $\dot{X}_{kj}= \mathrm{sign}(\widetilde{X}_{kj}+\Gamma_{kj})$, $j=1,2$, where $\Gamma_{kj}\sim \mathrm{uni}([-\gamma,\gamma]^d)$. Motivated by Corollary \ref{corollary1}, we propose an intermediate estimator \begin{equation}
    \bm{\breve{\Sigma}} = \frac{\gamma^2}{2n}\sum_{k=1}^n \Big[\dot{X}_{k1}\dot{X}_{k2}^T+\dot{X}_{k2}\dot{X}_{k1}^T \Big],
    \label{2.15}
\end{equation}
which extends (\ref{2.1}) to heavy-tailed data. Element-wise error for $\bm{\breve{\Sigma}}$ is given in Theorem \ref{theorem4}.
\begin{theorem}
Assume (\ref{2.14}) holds. For some fixed $\delta \geq 1$ and $C_3,C_4$ ($C_4>C_3$), we set the truncation parameter $\eta$ and the dithering scale $\gamma$ by 
\begin{equation}
\begin{cases}
 \displaystyle     \eta = C_3M^{1/4}\Big(\frac{n}{\delta \log d}\Big)^{1/8} \\
   \displaystyle   \gamma = C_4M^{1/4}\Big(\frac{n}{\delta \log d}\Big)^{1/8}
\end{cases},
    \label{2.16}
\end{equation}
Then for $\bm{\breve{\Sigma}}= [\breve{\sigma}_{ij}]$ given in (\ref{2.15}), we have
\begin{equation}
\mathbbm{P}\Big(|\breve{\sigma}_{ij}-\sigma^*_{ij}| \lesssim \sqrt{M}\Big[\frac{\delta \log d}{n}\Big]^{1/4}\Big) \geq 1-2d^{\delta} .
    \label{2.17}
\end{equation}
Moreover, we have the error bound under max norm 
\begin{equation}
\mathbbm{P}\Big(\|\bm{\breve{\Sigma}- \Sigma^*}\|_{\max} \lesssim \sqrt{M}\Big[\frac{\delta \log d}{n}\Big]^{1/4}\Big) \geq 1-2d^{2-\delta} .
    \label{2.18}
\end{equation}
\label{theorem4}
\end{theorem}

Parallel to the sub-Gaussian regime, we use an additional hard thresholding step to promote sparsity. That is, based on the intermediate estimator $\bm{\breve{\Sigma}}$ in (\ref{2.15}), we choose some suitable thresholding parameter $\zeta$ and define the estimator
\begin{equation}
    \bm{\widehat{\Sigma}} = \mathcal{T}_\zeta \bm{\breve{\Sigma}}.
    \label{2.19}
\end{equation}
We show the element-wise and operator norm statistical rates in Theorem \ref{theorem5}, Theorem \ref{theorem6}.

\begin{theorem}
Assume (\ref{2.14}) holds, $\delta $ is the same as Theorem \ref{theorem4}, and the truncation threshold $\eta$ and the dithering scale $\gamma$ are set as (\ref{2.16}) with some $C_3,C_4$. Then we set the threshold $\zeta$ in (\ref{2.19}) by 
\begin{equation}
\zeta = C_5 \sqrt{M}\Big(\frac{\delta \log d}{n}\Big)^{1/4}
    \label{2.20}
\end{equation}
where $C_5$ is a sufficiently large constant. Then for any $i,j\in[d]$ we have 
\begin{equation}
\mathbbm{P}\Big(|\widehat{\sigma}_{ij} - \sigma^*_{ij}|\lesssim \min\Big\{|\sigma_{ij}^*|, \sqrt{M}\Big[\frac{\delta \log d}{n}\Big]^{1/4}\Big\}\Big) \geq 1-2d^{-\delta}.
    \label{2.21}
\end{equation}
\label{theorem5}
\end{theorem}
\begin{theorem}
Assume Assumption \ref{assumption1}, (\ref{2.14}) hold, $\delta $ is fixed and the same as Theorem \ref{theorem4}, \ref{theorem5} (set $\delta \geq 4$), the truncation threshold $\eta$, the dithering scale $\gamma$, the threshold $\zeta$ are set as (\ref{2.16}), (\ref{2.20}) for some specified $C_3,C_4,C_5$. Besides, assume that $\delta \log d/n$ is sufficiently small. Let $p=\delta/4$, then we have the bound for the moment of order $p$ 
\begin{equation}
\Big(\mathbbm{E}\|\bm{\widehat{\Sigma}- \Sigma^*}\|_{\mathrm{op}}^p\Big)^{1/p}  \lesssim  sM^{(1-q)/2}\Big[\frac{\delta \log d}{n}\Big]^{(1-q)/4}.
    \label{2.23}
\end{equation}
Moreover, we bound the probability tail of operator norm error 
\begin{equation}
\mathbbm{P} \Big(\bm{\|\widehat{\Sigma}- \Sigma^*}\|_{\mathrm{op}} \lesssim  sM^{(1-q)/2}\Big[\frac{\delta \log d}{n}\Big]^{(1-q)/4} \Big)\geq 1-\exp(-\delta).
    \label{2.24}
\end{equation}
\label{theorem6}
\end{theorem}


\section{Sparse Linear Regression}
\label{section3}
We intend to establish our results for sparse linear regression (Section \ref{section3}) and low-rank matrix completion (Section \ref{section4}) under the unified framework of trace regression, which should be established first. Trace regression with $\bm{\Theta^*}\in \mathbb{R}^{d\times d}$ as desired signal is formulated as 
\begin{equation}
Y_k = \big<\bm{X_k,\Theta^*}\big> + \epsilon_k,
    \label{3.1}
\end{equation}
where $\bm{X_k} \in \mathbb{R}^{d\times d}$ is   covariate, $\epsilon_k$ is  additive noise. To handle high-dimensional scaling, $\bm{\Theta^*}$ is assumed to be (approximately) low-rank (e.g., \cite{negahban2011estimation,negahban2012restricted,fan2021shrinkage}) 
\begin{equation}
\sum_{k=1}^d |\sigma_k(\bm{\Theta^*})|^q \leq r,\ \mathrm{for\ some\ }0\leq q< 1,
    \label{3.2}
\end{equation}
where $\sigma_1(\bm{\Theta^*})\geq \sigma_2(\bm{\Theta^*})\geq ...\geq \sigma_d(\bm{\Theta^*})$ are the singular values of $\bm{\Theta^*}$. For this low-rank trace regression problem, a standard approach to estimate or reconstruct $\bm{\Theta^*}$ is via the M-estimator (e.g.,\cite{negahban2012unified})
\begin{equation}
\bm{\widehat{\Theta}} \in \mathop{\arg\min}\limits_{\bm{\Theta}\in \mathcal{S}} \ \mathcal{L}(\bm{\Theta}) + \lambda \|\bm{\Theta}\|_{\mathrm{nu}},
    \label{3.3}
\end{equation}
where $\mathcal{L}(\bm{\Theta})$ is a loss function that requires $\bm{\widehat{\Theta}}$ to fit the data $\{(\bm{X}_k,Y_k)\}$, $\|\bm{\Theta}\|_{\mathrm{nu}}$ is the penalty that promotes low-rankness. In \cite{negahban2011estimation} Negahban and Wainwright first established a general framework to obtain convergence rate for trace regression when $\mathcal{L}(\bm{\Theta})$ is a quadratic loss, and then many subsequent papers developed and extended the theoretical framework, to name a few, negative log-likelihood loss function \cite{fan2019generalized},  other estimation problems such as matrix completion with sparse corruption \cite{klopp2017robust} and sparse high-dimensional time series \cite{basu2015regularized}, extension to quaternion field \cite{chen2022color}. For data fitting term $\mathcal{L}(\bm{\Theta})$, a standard quadratic loss (i.e., $\ell_2$ loss) based on full data is  
\begin{equation}
\begin{aligned}
   \mathcal{L}(\bm{\Theta})=\frac{1}{2n}\sum_{k=1}^n|Y_k - \left<\bm{X_k,\Theta}\right>|^2 = \frac{1}{2}\mathrm{vec}(\bm{\Theta})^T\bm{\Sigma}_{XX}\mathrm{vec}(\bm{\Theta})-\left<\bm{\Sigma}_{Y\bm{X}},\bm{\Theta}\right> + \mathrm{constant}, 
\end{aligned}
    \nonumber
\end{equation}
where $\bm{\Sigma}_{XX}= \sum_{k=1}^n\mathrm{vec}(\bm{X_k})\mathrm{vec}(\bm{X_k})^T/n$, $\bm{\Sigma}_{Y\bm{X}} = \sum_{k=1}^nY_k\bm{X}_k/n$. However, this standard quadratic loss does not directly apply to our setting where full data are not available. In order to introduce some flexibility, we consider a generalized quadratic loss
\begin{equation}
\mathcal{L}(\bm{\Theta}) =  \frac{1}{2}\mathrm{vec}(\bm{\Theta})^T\bm{Q}\mathrm{vec}(\bm{\Theta})-\left<\bm{B},\bm{\Theta}\right>,
    \label{3.4}
\end{equation}
where $\bm{Q}\in \mathbb{R}^{d^2\times d^2}$ is symmetric, $\bm{B} \in \mathbb{R}^{d\times d}$. We present a framework for trace regression in Lemma \ref{lemma2}. \textcolor{black}{Note that  \cite[Theorem 1]{fan2021shrinkage} is only for $\bm{Q},B$ in (\ref{3.4}) being the (truncated) sample covariance, hence Lemma \ref{lemma2} can be viewed as its extension to more general $\bm{Q},B$ that suffices for our needs}. Besides,  our version is refined to be more technically amenable  since a useful relation (\ref{3.6}) is established even without the restricted strong convexity (\ref{3.7}). One shall see that (\ref{3.6}) can simplify the proofs of Theorems \ref{1bitcssg}, \ref{1bitcsht}, \ref{theorem9}, \ref{theorem10}.

\begin{lem}
Consider trace regression (\ref{3.1}) with (approximate) low-rankness (\ref{3.2}), the estimator is given by (\ref{3.3}) where the loss function is a generalized quadratic loss (\ref{3.4}). Let $\bm{\widehat{\Delta} = \widehat{\Theta}-\Theta^*}$. If $\bm{Q}$ is positive semi-definite, and $\lambda$ satisfies \begin{equation}
\lambda \geq 2\|\mathrm{mat}(\bm{Q}\cdot\mathrm{vec}(\bm{\Theta^*}))-\bm{B}\|_{\mathrm{op}},
    \label{3.5} 
\end{equation} then it holds that  
\begin{equation}
\|\bm{\widehat{\Delta}}\|_{\mathrm{nu}} \leq 10 r^{\frac{1}{2-q}}\|\bm{\widehat{\Delta}}\|_{\mathrm{F}}^{\frac{2-2q}{2-q}}.
    \label{3.6}
\end{equation}
Moreover, if the restricted strong convexity (RSC) holds, i.e., there exists $\kappa >0$ such that 
\begin{equation}
\mathrm{vec}(\bm{\widehat{\Delta}})^T\bm{Q} \mathrm{vec}(\bm{\widehat{\Delta}}) \geq \kappa \|\bm{\widehat{\Delta}}\|_{\mathrm{F}}^2,
    \label{3.7}
\end{equation}
then we have the convergence rate for Frobenius norm and nuclear norm
\begin{equation}
 \|\bm{\widehat{\Delta}}\|_{\mathrm{F}}\leq 30\sqrt{r}\Big(\frac{\lambda}{\kappa}\Big)^{1-q/2} \ \mathrm{and}\  \|\bm{\widehat{\Delta}}\|_{\mathrm{nu}} \leq 300r\Big(\frac{\lambda}{\kappa}\Big)^{1-q}.
 \label{3.8}
\end{equation}
\label{lemma2}
\end{lem}
With the preliminary of trace regression we now go into sparse linear regression \begin{equation}
    Y_k = X_k^T\Theta^* + \epsilon_k,
    \label{3.9}
\end{equation}
where $\Theta^*\in \mathbb{R}^d$ is the desired signal, $X_k$ is the covariate (or sensing vector), $\epsilon_k$ is noise independent of $X_k$. In addition, $\Theta^*$ is approximately sparse.
\begin{assumption}
{\rm (Approximate sparsity on vector)} For a specific $0\leq q<1$, the desired signal $\Theta^* = [\theta^*_1,...,\theta^*_d]^T$ satisfies
\begin{equation}
\sum_{i=1}^d|\theta^*_i|^q \leq s.
    \label{3.10}
    \end{equation}
\label{assumption2}
\end{assumption}
It is not hard to see that (\ref{3.9}), (\ref{3.10}) are encompassed by (\ref{3.1}), (\ref{3.2}) if $\bm{X_k,\Theta^*}$ are diagonal, i.e., $\bm{X_k} = \mathrm{diag}(X_k)$, $\bm{\Theta^*} = \mathrm{diag}(\Theta^*)$, so we consider analogue of (\ref{3.3}) as the estimator. The first issue is the choice of loss function since the existing methods are invalid: we can neither use the quadratic loss as \cite{negahban2011estimation,fan2021shrinkage} without full data, nor the negative log-likelihood as \cite{fan2019generalized} due to the noise $\epsilon_k$ with unknown distribution. Instead, we resort to a generalized quadratic loss given in (\ref{3.4}) to proceed. For sparse linear regression, particularly, we let $\mathcal{L}(\Theta) = \frac{1}{2}\Theta^T\bm{Q}\Theta -B^T \Theta$ where $\bm{Q}\in \mathbb{R}^{d\times d}$ is symmetric, $B\in \mathbb{R}^d$. Thus, our estimator is given by
\begin{equation}
\widehat{\Theta} \in \mathop{\arg\min}\limits_{\Theta \in \mathbb{R}^d} \ \frac{1}{2}\Theta^T \bm{Q} \Theta - B^T\Theta + \lambda \|\Theta\|_{1}.
\label{3.11}
\end{equation}
Lemma \ref{lemma2} implies the following Corollary. 

\begin{cor}
Consider linear regression (\ref{3.9}) with (approximate) sparsity (\ref{3.10}),  the estimator $\widehat{\Theta}$ is given by (\ref{3.11}). Let $\widehat{\Delta}=\widehat{\Theta}-\Theta^*$. If $\bm{Q}$ is positive semi-definite, $\lambda$ satisfies
\begin{equation}
\lambda \geq 2\|\bm{Q}\Theta^* - B \|_{\max},
    \label{3.12}
\end{equation}
then it holds that 
\begin{equation}
    \label{72add_4}
    \|\widehat{\Delta}\|_1 \leq 1 0 s^{\frac{1}{2-q}}\| \widehat{\Delta}\|_2^{\frac{2-2q}{2-q}}.
\end{equation}
Moreover, if for some $\kappa >0$ we have the restricted strong convexity
\begin{equation}
    \widehat{\Delta}^T\bm{Q}\widehat{\Delta} \geq \kappa \|\widehat{\Delta}\|_2^2,
    \label{3.13}
\end{equation}
then we have the error bound for $\ell_2$ and $\ell_1$ norm
\begin{equation}
\|\widehat{\Delta}\|_2 \leq 30\sqrt{s}\Big(\frac{\lambda}{\kappa}\Big)^{1-{q}/{2}}\ ~\mathrm{and}~~ \ \|\widehat{\Delta}\|_1 \leq 300 s\Big(\frac{\lambda}{\kappa}\Big)^{1-q} 
    \label{3.14}
\end{equation}
\label{corollary2}
\end{cor}

It remains to properly specify $\bm{Q},B$ in (\ref{3.11}). Note that the expected quadratic risk is given by $$\mathbbm{E}|Y_k - X_k^T\Theta|^2 = \Theta^T \mathbbm{E}(X_kX_k^T)\Theta -(\mathbbm{E}(Y_kX_k))^T\Theta +\mathrm{constant},$$
 thus a general guideline to choose $\bm{Q}$, $B$ is that $\bm{Q}$ should be close to the covariance matrix of $X_k$, and $B$ should well approximate the covariance $\mathbbm{E}(Y_kX_k)$. Naturally, based on 1-bit data we can still use $\bm{\widehat{\Sigma}}$ in (\ref{2.4}) or $\bm{\breve{\Sigma}}$ in (\ref{2.1}) as $\bm{Q}$. Nevertheless, the issue is that they may not be positive semi-definite, while the positive semi-definiteness of $\bm{Q}$ is an indispensable condition in Corollary \ref{corollary2}. To resolve the issue, we assume  $\bm{\Sigma}_{XX}=\mathbbm{E}X_kX_k^T$ is column-wisely sparse. We defer an illustration of this assumption to Remark \ref{remark3}.
 \begin{assumption}
 $X_1,...,X_n$ are i.i.d. drawn from a zero-mean random vector with covariance matrix $\bm{\Sigma}_{XX} = \mathbbm{E}X_kX_k^T= [\sigma_{ij}] $ satisfying Assumption \ref{assumption1} under parameter $(0,s_0)$, i.e., the number of non-zero elements in each column is less than $s_0$. Besides, $\bm{\Sigma}_{XX}$ is positive definite, and for some absolute constant $\kappa_0  >0$ it satisfies $ \lambda_{\min}(\bm{\Sigma}_{XX}) \geq 2\kappa_0$. 
\label{assumption3}
\end{assumption}
Under Assumption \ref{assumption3}, our estimator $\bm{\widehat{\Sigma}}$ defined in (\ref{2.4}) for sub-Gaussian data, or (\ref{2.19}) for heavy-tailed data, is positive definite with high probability. Thus, we set $\bm{Q} = \bm{\widehat{\Sigma}}$ in (\ref{3.11}). Note that $\mathbbm{E}(Y_kX_k)$ is also covariance, enlightened by Corollary \ref{corollary1}, 
 we similarly set 
\begin{equation}
\widehat{\Sigma}_{YX} = \frac{1}{n}\sum_{k=1}^n  \gamma^2\cdot \dot{Y}_{k}\dot{X}_{k1}.
    \label{3.17}
\end{equation}
 Now we have specified our estimator as
\begin{equation}
\widehat{\Theta} \in \mathop{\arg\min}\limits_{\Theta \in \mathbb{R}^d} \ \frac{1}{2}\Theta^T \bm{\widehat{\Sigma}} \Theta - \widehat{\Sigma}_{YX}^T\Theta + \lambda \|\Theta\|_{1}.
    \label{3.18}
\end{equation}

\subsection{Sub-Gaussian Data}
We assume the sub-Gaussian, zero-mean covariate and sub-Gaussian noise satisfying $\|X_k\|_{\psi_2} \leq \sigma_1$, $\|\epsilon_k\|_{\psi_2}\leq \sigma_2$, and $\|\Theta^*\|_2 \leq R = O(1)$. In this setting, we have $\|Y_k\|_{\psi_2} \leq \|X_k^T\Theta^*\|_{\psi_2}+\|\epsilon_k\|_{\psi_2}\leq \|\Theta^*\|_2 \|X_k\|_{\psi_2}+\|\epsilon_k\|_{\psi_2} = O(\max\{\sigma_1,\sigma_2\})$. To lighten notations without losing generality, we assume for some $\sigma>0$ \begin{equation}
   \max\big\{\|X_k\|_{\psi_2},\|Y_k\|_{\psi_2}\big\}\leq \sigma
    \label{3.19}
\end{equation} 
and use the uniform noise with the same dithering scale $\gamma$ to dither $X_k$ and $Y_k$ before 1-bit quantization. More precisely, we choose dithering noise $\Gamma_{k1},\Gamma_{k2}\sim \mathrm{uni}([-\gamma ,\gamma]^d),\Lambda_k\sim \mathrm{uni}([-\gamma,\gamma])$ with $\gamma$ in (\ref{2.6}), then we obtain the 1-bit data $(\dot{X}_{k1},\dot{X}_{k2},$ $\dot{Y}_k)$.

We mention that our result directly extends to more general setting where $\|X_k\|_{\psi_2}$, $\|Y_k \|_{\psi_2}$  may vary a lot. Indeed, we can adaptively choose dithering scale according to $\|X_k\|_{\psi_2}$ and $\|Y_k \|_{\psi_2}$, for instance, $\Gamma_{k1},\Gamma_{k2}\sim \mathrm{uni}([-\gamma_X,\gamma_X]^d)$, $\Lambda_k \sim \mathrm{uni}([\gamma_Y,\gamma_Y])$. In our numerical simulations, we also applied different dithering scales to $X_k$ and $Y_k$ to improve the recovery.

In   Theorem \ref{theorem7} we will give the near minimax statistical rate for the estimator $\widehat{\Theta}$. The idea is to invoke Corollary \ref{corollary2}, and this requires (\ref{3.12}) and (\ref{3.13}). To properly set $\lambda$ to confirm (\ref{3.12}), it suffices to bound $\|\bm{\widehat{\Sigma}}\Theta^* - \widehat{\Sigma}_{YX}\|_{\max}$ from above. Combining Assumption \ref{assumption3} and results in Section \ref{section2}, we can show (\ref{3.13}) holds with high probability.   
\begin{theorem}
Assume (\ref{3.9}), Assumption \ref{assumption2}, (\ref{3.19}) hold, $\|\Theta^*\|_{2}\leq R $ for some absolute constant $R$, and  Assumption \ref{assumption3} holds for some fixed integer $s_0$. Before the quantization we dither the data with $\gamma$ in (\ref{2.6}). We consider $\widehat{\Theta}$ given by (\ref{3.18}) where $\bm{\widehat{\Sigma}},\widehat{\Sigma}_{YX}$ are respectively set as (\ref{2.4}), (\ref{3.17}), and $\zeta$ is given by (\ref{2.9}). Moreover, we choose $\lambda$ by 
\begin{equation}
\lambda = C_6\log n \sqrt{\frac{\delta \log d}{n}}
    \label{3.20}
\end{equation}
with sufficiently large $C_6$. Let $\widehat{\Delta}=\widehat{\Theta}- \Theta^*$. When $\frac{(\log n)^2\log d }{n}$ is sufficiently small, with probability at least $1-\exp(-\delta)- 2d^{2-\delta}$, we have 
\begin{equation}
    \begin{cases}
 \displaystyle     \|\widehat{\Delta}\|_2\lesssim \sqrt{s}\Big(\sigma^2 \log n \sqrt{\frac{\delta \log d}{n}}\Big)^{1-q/2}\\
    \displaystyle  \|\widehat{\Delta}\|_1\lesssim ~s\Big(\sigma^2 \log n \sqrt{\frac{\delta \log d}{n}}\Big)^{1-q}
        \label{3.21}
    \end{cases}.
\end{equation}
\label{theorem7}
\end{theorem}
\begin{rem}
Compared to the sample covariance $\sum_{k=1}^n{X_k^TX_k}/{n}$, the proposed 1-bit covariance matrix estimator $\bm{\widehat{\Sigma}}$ lacks positive semi-definiteness. We address the issue by assuming column-wise sparsity of $\bm{\Sigma}_{XX}$, which together with $\lambda_{\min}(\bm{\Sigma}_{XX})=\Omega(1)$ can provide positive definiteness under high-dimensional scaling. This assumption is also used in \cite{yang2015closed} to resolve the same issue. As an example, this accommodates isotropic sensing vectors that is conventionally adopted in compressed sensing literature \cite{candes2011probabilistic,plan2017high,dirksen2021non}. In addition, we have removed this (a bit uncommon) assumption in our subsequent work \cite{chen2022quantizing}. 
\label{remark3}
\end{rem}

\subsection{Heavy-tailed Data}
We then switch to the heavy-tailed case where $X_k$ and $\epsilon_k$ are only assumed to possess bounded $4$-th moment. We consider the scaling of the desired signal as $\|\Theta^*\|_2 \leq R = O(1)$. Moreover, we assume $\mathbbm{E}|V^TX_k|^4 \leq M_1$ for any $V\in\mathbb{R}^d, \|V\|_2 \leq  1$, and $\mathbbm{E}|\epsilon_k|^4\leq M_2$. Then we have the fourth moment of $Y_k$ is also bounded by $O\big(R^4M_1 + M_2\big)$.
To lighten the notation without losing generality, we assume the same upper bound $M$ for covariate and response: \begin{equation}
\max\big\{\sup_{\|V\|_2\leq 1}\mathbbm{E}|V^TX_k|^4,\mathbbm{E}|Y_k|^4 \big\} \leq M,
    \label{3.22}
\end{equation} which allows us to use the same truncation parameter $\eta$ and dithering scale $\gamma$ for $X_k$ and $Y_k$.

Similar to the same comment for sub-Gaussian case, if the fourth moment of $X_k$ and $Y_k$ have different scales, our method still works under different truncation parameters $(\eta_X,\eta_Y)$ and dithering parameters $( \gamma_X,\gamma_Y)$. Moreover, it is straightforward to adapt our method to the mixing case studied in \cite{fan2021shrinkage} where $X_k$ is sub-Gaussian but $\epsilon_k$ (and hence $Y_k$) is heavy-tailed. In this mixing setting, only the responses are treated as heavy-tailed data and truncated before dithering.
\begin{theorem}
Assume (\ref{3.9}), Assumption \ref{assumption2}, (\ref{3.22}) hold, $\|\Theta^*\|_{2}\leq R$ for some absolute constant $R$, and   Assumption \ref{assumption3} holds for some fixed integer $s_0$. By setting $\eta,\gamma$ as (\ref{2.16}) such that $\gamma > \eta$, we first truncate $(X_k,Y_k)$ element-wisely to $(\widetilde{X}_k,\widetilde{Y}_k)$ with parameter $\eta$, then dither the truncated data with uniform noise on $[-\gamma,\gamma]$ and quantize the data to $(\dot{X}_{k1},\dot{X}_{k2},\dot{Y}_k)$ finally. We consider $\widehat{\Theta}$ in (\ref{3.18}) where $\bm{\widehat{\Sigma}}$ is given in (\ref{2.19}) with $\zeta$ set as (\ref{2.20}), $\widehat{\Sigma}_{YX}$ is given in (\ref{3.17}). Moreover, we choose 
\begin{equation}
    \lambda  =  C_7\sqrt{M}\Big(\frac{\delta \log d}{n}\Big)^{{1}/{4}}
    \label{3.23}
\end{equation}
with sufficiently large $C_7$. Let $\widehat{\Delta}= \widehat{\Theta}-\Theta^*$. When ${\log d}/{n}$ is sufficiently small, with probability at least $1-\exp(-\delta)-2d^{2-d}$, we have 
\begin{equation}
\begin{cases}
 \displaystyle  \|\widehat{\Delta}\|_2\lesssim \sqrt{s}M^{1/2-q/4}\Big(\frac{\delta \log d}{n}\Big)^{1/4-q/8}\\
   \displaystyle  \|\widehat{\Delta}\|_1 \lesssim ~s M^{(1-q)/2}\Big(\frac{\delta \log d}{n}\Big)^{(1-q)/4}
\end{cases}.
\label{3.24}
\end{equation}
\label{theorem8}
\end{theorem}
We emphasize that our method does not rely on the full knowledge of  $\bm{\Sigma}_{XX}$; indeed, our method applies as long as $\bm{\Sigma}_{XX}$ satisfies Assumption \ref{assumption3}. Note that when $\bm{\Sigma}_{XX}$ is known as a priori,  we can  directly set $\bm{Q} = \bm{\Sigma}_{XX}$ in (\ref{3.11}), and the same error rates can be obtained by the similar techniques. 

\subsection{1-bit Compressed Sensing}
\textcolor{black}{
We just studied  sparse linear regression based on the 1-bit quantized covariates and responses $(\dot{X}_{k1},\dot{X}_{k2},\dot{Y}_k)$, while the only related problem studied in existing works is 1-bit compressed sensing (1-bit CS). In 1-bit CS,   one considers the same linear model (\ref{3.9}) and wants to  estimate the sparse underlying signal $\Theta^*$ based on $(X_k,\dot{Y}_k)$, where $X_k$ denotes the full covariate, and $\dot{Y}_k\in \{-1, 1\}$ is the 1-bit quantized version of   the response $Y_k$. In particular, earlier works mainly studied a direct quantization with $\dot{Y}_k = \mathrm{sign}(X^T\Theta^*)$ (see, e.g., \cite{boufounos20081,jacques2013robust,plan2012robust,plan2013one}), while recent works (e.g., \cite{knudson2016one,baraniuk2017exponential,dirksen2021non,dirksen2018robust,thrampoulidis2020generalized}) began to consider  dithered quantization that are more relevant to our work, i.e., $\dot{Y}_k = \mathrm{sign}(X^T\Theta^* + \Lambda)$ for some dithering noise $\Lambda$. Specifically, by the additional dithering step, these works overcome several limitations and present better results. For instance, full reconstruction with norm \cite{knudson2016one}, exponentially-decaying error rate  \cite{baraniuk2017exponential}, and extension to non-Gaussian sensing vectors \cite{dirksen2021non,dirksen2018robust,thrampoulidis2020generalized}.}

Since one still has full knowledge on $X_k$ in 1-bit CS, our problem setting is novel and evidently more tricky. From a practical viewpoint, due to the binary covariate, the storage and communication costs are further lowered in our method. Technically, the key element that allows quantization of covariate is the new 1-bit sparse covariance matrix estimator developed in Section \ref{section2}. To facilitate presentation and future study, we term this new setting as {\it  1-bit quantized-covariate compressed sensing (1-bit QC-CS)} to distinguish with the canonical 1-bit CS.

Note that it is unfair to compare our Theorem \ref{theorem7}, \ref{theorem8} with existing results for 1-bit CS.  \textcolor{black}{To see the contributions of this paper more explicitly, we analogously establish results for 1-bit CS where full-precision $X_k$ are available, under both sub-Gaussian and heavy-tailed regimes. Similar to (\ref{3.18}), we formulate the estimation as a convex programming problem, but  substitute $\bm{\widehat{\Sigma}}$ in (\ref{3.18}) with the sample covariance matrix $\bm{\widehat{\Sigma}}_{XX}= \sum_{k=1}^n X_kX_k^T/n$ for sub-Gaussian $X_k$, or the truncated sample covariance matrix $\bm{\widehat{\Sigma}}_{\tilde{X}\tilde{X}} = \sum_{k=1}^n \widetilde{X}_k\widetilde{X}_k^T / n$ for heavy-tailed $X_k$. Here, for heavy-tailed case, we truncate $X_k$ element-wisely, but we distinguish the truncation threshold of $X_k$, $Y_k$ by different notations $\eta_X$, $\eta_Y$ and they are set to be different values. More precisely, the $i$-th entry of $\widetilde{X}_k$ is given by $\widetilde{X}_{k,i} = \sign(X_{k,i})\min\{|X_{k,i}|,\eta_X\}$, while before the dithered quantization, $Y_k$ is truncated to be $\widetilde{Y}_k = \sign(Y_k)\min\{| Y_k|,\eta_Y\}$.}

 Although the results are similarly established by the framework of trace regression,
we   feel obliged to note some differences. Let us consider the sub-Gaussian regime for illustration. Firstly, the column-wise sparsity of $\bm{\Sigma}_{XX}$ in Assumption \ref{assumption3}, whose main aim is to guarantee positive semi-definiteness of the 1-bit covariance matrix estimator $\bm{\widehat{\Sigma}}$, can now be removed as $\bm{\widehat{\Sigma}}_{XX}$  is automatically positive semi-definite. But on the other hand, without this assumption, we no longer have a dimension-free upper bound on   $\|\bm{\widehat{\Sigma}}_{XX}- \bm{\Sigma}_{XX}\|_{\mathrm{op}}$, hence the proof cannot proceed to (\ref{add72_1}). Indeed, we only have dimension-free upper bound on $\|\bm{\widehat{\Sigma}}_{XX}- \bm{\Sigma}_{XX}\|_{\max}$. In heavy-tailed case we hence impose a stronger scaling $\| \Theta^*\|_1 \leq R$, (which is also used in the heavy-tailed case of sparse linear regression in \cite[Lemma 1(b)]{fan2021shrinkage}). In addition, we need to establish the restricted strong convexity (\ref{3.7}) in Lemma \ref{lemma2} via some additional technicalities.

In the next two theorems we present our results on 1-bit CS, which are directly comparable to the prior results of 1-bit CS. To facilitate the flow of our presentation, a detailed comparison is postponed to Appendix \ref{appendxD}. One shall see that,  the following two results improve on existing ones from some respect. 
 
\begin{theorem}
\label{1bitcssg}
  {\rm (1-bit CS with sub-Gaussian data){\bf \sffamily.}}  Assume (\ref{3.9}), Assumption \ref{assumption2} hold, $\|X_k\|_{\psi_2}\leq \sigma_1$, $\|\epsilon_k\|_{\psi_2}\leq \sigma_2$, $\|\Theta^*\|_2\leq R $ with absolute constants $\sigma_1$, $\sigma_2$, $R$. For the zero-mean covariate $X_k$, define the covariance matrix $\bm{\Sigma}_{XX} = \mathbbm{E} X_kX_k^T$ and we assume  $\lambda_{\min}(\bm{\Sigma}_{XX})\geq 2\kappa_0 $ for some absolute constant $\kappa_0>0$. We quantize $Y_k$ to be $\dot{Y}_k = \sign(Y_k + \Lambda_k)$ with $\Lambda_k$ uniformly distributed on $[-\gamma,\gamma]$, and   we set $\gamma = C_{8}'\sqrt{  \log n}$ for sufficiently large $C_8'$. The estimation is formulated as a convex programming problem  \begin{equation}
  \label{add72_5}
\widehat{\Theta} \in \mathop{\arg\min}\limits_{\Theta\in\mathbb{R}^d} \ \frac{1}{2}\Theta^T \bm{\widehat{\Sigma}}_{XX} \Theta - \widehat{\Sigma}_{YX}^T\Theta + \lambda \|\Theta\|_{1}.
 \end{equation}
Moreover, we set $\bm{\widehat{\Sigma}}_{XX}$, $\widehat{\Sigma}_{YX}^T$ and $\lambda$ in (\ref{add72_5}) as 
 \begin{equation}
     \label{add72_6}
      \bm{\widehat{\Sigma}}_{XX}=\frac{1}{n}\sum_{k=1}^n X_kX_k^T, ~ ~  \widehat{\Sigma}_{YX}=  \frac{1}{n}\sum_{k=1}^n \gamma\cdot\dot{Y}_K X_k,~~\lambda =C_8  \sqrt{\frac{\delta \log d \log n}{n}}~~
 \end{equation}
 with some sufficiently large $C_8$. Let $\widehat{\Delta}=\widehat{\Theta}-\Theta^*$, then when $s\big(\frac{\delta \log d}{n}\big)^{1-q/2}$ is sufficiently small, with probability at least $1-7d^{2-\delta}$, we have 
 \begin{equation}
     \label{add72_7}
     \begin{cases}
       \displaystyle \|\widehat{\Delta}\|_2 \lesssim \sqrt{s}\Big( \sqrt{\frac{\delta \log d\log n}{n}}\Big)^{1-q/2} \\
     \displaystyle   \|\widehat{\Delta}\|_1 \lesssim~ s\Big( \sqrt{\frac{\delta \log d\log n}{n}}\Big)^{1-q}
     \end{cases}. 
 \end{equation}
 \end{theorem}

By taking advantage of the full covariate, in heavy-tailed regime of 1-bit CS we show the $\ell_2$ norm error rate $O\big(s^{2/3}\big(\frac{\log d}{n}\big)^{1/3-q/6}\big)$, which is faster than the corresponding rate for 1-bit QC-CS in Theorem \ref{theorem8}. 

\textcolor{black}{\begin{theorem}
\label{1bitcsht}
{\rm (1-bit CS with heavy-tailed data){\bf \sffamily.}} Assume (\ref{3.9}), Assumption \ref{assumption2}, (\ref{3.22}) hold ($M$ in (\ref{3.22}) is an absolute constant), $\|\Theta^*\|_1 \leq R  $ for some absolute constant $R$. For the zero-mean covariate $X_k$ we let $\bm{\Sigma}_{XX}=\mathbbm{E}X_kX_k^T$ and assume $\lambda_{\min}(\bm{\Sigma}_{XX})\geq 2\kappa_0 $ for some absolute constant $\kappa_0>0$. We element-wisely truncate $X_k$ to be $\widetilde{X}_k$ with threshold $\eta_X $, and truncate $Y_k$ to be $\widetilde{Y}_k$ with threshold $\eta_Y$. Then, $\widetilde{Y}_k$ is dithered and quantized to be $\dot{Y}_k = \sign(\widetilde{Y}_k+\Lambda_k)$ with $\Lambda_k$ uniformly distributed on $[-\gamma,\gamma]$. For specific $\delta >0$, we set these signal processing parameters as 
\begin{equation}
    \label{signalpa}
    \eta_X = C_{9} \Big(\frac{n}{\delta \log d}\Big)^{\frac{1}{4}},~~ \eta_Y = C_{10} \Big(\frac{n}{\delta \log d}\Big)^{\frac{1}{6}},~~ \gamma = C_{11} \Big(\frac{n}{\delta \log d}\Big)^{\frac{1}{6}},
\end{equation}
where $C_{11}>C_{10}$ to give $\gamma >\eta_Y$. The estimation is formulated as a convex programming problem 
\begin{equation}
  \label{add72_8}
\widehat{\Theta} \in \mathop{\arg\min}\limits_{\Theta\in\mathbb{R}^d}  \ \frac{1}{2}\Theta^T \bm{\widehat{\Sigma}}_{\tilde{X}\tilde{X}} \Theta - \widehat{\Sigma}_{YX}^T\Theta + \lambda \|\Theta\|_{1}.
 \end{equation}
 Moreover, we set $\bm{\widehat{\Sigma}}_{\tilde{X}\tilde{X}} $, $\widehat{\Sigma}_{YX}$ and $\lambda$ in (\ref{add72_8}) as
 \begin{equation}
     \label{add72_9}
      \bm{\widehat{\Sigma}}_{\tilde{X}\tilde{X}}=\frac{1}{n}\sum_{k=1}^n \widetilde{X}_k\widetilde{X}_k^T, ~~   \widehat{\Sigma}_{YX}=  \frac{1}{n}\sum_{k=1}^n \gamma\cdot\dot{Y}_k  \widetilde{X}_k,~~\lambda =C_{12} \Big(\frac{\delta \log d  }{n}\Big)^{1/3}~~
 \end{equation}
  with some sufficiently large $C_{12}$. Let $\widehat{\Delta}=\widehat{\Theta}-\Theta^*$.  Under sufficiently small  $s^2\big({\frac{\delta \log d}{n}}\big)^{1-q/2}$, and we further assume  $s\big(\frac{\delta \log d}{n}\big)^{\frac{1}{2}-\frac{q}{3}}=O(1)$ for $q\in (0,1)$, then with probability at least $1-O(d^{2-\sqrt{\delta}})$, we have 
  \begin{equation}
      \label{add72_10}
      \begin{cases}
      \displaystyle   \|\widehat{\Delta}\|_2 \lesssim \sqrt{s}\Big(\frac{\delta \log d}{n}\Big)^{\frac{1-q/2}{3}}\\
     \displaystyle    \|\widehat{\Delta}\|_1 \lesssim s\Big(\frac{\delta \log d}{n}\Big)^{\frac{1-q}{3}}
      \end{cases}.
  \end{equation}
\end{theorem}
}

\vspace{2mm}
We conclude this section by deriving an information-theoretic lower bound regarding Theorem \ref{1bitcsht}. To be concise we only deal with   exactly sparse $\Theta^*$, i.e., $s$-sparse $\Theta^*$. As Theorem \ref{1bitcsht} we consider the set of parameters \begin{equation}
    \mathscr{K}(s,R)= \big\{\Theta\in \mathbb{R}^d:\|\Theta\|_0\leq s,\|\Theta\|_1\leq R\big\}.
\end{equation}  
\begin{theorem}\label{lower2}
    Given $n,d,s,R$ and covariates $\{X_k:k\in [n]\}$, and assume $s\leq \frac{d}{8}$. For some underlying $\Theta \in \mathscr{K}(s,R)$, we suppose the responses $(\dot{Y}_k)_{k=1}^n$ are obtained as in Theorem \ref{1bitcsht}, i.e., truncation, dithering and quantization with parameters in (\ref{add72_9}), and we assume $\eta_Y <\frac{8}{9}\gamma$. Consider any algorithm which, for any underlying $\Theta\in \mathscr{K}(s,R)$, takes $\{X_k:k\in [n]\}$ and the corresponding $(\dot{Y}_k)_{k=1}^n$ as input and returns $\widehat{\Theta}$. If $  n\gtrsim K_u^{-1}\big(\frac{s}{R}\big)^3\log\frac{d}{2s}$, and $\sum_{k=1}^n |X_k^T V|^2 \leq nK_u \|V\|^2_2$ holds for some $K_u$ and for all $2s$-sparse $V\in \mathbb{R}^d$, then there exists $\Theta_0\in \mathscr{K}(s,R)$ such that with probability at least $\frac{3}{4}$, \begin{equation}
        \|\widehat{\Theta}-\Theta_0\|_2\geq \Big(\frac{\log \frac{d}{2s}}{\log d}\Big)^{\frac{1}{6}}\sqrt{\frac{s}{K_u}}\Big(\frac{\log \frac{d}{2s}}{n}\Big)^{1/3}.
    \end{equation} 
\end{theorem}
The proof follows  similar courses   as in \cite[Theorem 1]{raskutti2011minimax}, while the main difference is on bounding the Kullback–Leibler divergence because of the 1-bit observation $\dot{Y}_k$. The conditions are quite benign: $\sum_{k=1}^n|X_k^TV|^2\leq nK_u\|V\|_2^2$ for $2s$-sparse $V$ is satisfied by a large class of random $X_k$ with bounded or logarithmic $K_u$ (e.g., i.i.d., sub-Gaussian or sub-exponential $X_k$);   $n\geq K_u^{-1}\big(\frac{s}{R}\big)^3\log\frac{d}{2s}$ is in the interesting high-dimensional scaling and assumed to guarantee the packing set is a subset of $\mathscr{K}(s,R)$. Note that the lower bound matches our upper bound for $\ell_2$ error in Theorem \ref{1bitcsht}, up to logarithmic factors and the parameter $K_u$. Thus, for estimation of $\Theta^*$ from the observed data $(X_k,\dot{Y}_k)$, the rate in Theorem \ref{1bitcsht} is almost tight, and significantly faster rate is not achievable without changing the process of producing $\dot{Y}_k$.

In the proof, the key point that lifts the regular lower bound $\tilde{\Omega}(\frac{\sqrt{s}}{\sqrt{n}})$ to $\tilde{\Omega}(\frac{\sqrt{s}}{n^{1/3}})$ 
is our choice of dithering scale (i.e., $\gamma \asymp \big(\frac{n}{\log d}\big)^{1/6}$). Note that it is essentially larger than $\gamma \asymp \sqrt{\log n}$ used in the sub-Gaussian case (Theorem \ref{1bitcssg}). Such a larger dithering scale can be understood as the price we pay for dealing with heavy-tailed noise.

\section{Low-rank Matrix Completion}
\label{section4}
Matrix completion refers to the problem of recovering a low-rank matrix with incomplete observations of the entries, which is motivated by recommendation system, system identification, quantum state tomography, image inpainting, and many others, see \cite{davenport2016overview,chen2018harnessing,bennett2007netflix,gross2010quantum,fazel2003log,chen2022color,chen2019low} for instance. The literature can be roughly divided into two lines, exact recovery and approximate recovery (i.e., statistical estimation). To establish exact recovery guarantee, the underlying matrix is required to satisfy a quite stringent incoherence condition proposed and developed in \cite{candes2009exact,candes2010matrix,candes2010power,recht2011simpler}. By contrast, it was shown that matrix with low spikiness could be well approximated (or estimated) under much more relaxed condition \cite{negahban2012restricted,klopp2014noisy,koltchinskii2011nuclear}. This Section is intended to study the estimation problem of matrix completion via the binary data produced by our 1-bit quantization scheme. For simplicity we consider square matrix and formulate the model as \begin{equation}
Y_k = \left<\bm{X_k,\Theta^*}\right> +\epsilon_k
    \label{4.1}
\end{equation} 
where $\bm{\Theta^*}$ is the underlying low-rank data matrix of interest, $\bm{X_k}$ distributed on $\{e_ie_j^T:i,j\in[d]\}$ is the sampler that extracts one entry of $\bm{\Theta^*}$, $Y_k$ is the $k$-th observation corrupted by noise $\epsilon_k$ independent of $\bm{X_k}$. We consider a random, uniform sampling scheme \begin{equation}
    \bm{X_1},...,\bm{X_n}\ \text{are i.i.d. uniformly distributed on } \{e_ie_j^T:i\in[d],j\in[d]\},
    \label{4.2}
\end{equation}     
but we mention that the results can be directly adapted to non-uniform sampling scheme, see \cite{klopp2014noisy}. To embrace more real applications, $\bm{\Theta^*}$ is assumed to be approximately low-rank \cite{negahban2012restricted}. 
\begin{assumption}
{\rm (Approximate low-rankness on matrix){\bf \sffamily.}} Let $\sigma_1(\bm{\Theta^*})\geq ...\geq \sigma_d(\bm{\Theta^*})$ be singular values of $\bm{\Theta^*}$, $0\leq q<1$. For some $r>0$ it holds that 
\begin{equation}
\sum_{k=1}^d\sigma_k(\bm{\Theta^*})^q \leq r.
    \label{4.3}
\end{equation}
\label{assumption4}
\end{assumption}
Since $\bm{X_k}$ only has $d^2$ values, we can use $\lceil2\log_2 d \rceil $ bits to encode $\bm{X_k}$ without losing any information. Therefore, we only quantize $Y_k$ to binary data $\dot{Y}_k$ and study the estimation via $(\bm{X_k},\dot{Y}_k)$. Similar to our prior developments, we use a generalized quadratic loss (\ref{3.4}) with $\bm{Q},\bm{B}$ specified to be
 \begin{equation}
    \bm{Q}= \frac{1}{n}\sum_{k=1}^n \mathrm{vec}(\bm{X_k})\mathrm{vec}(\bm{X_k})^T ~,~~\bm{B}  = \frac{1}{n}\sum_{k=1}^n\gamma \cdot \dot{Y}_k \bm{X_k}. 
    \label{4.4}
\end{equation} 
The spikiness of $\bm{\Theta^*}$ is defined as $ \frac{d\|\bm{\Theta^*}\|_{\max}}{\|\bm{\Theta^*}\|_{\mathrm{F}}} $ in \cite{negahban2012restricted}, and note that completing a matrix with high spikiness (close to $d$) is an ill-posed problem per se \cite{davenport2016overview}. Besides the spikiness, a similar but more straightforward assumption is   a max-norm constraint (e.g., \cite{klopp2014noisy,chen2022color,davenport20141,cai2013max}). 
Here, we adopt this more straightforward condition and assume \begin{equation}
    \|\bm{\Theta^*}\|_{\max}\leq \alpha^*.
    \label{4.5}
\end{equation}
Substitute (\ref{4.4}), (\ref{3.4}) into (\ref{3.3}), together with the max-norm constraint (\ref{4.5}), we now define our estimator via the following convex programming problem
\begin{equation}
\begin{aligned}
\bm{\widehat{\Theta}} \in& \mathop{\arg\min}\limits_{\|\bm{\Theta}\|_{\max}\leq {\alpha^*}} \ \frac{1}{2}\mathrm{vec}(\bm{\Theta})^T\bm{Q}\mathrm{vec}(\bm{\Theta})-\left<\bm{B},\bm{\Theta}\right> + \lambda \|\bm{\Theta}\|_{\mathrm{nu}}\\
=&\mathop{\arg\min}\limits_{\|\bm{\Theta}\|_{\max}\leq {\alpha^*}} \ \frac{1}{2n}\sum_{k=1}^n \big(\big<\bm{X_k},\bm{\Theta}\big>-\gamma\cdot \dot{Y}_k\big)^2 + \lambda \|\bm{\Theta}\|_{\mathrm{nu}}
\end{aligned}
    \label{4.6}
\end{equation}
Compared with the program (\ref{3.18}) involving $\bm{\widehat{\Sigma}}$ used in sparse linear regression, (\ref{4.6}) is more intuitive since we simply replace the full observation $Y_k$ in a standard quadratic loss with its 1-bit surrogate $\gamma \cdot \dot{Y}_k$. Such choice can be readily explained by  Lemma \ref{lemma1}.

Applying Lemma \ref{lemma2} to the problem set-up of low-rank matrix completion directly gives the following Corollary \ref{corollary3}.

\begin{cor}
Consider (\ref{4.1}) under random sampling (\ref{4.2}), $\bm{\Theta^*}$ satisfies Assumption \ref{assumption4} and (\ref{4.5}). Consider $\bm{\widehat{\Theta}}$ in (\ref{4.6}). Let $\bm{\widehat{\Delta}=\widehat{\Theta}-\Theta^*}$. If 
\begin{equation}
\lambda \geq 2\big\|\frac{1}{n}\sum_{k=1}^n\big[\left<\bm{X_k,\Theta^*}\right>-\gamma \cdot\dot{Y}_k\big]\bm{X_k}\big\|_{\mathrm{op}},
    \label{4.7}
\end{equation}
then it holds that \begin{equation}
\|\bm{\widehat{\Delta}}\|_{\mathrm{nu}} \leq 10 r^{\frac{1}{2-q}}\|\bm{\widehat{\Delta}}\|_{\mathrm{F}}^{\frac{2-2q}{2-q}}.
    \label{4.8}
\end{equation} Moreover, if  the RSC holds, i.e., for some $\kappa >0$  \begin{equation}
\frac{1}{n}\sum_{k=1}^n |\big<\bm{X_k,\widehat{\Delta}}\big>|^2 \geq \kappa \|\bm{\widehat{\Delta}}\|_{\mathrm{F}}^2,  
    \label{4.9}
\end{equation}
then we have 
\begin{equation}
 \|\bm{\widehat{\Delta}}\|_{\mathrm{F}}\leq 30\sqrt{r}\Big(\frac{\lambda}{\kappa}\Big)^{1-{q}/{2}} \ \mathrm{and}\ ~ \|\bm{\widehat{\Delta}}\|_{\mathrm{nu}} \leq 300r\Big(\frac{\lambda}{\kappa}\Big)^{1-q}.
 \label{4.10}
\end{equation}
\label{corollary3}
\end{cor}
\subsection{Sub-Gaussian noise}
We first consider sub-Gaussian noise $\epsilon_k $ satisfying
\begin{equation}
    \label{75add_2}
   \mathbbm{E}\epsilon_k=0,~ \|\epsilon_k\|_{\psi_2} \leq \sigma.
\end{equation}
To invoke Corollary \ref{corollary3} and obtain the statistical rate, we need to choose suitable $\lambda$ that guarantees (\ref{4.7}) with high probability. Thus, we upper bound the right hand side of (\ref{4.7}) first. 
\begin{lem}
Consider (\ref{4.1}) under sampling scheme (\ref{4.2}), max-norm constraint (\ref{4.5}), and sub-Gaussian noise assumption (\ref{75add_2}). For a specific $\delta >1$, we choose the dithering noise scale $\gamma$ by 
\begin{equation}
    \gamma = C_{13} \max\{\alpha^*,\sigma\}\sqrt{{\log \Big(\frac{n}{\delta d \log (2d)}\Big)} }
    \label{4.11}
\end{equation}
with some sufficiently large $C_{13}$ such that $\gamma \geq 2\max\{\alpha^*, \sigma\}$. If $\frac{\delta d \log d}{n}$ is sufficiently small, we have 
\begin{equation}
\big\|\frac{1}{n}\sum_{k=1}^n\big[\left<\bm{X_k,\Theta^*}\right>-\gamma\cdot \dot{Y}_k\big]\bm{X_k}\big\|_{\mathrm{op}} \lesssim \max\{\alpha^*,\sigma\} \sqrt{\log n \frac{\delta   \log d}{nd}}
    \label{4.12}
\end{equation}
with probability higher than $1-2d^{1-\delta}$. 
\label{lemma3}
\end{lem}
It remains to consider (\ref{4.9}). To lighten the notation we use $\mathscr{X}=(\bm{X_1},...,\bm{X_n})$
to denote the observed positions and define $\mathcal{F}_{\mathscr{X}}(\bm{\Theta}) = n^{-1}\sum_{k=1}^n|\big< \bm{X_k},\bm{\Theta}\big>|^2$. It is known that $\mathcal{F}_{\mathscr{X}}(\bm{\Theta})\geq \kappa\| \bm{\widehat{\Delta}}\|^2_{\mathrm{F}}$ may not always hold under high-dimensional scaling and the special covariate (\ref{4.2}). In this case, one often needs to establish (\ref{4.9}) with a relaxed (tolerance) term   \cite[Definition 2]{negahban2012unified}. To this end, Negahban and Wainwright first established such relaxed RSC over a constraint set in \cite[Theorem 1]{negahban2012restricted}. Later, in   \cite[Lemma 12]{klopp2014noisy}, Klopp considered a different constraint set and provided a refined proof, but only for the exact low-rank setting, i.e., $q=0$ in Assumption \ref{assumption4}. More recently, in  \cite[Lemma 5]{chen2022color}, Chen and Ng considered a constraint set depending on $q\in [0,1)$ and extended the   proof in \cite{klopp2014noisy} to approximate low-rank regime. As a consequence, a simpler and much shorter proof for the error bound in \cite{negahban2012restricted} could be obtained, see more discussions in \cite{chen2022color}. Here we show the relaxed RSC over the constraint set defined in \cite{chen2022color}, see $\mathcal{C}(\psi)$ in (\ref{4.13}). 
\begin{lem}
For a specific $\delta$ and sufficiently large $\psi$, we consider the constraint set
\begin{equation}
\begin{aligned}
\mathcal{C}(\psi) = \big\{\bm{\Theta}\in \mathbb{R}^{d\times d}:&\|\bm{\Theta}\|_{\max}\leq 2\alpha^*,\|\bm{\Theta}\|_{\mathrm{nu}}\leq 10r^{\frac{1}{2-q}}\|\bm{\Theta}\|_{\mathrm{F}}^{\frac{2-2q}{2-q}},\\
&\|\bm{\Theta}\|_{\mathrm{F}}^2\geq (\alpha^*d)^2\sqrt{\frac{\psi\delta\log(2d)}{n}} \big\}.
\end{aligned}
    \label{4.13}
\end{equation}
Then there exists some absolute constant $\kappa\in (0,1)$, such that with probability at least $1-d^{-\delta}$, it holds that 
\begin{equation}
\mathcal{F}_{\mathscr{X}}(\bm{\Theta})\geq \kappa d^{-2}\|\bm{\Theta}\|_{\mathrm{F}}^2 - T_0,\ \forall \bm{\Theta} \in \mathcal{C}(\psi),
    \label{4.14}
\end{equation}
where the relaxation term $T_0$ is given by 
\begin{equation}
T_0 = \frac{r}{(2-q)d^q}\Big(240\alpha^*\sqrt{\frac{d\log(2d)}{n}}\Big)^{2-q}.
    \label{4.15}
\end{equation}
\label{lemma31}
\end{lem}
We are now ready to derive the statistical bound of the estimation error $\bm{\widehat{\Delta}} = \bm{\widehat{\Theta}}-\bm{\Theta^*}$. The main idea is parallel to previous works \cite{negahban2012restricted,klopp2014noisy,chen2022color}, i.e., to discuss whether $\bm{\widehat{\Delta}}$ belongs to $\mathcal{C}(\psi)$. Note that this only hinges on the third constraint in (\ref{4.13}), since the first two constraints are automatically satisfied by $\bm{\widehat{\Delta}}$, see (\ref{4.5}) and (\ref{4.8}).

\begin{theorem}
Under the setting of Lemma \ref{lemma3}, assume $\bm{\Theta^*}$ satisfies Assumption \ref{assumption4}, we consider the estimator $\bm{\widehat{\Theta}} $ defined in (\ref{4.6}). Moreover, we set $\lambda$ by 
\begin{equation}
\lambda  = C_{14} \max\{\alpha^*,\sigma\}\sqrt{\log n \frac{\delta   \log d}{nd}}
    \label{4.16}
\end{equation}
with sufficiently large $C_{14}$, assume $\frac{\delta d \log d}{n}$ is sufficiently small, $r\gtrsim d^{q}$, $n\lesssim d^2\log(2d)$, then with probability higher than $1-3d^{1-\delta} $, we have 
\begin{equation}
    \begin{cases}
\displaystyle     \|\bm{\widehat{\Delta}}\|_{\mathrm{F}}^2 /d^2\lesssim rd^{-q}\Big(\max\{(\alpha^*)^2,\sigma^2\}\log n \frac{\delta  d \log d}{n}\Big)^{1-{q}/{2}}\\
    \displaystyle  \|\bm{\widehat{\Delta}}\|_{\mathrm{nu}}/d \lesssim rd^{-q}
      \Big(\max\{\alpha^*,\sigma\}\sqrt{\log n \frac{\delta  d \log d}{n}}\Big)^{1-q}.
     \label{5.17}
    \end{cases}
\end{equation}
\label{theorem9}
\end{theorem}

\begin{rem}
Under a specific scaling $\|\bm{\Theta^*}\|_{\mathrm{F}}=1$, $\bm{X_k}= d\cdot e_ie_j^T$ adopted in \cite{negahban2012restricted,fan2021shrinkage}, our bound for the mean square error $d^{-2}\|\bm{\widehat{\Delta}}\|_{\mathrm{F}}^2$ is equivalent to $$\|\bm{\widehat{\Delta}}\|_{\mathrm{F}}^2\lesssim r\Big(\max\{\alpha(\bm{\Theta^*})^2,\sigma^2\}\frac{d\log d \log n}{n}\Big)^{1-q/2},$$
where $\alpha(\bm{\Theta^*})= \frac{d \| \bm{\Theta^*}\|_{\mathrm{\max}}}{\|\bm{\Theta^*}\|_{\mathrm{F}}}\in [1,d]$ is the spikiness   of the desired $\bm{\Theta^*}$. Compared with the full-data-based estimator in \cite{negahban2012restricted} that achieves near minimax rate \cite[Theorem 3]{negahban2012restricted}, our 1-bit estimator only degrades by a minor factor $\log n $, hence is also near minimax. It is quite striking that the underlying matrix can be recovered fairly well from merely 1-bit observation.
\label{remark4}
\end{rem}

\subsection{Heavy-tailed noise }
The heavy-tailed noise is assumed to have bounded second moment in this part, i.e., \begin{equation}
    \label{75_3}
    \mathbbm{E}\epsilon_k = 0,~\mathbbm{E}|\epsilon_k|^2\leq M.
\end{equation} 

Note that the 1-bit response $\dot{Y}_k$ is obtained with the truncation step before the dithered quantization --- $Y_k$ is first truncated to be $\widetilde{Y}_k = \mathrm{sign}(Y_k)\min\{|Y_k|,\eta\}$, then  dithered and quantized to $\dot{Y}_k = \mathrm{sign}(\widetilde{Y}_k + \Lambda_k)$ where $\Lambda_k \sim \mathrm{uni}([-\gamma,\gamma
])$. To invoke Corollary \ref{corollary3}, we first upper bound the right hand side of (\ref{4.7}). 

\begin{lem}
Consider (\ref{4.1}) under sampling scheme (\ref{4.2}),  max-norm constraint (\ref{4.5}), and heavy-tailed noise assumption (\ref{75_3}). For a specific $\delta >1$, we set the truncation threshold $\eta$, dithering scale $\gamma$ as
\begin{equation}
\begin{cases}
  \displaystyle \eta = C_{15}\max\{\alpha^*,\sqrt{M}\}\Big(\frac{n}{\delta d \log d}\Big)^{{1}/{4}} \\
  \displaystyle \gamma =C_{16}\max\{\alpha^*,\sqrt{M}\}\Big(\frac{n}{\delta d \log d}\Big)^{{1}/{4}} 
\end{cases},
    \label{5.19}
\end{equation}
where $C_{16}>C_{15}$, $\gamma > 2\max\{\alpha^*, \sqrt{M}\}$. If $\frac{\delta d \log d}{n}$ is sufficiently small, we have 
\begin{equation}
\big\|\frac{1}{n}\sum_{k=1}^n\big[\left<\bm{X_k,\Theta^*}\right>-\gamma\cdot \dot{Y}_k\big]\bm{X_k}\big\|_{\mathrm{op}} \lesssim \max\{\alpha^*,\sqrt{M}\} \Big(\frac{\delta \log d}{nd^3}\Big)^{1/4}
    \label{5.20}
\end{equation}
with probability higher than $1-2d^{1-\delta}$.
\label{lemma4}
\end{lem}
Parallel to proof of Theorem \ref{theorem9}, a discussion on whether $\bm{\widehat{\Delta}}\in\mathcal{C}(\psi)$ unfolds some key relations that further lead to the desired error bounds. The result is given in Theorem \ref{theorem10}.
\begin{theorem}
Under the setting of Lemma \ref{lemma4}, assume $\bm{\Theta^*}$ satisfies Assumption \ref{assumption4}, we consider the estimator $\bm{\widehat{\Theta}} $ defined in (\ref{4.6}). Moreover, we set $\lambda$ as 
\begin{equation}
\lambda  = C_{17}\max\{\alpha^*,\sqrt{M}\} \Big(\frac{\delta\log d}{nd^3}\Big)^{{1}/{4}}
    \label{5.21}
\end{equation}
with sufficiently large $C_{17}$. Assume $\frac{\delta d \log d}{n}$ is sufficiently small, $r\gtrsim d^q$, then with probability at least $1-3d^{1-\delta}$, we have 
\begin{equation}
    \begin{cases}
      \displaystyle \|\bm{\widehat{\Delta}}\|_{\mathrm{F}}^2/d^2 \lesssim rd^{-q}\Big(\max\{(\alpha^*)^2,M\} \sqrt{\frac{\delta d \log d}{n}}\Big)^{1-{q}/{2}}\\
      \displaystyle \|\bm{\widehat{\Delta}}\|_{\mathrm{nu}}/d \lesssim rd^{-q}
      \big(\max\{\alpha^*,\sqrt{M}\} \Big(\frac{\delta d\log d}{n}\Big)^{{1}/{4}}\big)^{1-q}
     \label{5.22}
    \end{cases}.
\end{equation}
\label{theorem10}
\end{theorem}
The result is   consistent with previous two estimation problems --- the error rates become essentially slower in the 1-bit heavy-tailed case. Similarly, this is the outcome of a bias-and-variance trade-off.

To shed some light on the fundamental difficulty of estimating $\bm{\Theta^*}$ from $(\bm{X}_k,\dot{Y}_k)$, we derive an information-theoretic lower bound in the following. Specifically, we consider the exact low-rank case ($q=0$) with the set of parameters \begin{equation}
\mathscr{K}(r,\alpha^*)= \big\{\bm{\Theta}\in \mathbb{R}^{d\times d}: \mathrm{rank}(\bm{\Theta})\leq r, \|\bm{\Theta}\|_{\max}\leq \alpha^*\big\}.
\end{equation}
\begin{theorem}\label{lower1}
    Given $n,d,r,\alpha^*$. For some underlying $\bm{\Theta}\in \mathscr{K}(r,\alpha^*)$, we suppose the data $(\bm{X}_k,\dot{Y}_k)_{k=1}^n$ are generated as in Theorem \ref{theorem10}, i.e., truncation, dithering and quantization with parameters (\ref{5.19}), and we assume $\gamma>\eta> \alpha^*$. Consider any algorithm which, for any underlying  $\bm{\Theta}\in \mathscr{K}(r,\alpha^*)$, takes the corresponding $(\bm{X}_k,\dot{Y}_k)_{k=1}^n$ as input and returns $\bm{\widehat{\Theta}}$. Then for some absolute constants $D_1,D_2$, there exists $\bm{\Theta}_0 \in \mathscr{K}(r,\alpha^*)$ such that with probability at least $\frac{3}{4}$, \begin{equation}\label{e4.25}
        \|\bm{\widehat{\Theta}}-\bm{\Theta}_0\|_F^2/d^2 \geq  \min\Big\{D_1\alpha^*, D_2((\alpha^*)^2+M)r\sqrt{\frac{d}{n\log d}}\Big\}.
    \end{equation} 
\end{theorem}

In the above lower bound, of primary interest is the second term --- it dominates the first term in the regular scaling $n \gtrsim r^2d $, and nearly matches the upper bound of Theorem \ref{theorem10} under $q=0$, up to a factor of $\log d$. Recall that the error rate in Theorem \ref{theorem10} suffers from essential degradation compared to the sub-Gaussian case. While the  lower bound indicates that, if one only has access to $(\bm{X}_k,\dot{Y}_k)_{k=1}^n$ (produced by our quantization scheme with the chosen parameters), the upper bound is indeed almost tight. In other words,     there exists no estimator as a function of $(\bm{X}_k,\dot{Y}_k)_{k=1}^n$ that could achieve error rate significantly faster than $\bm{\widehat{\Theta}}$ in Theorem \ref{theorem10}. 

The proof of Theorem \ref{lower1} is information-theoretic and inspired by \cite[Theorem 3]{davenport20141}, but requires some modifications because of different parameter sets and sampling schemes. Similarly to Theorem \ref{lower2}, the lifted lower bound mainly stems from the dithering scale (\ref{5.19}) that is larger than the sub-Gaussian counterpart  (\ref{4.11}).

To close this section, we point out that our method for 1-bit matrix completion is new and essentially different from the existing likelihood approach (see, e.g., \cite{davenport20141,cai2013max}). Notably, our method can deal with unknown pre-quantization noise $\epsilon_k$ that can be sub-Gaussian or heavy-tailed; note that such unknown noise precludes the standard likelihood approach. A review of prior works and more detailed comparison can be found in Appendix \ref{appendxD}.

\section{An Overview of the Techniques}
\label{proofsketch}
While deferring all the proofs to appendices, we provide an overview of the   techniques used in this work. We focus on the derivation of upper bounds. We detail the sub-Gaussian regime, and use concrete example for heavy-tailed case to illustrate that the same technicalities can derive the presented results with an optimal choice of parameters. Finally, we   compare our work with \cite{fan2021shrinkage} to illustrate the main technical reason why 1-bit quantization of heavy-tailed data leads to rate degradation.


\subsection{Sub-Gaussian Case}
For sparse covariance matrix estimation, the element-wise error rate of $\bm{\breve{\Sigma}}$ in Theorem \ref{theorem1} is a fundamental element. Unlike the full data case where $\mathbbm{E}\big(X_{k,i}X_{k,j}\big) = \sigma^*_{ij}$,  $\mathbbm{E}\breve{\sigma}_{ij} = \sigma^*_{ij}$  may not  hold due to the possibility of $|X_{k,i}|>\gamma$. Thus, we first divide the element-wise error into a concentration term $R_1$  and a bias term $R_2$
\begin{equation}
\nonumber
     |\breve{\sigma}_{ij}-\sigma^*_{ij}| \leq |\breve{\sigma}_{ij} -\mathbbm{E}\breve{\sigma}_{ij}|+|\mathbbm{E}\breve{\sigma}_{ij}- \sigma^*_{ij}|:=R_1+R_2.
\end{equation}
Since the quantized data is bounded, a fast concentration rate for $R_1$ is guaranteed by Hoeffding's inequality, while $R_2$ can be controlled by   standard arguments.  We   strike a balance between $R_1$, $R_2$ by setting $\gamma = O\big(\sqrt{\log \big(\frac{n}{\log d}\big)}\big)$, then the concentration term $R_1 =O \big(\sqrt{\frac{\log d (\log n)^2}{n}}\big)$ dominates the error, hence the error bound only degrades by a factor $\log n$ compared with $O\big(\sqrt{\frac{\log d}{n}}\big)$ for the full-data sample covariance matrix.

Recall that our estimator $\bm{\widehat{\Sigma}}$ is defined by element-wisely hard thresholding $\bm{\breve{\Sigma}}$, and  the procedures to show  operator norm error rate of $\bm{\widehat{\Sigma}}$ are  parallel to corresponding results for the full-data-based hard thresholding estimator $\mathcal{T}_\zeta \big(\sum_{k=1}^nX_kX_k^T/n\big)$ in \cite{cai2012minimax}. 
In brief,  some discussions unfold the  element-wise rate  $|\widehat{\sigma}_{ij}-\sigma^*_{ij}|=O\big(\min\{|\sigma^*_{ij}|,  \sqrt{\frac{\log d(\log n)^2}{n}}\}\big)$, which is tighter than the bound for $| \breve{\sigma}_{ij}-\sigma^*_{ij}|$. This tighter rate, together with the sparsity, can yield a dimension-free  bound for the dominating   term of operator norm error. Despite a similar proof strategy, we need more involved analyses to deal with some new challenges from the data quantization. These additional efforts, for example, can be seen in the treatment of $R_2$  (\ref{A.8}). 


For sparse linear regression (including 1-bit QC-CS, 1-bit CS) and matrix completion, 
we derive the error rates for each problem based on Lemma \ref{lemma2}, a framework of trace regression. Compared with the key lemma (Theorem 1) in \cite{fan2021shrinkage} , we present Lemma \ref{lemma2} in a   more general form that accommodates generalized quadratic loss (\ref{3.4}), and the purpose is that more flexible $\bm{Q}$, $B$ constructed from the binary data can be used. The advantage of using such framework is a rather clear proof roadmap constituted by two steps:
\begin{itemize}
    \item {\it Step 1.}
Bound $\|\mathrm{mat}(\bm{Q}\cdot\mathrm{vec}(\bm{\Theta^*}))-\bm{B}\|_{\mathrm{op}}$ from above and choose   $\lambda$ that   guarantee (\ref{3.5}); 
\item {\it Step 2.}
  Establish the restricted strong convexity (\ref{3.7}), and invoke (\ref{3.8}) to obtain the error rate.
\end{itemize}

We first discuss {\it Step 1}. In sparse linear regression we need $\lambda \geq 2\|\bm{Q}\Theta^*-B\|_{\max}$ with some $\bm{Q}, B$ approximating $\bm{\Sigma}_{XX} = \mathbbm{E}X_kX_k^T$, $\Sigma_{YX} = \mathbbm{E}Y_kX_k$, respectively. Thus, by noting $\bm{\Sigma}_{XX}\Theta^* =\Sigma_{YX} $ it can be divided as two approximation error terms 
$$ \| \bm{Q}\Theta^*-B\|_{\max} \leq  \underbrace{\|(\bm{Q}-\bm{\Sigma}_{XX} )\Theta^*\|_{\max}}_{\mathrm{approximation~term~I}} + \underbrace{\| B - \Sigma_{YX} \|_{\max}}_{\mathrm{approximation~term~II}}.$$ 
One possibility to control the approximation error term  is via existing results. For instance, in 1-bit QC-CS we set $\bm{Q}$ to be the proposed sparse covariance matrix estimator $\bm{\widehat{\Sigma}}$. Thus, the bound of term I follows from results in Section \ref{section2} (see, e.g., (\ref{add72_1})). On the other hand, we can also adopt a standard strategy of bounding the concentration error and the deviation (i.e., bias). For example, we can divide term II into (see, e.g., $R_2$, $R_3$ in (\ref{73p_7}))
$$ \|B - \Sigma_{YX}\|_{\max }\leq \underbrace{ \|B - \mathbbm{E} B \|_{\max} }_{\mathrm{concentration~term~II.1}}+ \underbrace{\| \mathbbm{E} \big( B - Y_kX_k\big) \|_{\max}}_{\mathrm{bias~term~II.2}}.$$
For matrix completion the methodology is similar, see (\ref{713_4}) for example. We apply various concentration inequalities to bound the concentration terms, to name a few,   Bernstein's inequality (\ref{713_7}), (\ref{713_6}), matrix Bernstein's inequality (\ref{713_5}). In contrast, more standard tools like Cauchy-Schwarz, Markov's inequality can upper bound bias terms. Let  $\lambda_{\mathrm{full}}$ denote the optimal choice of $\lambda$ in the full-data settings (see, e.g., \cite{negahban2011estimation,negahban2012restricted,fan2021shrinkage,chen2022color}). As it comes out, in the sub-Gaussian regime of our 1-bit setting, one can always strike an almost perfect balance among all the  terms such that  $\lambda = \mathrm{Poly}(\log n)\cdot \lambda_{\mathrm{full}}$ can guarantee    $\lambda \geq 2\|\mathrm{mat}(\bm{Q}\cdot\mathrm{vec}(\bm{\Theta^*}))-\bm{B}\|_{\mathrm{op}}$ (\ref{3.5}).

 {\it Step 2} concerns the restricted strong convexity of $\bm{Q}$ with regard to $\bm{\widehat{\Delta}}$. Note that this   mainly hinges on  the covariate. Thus, in 1-bit CS and matrix completion where full covariate is available, we can directly borrow existing results from the full-data settings with no quantization \cite{fan2021shrinkage,chen2022color}. For 1-bit QC-CS with quantized $X_k$, the desired RSC property straightforwardly follows from $\bm{\Sigma}_{XX}$'s sparsity and the resulting dimension-free bound of $\| \bm{\widehat{\Sigma}} - \bm{\Sigma}_{XX}\|_{\mathrm{op}}$. 
  Finally, we apply (\ref{3.8}) to obtain the error rates. Since  $\lambda = \mathrm{Poly}(\log n)\cdot \lambda_{\mathrm{full}}$ suffices for {\it Step 1}, under the dithered 1-bit quantization scheme, the error rate at worst degrades by logarithmic factor.

  \subsection{Heavy-tailed Case}
  In heavy-tailed regime we introduce truncation parameter $\eta$  and require $\gamma > \eta$. The  strategies and technical tools for the proofs are almost the same as sub-Gaussian regime, while the difference is that we can no longer strike a perfect balance  among all terms. We briefly give an example to demonstrate the proofs.  
  \begin{exmp}
  ({\it Theorem \ref{1bitcsht}.}) To our best knowledge, Theorem \ref{1bitcsht} presents the first computationally efficient method for 1-bit CS with heavy-tailed sensing vectors, and the rate $O\big(\sqrt{s}\sqrt[\leftroot{-3}\uproot{4}3]{\frac{\log d}{n}}\big)$ (for $s$-sparse $\Theta^*$) is still faster than the convex approach  in \cite{dirksen2021non} that is only for sub-Gaussian regime.  Let us start from {\it Step 1} and first decompose $\|\bm{\widehat{\Sigma}}_{\tilde{X}\tilde{X}} \Theta^* -  \widehat{\Sigma}_{YX}\|_{\max}$ into four terms (see notations given in (\ref{add72_9}))
  \begin{equation}
      \nonumber
      \begin{aligned}
     & \|\bm{\widehat{\Sigma}}_{\tilde{X}\tilde{X}} \Theta^* -  \widehat{\Sigma}_{YX}\|_{\max} \leq   \| (\bm{\widehat{\Sigma}}_{\tilde{X}\tilde{X}}-\bm{\Sigma}_{XX})\Theta^*\|_{\max} + \|\widehat{\Sigma}_{YX} - \Sigma_{YX}\|_{\max}:=\mathrm{I} + \mathrm{II}\vspace{15mm}\\
     &\leq  \underbrace{\big\|(\bm{\widehat{\Sigma}}_{\tilde{X}\tilde{X}}-\mathbbm{E}\widetilde{X}_{k}\widetilde{X}_k^T)\Theta^*\big\|_{\max} }_{\mathrm{concentration~term~I.1}}+\underbrace{ \big\| \mathbbm{E}\big(X_kX_k^T -\widetilde{X}_{k}\widetilde{X}_k^T \big)\Theta^*\big\|_{\max}}_{\mathrm{bias~term~I.2}} \\  &~~~~~~~~~~~~~~~~~~~~~~~~~~  + \underbrace{\big\| \widehat{\Sigma}_{YX} - \mathbbm{E}(\gamma \cdot \dot{Y}_k\widetilde{X}_k)\big\|_{\max}}_{\mathrm{concentration~term~II.1}}+ \underbrace{\big\| \mathbbm{E}\big(\gamma \cdot \dot{Y}_k\widetilde{X}_k - Y_kX_k\big) \big\|_{\max}}_{\mathrm{bias~term~II.2}}.
      \end{aligned}
  \end{equation}
  For two concentration terms, Bernstein's inequality 
  gives $\mathrm{I.1} = O\big(\sqrt{\frac{\log d}{n}}+ \eta_X^2 \frac{\log d}{n}\big)$ and $\mathrm{II.1} = O\big(\gamma \sqrt{\frac{\log d}{n}}+\frac{\gamma\cdot \eta_X\cdot\log d}{n}\big)$ with probability $1-d^{-\Omega(1)}$. For two bias terms, some probability arguments and bounded $4$-th moment can yield $\mathrm{I.2} = O\big(\frac{1}{\eta_X^2}\big)$ and $\mathrm{II.2} =O\big(\frac{1}{\eta_Y^2}+\frac{1}{\eta_X^2}\big)$. Recall that the heavy-tailed $Y_k$ would be quantized to 1-bit, and we require $\gamma > \eta_Y$. To achieve an optimal trade-off among $\eta_X,\eta_Y, \gamma$, we set $\eta_X \asymp \big(\frac{n}{\log d}\big)^{1/4}$, $\eta_Y, \gamma \asymp \big(\frac{n}{\log d}\big)^{1/6}$, which gives an overall upper bound $ \|\bm{\widehat{\Sigma}}_{\tilde{X}\tilde{X}} \Theta^* -  \widehat{\Sigma}_{YX}\|_{\max} = O\big(\sqrt[\leftroot{-3}\uproot{4}3]{\frac{\log d}{n}}\big)$. Hence, $\lambda \asymp \sqrt[\leftroot{-3}\uproot{4}3]{\frac{\log d}{n}}$ suffices for $\lambda \geq 2\|\bm{\widehat{\Sigma}}_{\tilde{X}\tilde{X}} \Theta^* -  \widehat{\Sigma}_{YX}\|_{\max}$.

  For {\it Step 2}, since the truncated sample covariance matrix $\bm{\widehat{\Sigma}}_{\tilde{X}\tilde{X}}$ also serves as a plug-in estimator for sparse linear regression in \cite{fan2021shrinkage}, we can directly borrow their Lemma 2(b). It should be pointed out that if  we treat $\widetilde{X}_k$  as   data bounded by $\eta_X$ and deal with I.1, II.1 via Hoeffding's inequality, we can only establish an essentially slower error rate. By contrast, Bernstein's inequality 
  enables us to make full use of $X_k$'s   bounded $4$-th moment and derive tighter bound.
  \hfill$\triangle$  
  \end{exmp}

  \subsection{Comparison With the Heavy-tailed Full-data Case}
Finally, we   compare the heavy-tailed, full-data setting in \cite{fan2021shrinkage} and our heavy-tailed, 1-bit quantized setting, to explain the main technical reason why near optimal rates are derived in \cite{fan2021shrinkage}, but ours are essentially slower. 
The key difference is on the effectiveness of the original moment constraint. Because the truncated data admits the same moment constraint as the original data
(e.g., $\mathbbm{E}|\widetilde{Y}_k|^4 \leq \mathbbm{E}|Y_k|^4\leq M$), \cite{fan2021shrinkage} can use (matrix) Bernstein's inequality to deal with the concentration term. Nevertheless, quantizing $\widetilde{Y}_k$ to its 1-bit surrogate $\gamma\dot{Y}_k$ ruins the moment constraint since $\mathbbm{E}|\gamma\cdot\dot{Y}_k|^4 =\gamma^4$. As a consequent, we can only derive a looser bound for the concentration term. 

We use sparse linear regression as a concrete example for illustration.

 \begin{exmp}
 ({\it sparse linear regression in \cite{fan2021shrinkage} and this work.}) In the proof of  \cite[Lemma 1]{fan2021shrinkage}, Bernstein's inequality 
 is used to deal with the concentration term $\|\frac{1}{n}\sum_{k=1}^n \widetilde{Y}_k\widetilde{X}_k-\mathbbm{E}\widetilde{Y}_k\widetilde{X}_k\|_{\max}$. Thanks to the moment constraints of $\widetilde{X}_k$, $\widetilde{Y}_k$, they can show \begin{equation}
     \big\|\frac{1}{n}\sum_{k=1}^n \widetilde{Y}_k\widetilde{X}_k-\mathbbm{E}\widetilde{Y}_k\widetilde{X}_k\big\|_{\max} = O\big(\sqrt{\frac{\log d}{n}}+\frac{\eta_X\eta_Y \log d}{n}\big)
     \label{718add4}
 \end{equation}
 with high probability. By contrast, in our Theorem \ref{1bitcsht} for 1-bit CS, the corresponding term is the concentration term II.1 in Example 1. Since $\gamma \cdot \dot{Y}_k$ fails to inherit the moment constraint from $Y_k$,   the same Bernstein's inequality only delivers (see (\ref{718_add2}), (\ref{718_add3}))
\begin{equation}
    \big\| \frac{1}{n}\sum_{k=1}^n \gamma\cdot\dot{Y}_k \widetilde{X}_k-\mathbbm{E}\gamma\cdot\dot{Y}_k \widetilde{X}_k\big\|_{\max} = O\Big(\gamma\big( \sqrt{\frac{\log d}{n}}+\frac{\eta_X \log d}{n}\big)\Big)
    \label{718add5}
\end{equation} with probability $1-d^{-\Omega(1)}$, which is worse since $\gamma$ becomes a common factor. Furthermore, in our Theorem \ref{theorem8} for 1-bit QC-CS the corresponding concentration term is $\|\frac{\gamma^2}{n}\sum_{k=1}^n \dot{Y}_k\dot{X}_k-\mathbbm{E} \gamma^2 \cdot \dot{Y}_k\dot{X}_k\|_{\max}$. Note that both covariate and response are quantized and hence lose the moment constraint. Thus, we directly invoke Hoeffding's inequality and obtain (see (\ref{72add2}), (\ref{718add1}))
\begin{equation}
    \big\|\frac{\gamma^2}{n}\sum_{k=1}^n \dot{Y}_k\dot{X}_k-\mathbbm{E} \gamma^2 \cdot \dot{Y}_k\dot{X}_k\big\|_{\max} = O\big(\gamma^2\sqrt{{\frac{\log d}{n}}}\big) \label{718add6}
\end{equation}
 where $\gamma^2$ appears as a leading multiplicative factor. It shall be clear that $\gamma$ or $\gamma^2$ appearing as a multiplicative factor of $\sqrt{\frac{\log d}{n}}$ leads to essential degradation. \hfill $\triangle$
  \end{exmp}



\section{Experimental Results}
In this section we present experimental results on synthetic data that can corroborate and demonstrate our theories. To facilitate  the presentation flow,  the simulation details and the algorithms  are provided in Appendix \ref{appendixE}. 
\label{section5}
\subsection{Sparse Covariance Matrix Estimation}
\begin{figure}[ht]
    \centering
    \includegraphics[scale = 0.8]{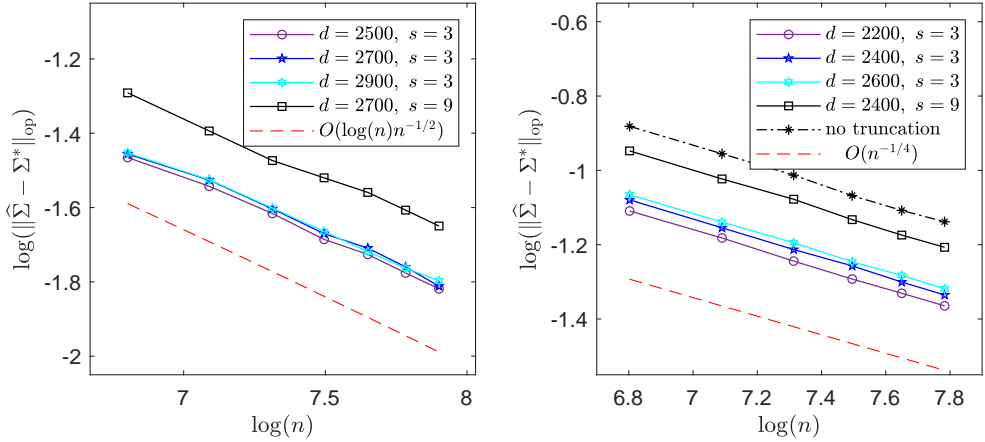}
    \caption{ Sparse covariance matrix estimation. Left: Sub-Gaussian; Right: Heavy-tailed.}
    \label{fig1}
\end{figure}

In our simulation $\bm{\Sigma^*}$ has exactly $s$-sparse columns. In sub-Gaussian regime, with high probability Theorem \ref{theorem3} provides the error bound
\begin{equation}
    \|\bm{\widehat{\Sigma}}- \bm{\Sigma^*}\|_{\mathrm{op}} \lesssim s  \sqrt{\frac{(\log n)^2\log d}{n}}.
    \label{5.1}
\end{equation}
Thus, the operator norm error is expected to only logarithmically depends on the ambient dimension $d$, while essentially depend on the    the sparsity $s$ (that can be viewed as the intrinsic dimension of the problem). We draw $X_k$ from multivariate Gaussian distribution to verify the theory. Specifically, we try $(d,s) =(2500,3)$, $(2700,3)$, $(2900,3)$, $(2700,9)$, and test the sample size $n=900:300:2700$ for each $(d,s)$. The log-log error curves for all $(d,s)$ are plotted on the left of Figure \ref{fig1}, with the theoretical curve $O\big(\frac{\log n}{\sqrt{n}}\big)$ also provided for comparison of the error rate. Clearly, the curves with different dimension $d$ but the same sparsity $s$ are almost coincident, which confirms the inessential dependency  on $d$ for the error. On the other hand, the estimation error depends on $s$ non-trivially since the   curve of $s=9$ is obviously higher. Moreover, the experimental curves are roughly parallel to the theoretical one, confirming a near optimal decreasing rate of $n^{-1/2}$.

In heavy-tailed regime, for $\Sigma^*$ with $s$-sparse columns Theorem \ref{theorem6} guarantees 
\begin{equation}
\|\bm{\widehat{\Sigma}}-\bm{\Sigma^*}\|_{\mathrm{op}} \lesssim s \Big(\frac{\log d}{n}\Big)^{1/4}.
    \label{5.2}
\end{equation}
 The relation between estimation error and parameters $s,d$ are similar to (\ref{5.1}), while the convergence rate becomes slower. In our simulations, heavy-tailed data are drawn from Student's t distribution. We test $(d,s) = (2200,3)$, $(2400,3)$, $(2600,3)$, $(2400,9)$ under sample size $n = 900:300:2400$.
 We report the results in the right figure of Figure \ref{fig1}. Consistent with the error bound, three curves with same $s$ but different $d$ are fairly close, while   larger $s$ ($s=9$) leads to essentially larger error. Although our theoretical rate $O\big(n^{-1/4}\big)$ does not match the optimal rate in the classical setting, these curves seem well aligned with the theoretical curve. Furthermore, we test $(d,s)=(2400,9)$ with the truncation step removed and then show the error curve with legend "no truncation". One shall see the estimation error becomes worse without truncation. Therefore, truncation is not merely of technical importance, but can indeed lower the estimation error in heavy-tailed regime.  

\subsection{Sparse Linear Regression}
\begin{figure}[ht]
    \centering
    \includegraphics[scale = 0.77]{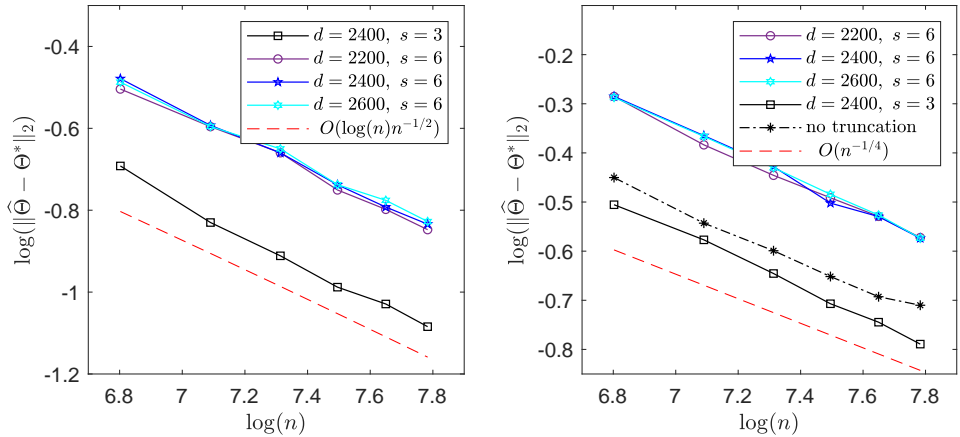}
    \caption{1-bit QC-CS. Left: Sub-Gaussian; Right: Heavy-tailed.}
    \label{fig2}
\end{figure}

\textbf{1-bit quantized-covariate compressed sensing (1-bit QC-CS).} In 1-bit QC-CS, both covariate $X_k$ and response $Y_k$ are quantized to 1-bit. Note that we use exactly sparse $\Theta^*$,  hence  in sub-Gaussian regime  Theorem \ref{theorem6} delivers the guarantee  
\begin{equation}
\|\widehat{\Theta}-\Theta^*\|_2 \lesssim \log n\sqrt{\frac{s\log d}{n}},
    \label{5.3}
\end{equation}
while for heavy-tailed regime the error bound in Theorem \ref{theorem7} reads as  
\begin{equation}
    \|\widehat{\Theta}-\Theta^*\|_2 \lesssim \sqrt{s}\Big(\frac{\log d}{n} \Big)^{1/4}.
    \label{5.4}
\end{equation}
With simulation  details given in Appendix \ref{appendixE}, we try $\Theta^*$ with $(d,s)=(2400,3)$, $(2200,6)$, $(2400,6)$, $(2600,6)$ under $n = 900:300:2400$. The experimental results in sub-Gaussian regime, heavy-tailed regime are shown as log-log curves on the left, the right of Figure \ref{fig2}, respectively. We also plot the theoretical rates for comparison. To show the efficacy of truncation in heavy-tailed regime, keeping other parameters unchanged, we test $(d,s)=(2400,3)$ without truncation step. The errors are accordingly shown as a curve with legend "no quantization".

The results corroborate the theory from several respects. Firstly,   the curves with the same $s$ but different $d$ are extremely close, while the errors under $s=6$ are significantly larger than $s=3$. This verifies (\ref{5.4}) and (\ref{5.5}) that exhibit   non-trivial dependence on the sparsity $s$ but only logarithmic dependence on the ambient dimension $d$. Secondly, since the experimental curves are fairly aligned with the theoretical ones, the theoretical convergence rates (regarding $n$) are verified.  Moreover, comparing the curve of $(d,s)=(2400,3)$ and "no quantization" on the right of Figure \ref{fig2}, shrinking heavy-tailed data indeed leads to more accurate estimation of $\Theta^*$.

\vspace{2mm}

\begin{figure}[ht]
    \centering
    \includegraphics[scale = 0.8]{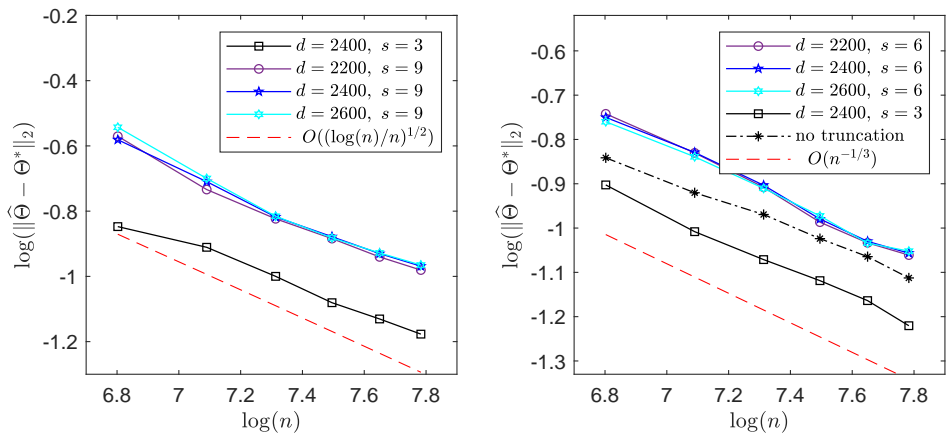}
    \caption{1-bit CS. Left: Sub-Gaussian; Right: Heavy-tailed.}
    \label{fig2.1}
\end{figure}

\noindent
\textbf{1-bit   compressed sensing (1-bit CS).} Different from the novel setting of 1-bit QC-CS, in 1-bit CS one has full covariate $X_k$ and only quantize $Y_k$ to 1-bit. Under $s$-sparse $\Theta^*$, Theorem \ref{1bitcssg} gives the near minimax error bound for sub-Gaussian regime \begin{equation}
    \label{6.6}
    \|\widehat{\Theta}-\Theta^*\|_2 \lesssim \sqrt{\frac{s\log d\log n}{n}},
\end{equation}
while Theorem \ref{theorem10} for heavy-tailed regime provides 
\begin{equation}
    \label{6.7}
        \|\widehat{\Theta}-\Theta^*\|_2 \lesssim \sqrt{s}\Big(\frac{\log d}{n}\Big)^{\frac{1}{3}}.
\end{equation}
 Under sample size $n = 900:300:2400$,   we test $(d,s)= (2400,3)$, $(2200,9)$, $(2400,9)$, $(2600,9)$ for sub-Gaussian case, while   $(d,s) = (2400,3)$, $(2200,6)$,  $(2400,6)$,  $(2600,6)$ for heavy-tailed case. For $(d,s)= (2400,3)$ with heavy-tailed data, we also conduct an independent simulation with the truncation of $X_k$, $Y_k$ removed but other conditions unchanged. The log-log error curves and the theoretical rates are plotted in Figure \ref{fig2.1}. The key implications of Figure \ref{fig2.1} are similar to those in Figure \ref{fig2} and can support our theoretical error bounds (\ref{6.6}), (\ref{6.7}).

\subsection{Low-rank Matrix Completion}

\begin{figure}[ht]
    \centering
    \includegraphics[scale = 0.71]{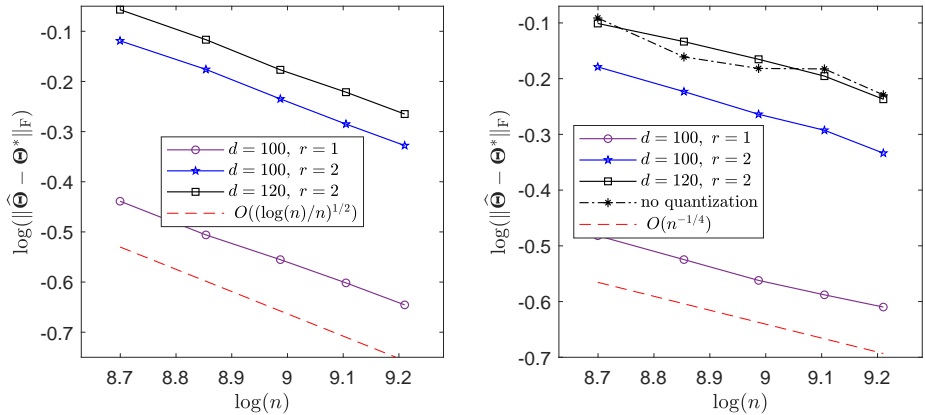}
    \caption{Low-rank matrix completion. Left: Sub-Gaussian; Right: Heavy-tailed.}
    \label{fig3}
\end{figure}

While the error bounds in Theorems \ref{theorem9}-\ref{theorem10} are stated under  $\frac{\|\bm{\widehat{\Delta}}\|_{\mathrm{F}}^2}{d^2}$,  we first adapt them to our   simulation   (see Appendix \ref{appendixE}) where the underlying exactly low-rank matrices have comparable spikiness $\alpha(\bm{\Theta^*})$ and unit Frobenius norm, and the noise is moderate compared with the signal. Specifically, under sub-Gaussian $\epsilon_k$, we   can translate MSE error bound in (\ref{5.17}) into  
  \begin{equation}
   {\|\bm{\widehat{\Theta}}-\bm{\Theta^*}\|_{\mathrm{F}}}  \lesssim \sqrt{\frac{rd\log d \log n}{n}},
    \label{5.5}
\end{equation}
and similarly  for heavy-tailed case (\ref{5.22})  
\begin{equation}
\|\bm{\widehat{\Theta}}-\bm{\Theta^*}\|_{\mathrm{F}} \lesssim   \Big(\frac{r^2d\log d}{n}\Big)^{1/4}.
    \label{5.6}
\end{equation}
To corroborate the theoretical error rates, we simulate the proposed 1-bit matrix completion method using $\bm{\Theta^*}$ with $(d,r)= (100,1)$, $(100,2)$, $(120,2)$, under the sample size $n=6000:1000:10000$. In heavy-tailed regime,  we also try $(d,r)=(120,2)$ with the response truncation step removed. The experimental results are plotted as log-log error curves in Figure \ref{fig3}.

Clearly, in both sub-Gaussian regime (left figure) and heavy-tailed regime (right figure), the errors significantly increase when either $r$ or $d$ becomes larger. This corroborates the implications of (\ref{5.5}), (\ref{5.6}) that the estimation error essentially hinges on $r$ and $d$. Moreover, the experimental curves are well aligned with the theoretical curve, hence the theoretical error rates are confirmed. Comparing two black curves of $(d,r)=(120,2)$ and "no quantization" in the right figure, the truncation step seems do not bring notable improvement to the recovery of  $\bm{\Theta^*}$. This is perhaps because the  the moderate noise $\frac{1}{250\sqrt{3}}\cdot t(\nu = 3)$ is used in the simulation, thus making the bias-and-variance trade-off less important. On the other hand, we believe a more significant advantage of using the truncation step can be observed under severer noise.

\section{Concluding Remarks}
\label{section6}

In this paper we propose a   dithered 1-bit quantization scheme and apply it to the estimation problems of sparse covariance matrix estimation, sparse linear regression and matrix completion. While adding a uniform dithering noise prior to quantization   was already seen in literature, our scheme involves a truncation step if the data are heavy-tailed. 
Under  high-dimensional scaling,  our estimators from merely   binary data can   recover the underlying parameters fairly well. In sub-Gaussian regime, the proposed estimators achieve near minimax rates. In heavy-tailed regime, the error rates become significantly slower because of a bias-and-variance trade-off. However, these results either represent the first ones under 1-bit quantization of heavy-tailed data, or already improve on prior results. Moreover, we also derive nearly matching information-theoretic lower bounds for Theorem \ref{1bitcsht}, \ref{theorem10} (heavy-tailed setting of 1-bit compressed sensing, matrix completion), showing that the rates are actually almost tight for estimation with the observed 1-bit data.

This work also provides new developments to each of the three estimation problems. Compared to \cite{dirksen2022covariance} that proposed the 1-bit covariance matrix estimator $\bm{\breve{\Sigma}}$, the results   in Section \ref{section2} can be viewed as a two-fold extension, that is, extension to high-dimensional scaling $(n<d)$  and to heavy-tailed distribution. 
For sparse linear regression, we first  propose and study a novel complete quantization setting where both covariate and response are quantized to 1-bit (Theorems \ref{theorem7}-\ref{theorem8}). Corresponding results for 1-bit compressed sensing (1-bit CS) are also presented  (Theorems \ref{1bitcssg}-\ref{1bitcsht}). Compared with   previous results on 1-bit CS, our sensing vector can be sub-Gaussian or heavy-tailed, and other advantages include faster rate 
and convex recovery program, see Appendix \ref{appendxD}. In Section \ref{section4}, while all existing papers for 1-bit matrix completion (1-bit MC) are in essence based on maximum likelihood estimation, our novel method can handle   pre-quantization random noise with unknown distribution, see Appendix \ref{appendxD}.

 We point out two open questions to close this work. The first one is concerned with the   slower error rates under the 1-bit quantization of heavy-tailed data. While  they are   nearly tight in 1-bit CS and 1-bit MC from our binary observations (Theorems \ref{lower2}, \ref{lower1}), it is still possible to design different (1-bit) quantization schemes for heavy-tailed data that allow faster rate. The possibility includes changing parameters in our quantization scheme. In fact, the lifted lower bounds in Theorems \ref{lower2}, \ref{lower1} 
are mainly due to our choices of larger dithering scale. Such choice is the price we pay for handling heavy-tailed data, but it is possible to achieve faster rate by using smaller dithering scale  (of course, deriving the faster rate in this case will rely on sharper technical tool). We leave  faster estimator in a heavy-tailed, 1-bit quantized case as future work. Secondly, our results in Section \ref{section3} are non-uniform, meaning that the recovery guarantee is valid for a fixed $\Theta^*$. Note that in nonlinear compressed sensing uniform guarantee is an important aspect and still eagerly pursued (see \cite{chen2022quantizing,xu2020quantized,chen2022uniform,genzel2022unified} for recent advances). Thus, it would be a good direction to strengthen Theorems \ref{theorem7}-\ref{1bitcsht} to a uniform ones that guarantee the recovery of all (approximately) sparse signals. For this direction, we conjecture that the main technical difficulties will lie in the 1-bit quantization and the heavy-tailed data (we refer to \cite[Theorem 12]{chen2022quantizing} for a uniform recovery guarantee under uniformly quantized heavy-tailed data).


\bibliographystyle{plain}
\bibliography{libr}

\appendix
\section{Proofs: Sparse Covariance Matrix Estimation}\label{appendixA}
\noindent{\textbf{Proof of Lemma \ref{lemma1}}.}
Since $X$ and $\Lambda$ are independent, we have 
\begin{equation}
    \begin{aligned}
    &\mathbbm{E}\Big[\gamma\cdot \mathrm{sign}(X+\Lambda)\Big] =  \mathbbm{E}_X\mathbbm{E}_{\Lambda}\Big[\gamma\cdot \mathrm{sign}(X+\Lambda)\Big] = \mathbbm{E}_X \Big[\gamma\cdot \mathbbm{P}(\Lambda \geq -X)+ (-\gamma)\cdot \mathbbm{P}(\Lambda <-X)\Big] \\
    &=   \mathbbm{E}_X  \Big[\gamma\cdot\Big(\frac{\gamma +X}{2\gamma} - \frac{\gamma-X}{2\gamma}\Big)\Big]
    =  \mathbbm{E}X.
    \nonumber
    \end{aligned} 
\end{equation}
note that the third equal sign relies on $\gamma \geq  B$.
\hfill $\square$

\vspace{3mm}
\noindent{\textbf{Proof of Corollary \ref{corollary1}}.} Since $\Lambda_1$ and $\Lambda_2$ are i.i.d. uniformly distributed on $[-\gamma,\gamma]$ and independent of $X,Y$, then by using Lemma \ref{lemma1} we have 
\begin{equation}
    \begin{aligned}
    & \mathbbm{E}\Big[\gamma^2\cdot \mathrm{sign}(X+\Lambda_1)\mathrm{sign}(Y+\Lambda_2)\Big] 
    =  \mathbbm{E}_{X,Y}\mathbbm{E}_{\Lambda_1}\mathbbm{E}_{\Lambda_2}\Big[\gamma^2\cdot \mathrm{sign}(X+\Lambda_1)\mathrm{sign}(Y+\Lambda_2)\Big] \\
    & = \mathbbm{E}_{X,Y}\Big(\mathbbm{E}_{\Lambda_1}\Big[\gamma\cdot \mathrm{sign}(X+\Lambda_1)\Big]\mathbbm{E}_{\Lambda_2}\Big[\gamma \cdot\mathrm{sign}(Y+\Lambda_2)\Big]\Big)=\mathbbm{E}XY,
      \nonumber
    \end{aligned}
\end{equation}
the result follows.   \hfill $\square$

\subsection{Sub-Gaussian Data}
\noindent{\textbf{Proof of Theorem \ref{theorem1}}.} For fixed $i,j$, triangle inequality yields \begin{equation}
    \label{tech1}
    |\breve{\sigma}_{ij}-\sigma^*_{ij}| \leq |\breve{\sigma}_{ij} -\mathbbm{E}\breve{\sigma}_{ij}|+|\mathbbm{E}\breve{\sigma}_{ij}- \sigma^*_{ij}|:=R_1+R_2.
\end{equation} It suffices to bound $R_1$, $R_2$ from above.

\noindent{\underline{\it Bound of $R_1$.}} We introduce the element-wise notation of the quantized data as $\dot{X}_{kj}=~$ $[\dot{X}_{kj,1},\dot{X}_{kj,2},...,\dot{X}_{kj,d}]^T,\ \forall\ k\in [n],j\in [2],$ then by (\ref{2.1}) $\breve{\sigma}_{ij} = \frac{1}{n}\sum_{k=1}^n \frac{\gamma^2}{2}\big[\dot{X}_{k1,i}\dot{X}_{k2,j} + \dot{X}_{k2,i}\dot{X}_{k1,j}\big]$. Since $\Big|\frac{\gamma^2}{2}\big[\dot{X}_{k1,i}\dot{X}_{k2,j} + \dot{X}_{k2,i}\dot{X}_{k1,j}\big]\Big|\leq \gamma^2$, Hoeffding's inequality (Proposition \ref{pro3}) yields $$\mathbbm{P}(|\breve{\sigma}_{ij}-\mathbbm{E}\breve{\sigma}_{ij}|\geq t)\leq 2\exp(-nt^2/2\gamma^4),\ \forall~ t>0.$$
We set $t = \gamma^2 \sqrt{\frac{2\delta \log d}{n}}$ and obtain 
\begin{equation}
\mathbbm{P}\Big(R_1\geq \gamma^2 \sqrt{\frac{2\delta \log d}{n}}\Big) \leq 2d^{-\delta}.
    \label{A.1}
\end{equation}

\noindent{\underline{\it Bound of $R_2$.}} By Corollary \ref{corollary1} and some algebra, we have 
\begin{equation}
   \begin{aligned}
    &R_2  =  \big|\mathbbm{E} \big(\gamma^2 \cdot \dot{X}_{k1,i}\dot{X}_{k2,j}-X_{k,i}X_{k,j}\big)\big|\\  & =    \Big|\mathbbm{E} [\gamma^2\dot{X}_{k1,i}\dot{X}_{k2,j}- X_{k,i}X_{k,j}][\mathbbm{1}(\{|X_{k,i}|\geq \gamma\}\cup \{|X_{k,j}|>\gamma\})] \Big|\\&\leq \mathbbm{E}|X_{k,i}X_{k,j}|\mathbbm{1}(|X_{k,i}|>\gamma) +\mathbbm{E}|X_{k,i}X_{k,j}|\mathbbm{1}(|X_{k,j}|>\gamma):=R_{21}+R_{22}.
    \nonumber
   \end{aligned}
\end{equation}
Note that $R_{21},R_{22}$ can be bounded likewise, thus we only show the upper bound of $R_{21}$. We use Cauchy-Schwarz inequality, and then Proposition \ref{pro1}, it yields
\begin{equation}
    \nonumber
    \begin{aligned}
   & R_{21} \leq \sqrt{\mathbbm{E}|X_{k,i}X_{k,j}|^2 }\cdot \sqrt{\mathbbm{P}(|X_{k,i}|>\gamma)}  \leq \sqrt{\frac{1}{2}\mathbbm{E}(|X_{k,i}|^4 + |X_{k,j}|^4)}\cdot \sqrt{\mathbbm{P}(|X_{k,i}|>\gamma)}\\& \lesssim \sqrt{\sigma^4} \cdot \sqrt{\exp\big(-\frac{D_1\gamma^2}{\sigma^2}\big)} \leq  \sigma^2 \exp\big(-\frac{D_1\gamma^2}{2\sigma^2}\big).
    \end{aligned}
\end{equation}
We further plug in (\ref{2.6}) and assume $C_1$ is sufficiently large such that $D_1C_1^2 \geq 1$, it delivers $R_{21} \lesssim \sigma^2 \sqrt{\frac{2\delta \log d}{n}}.$ Therefore, we conclude that \begin{equation}
    \label{A.4}
    R_2 = |\mathbbm{E}\breve{\sigma}_{ij}-\sigma^*_{ij}| \lesssim \sigma^2\sqrt{\frac{2\delta \log d}{n}}. 
\end{equation}

Combining (\ref{A.1}) and (\ref{A.4}) we derive $\mathbbm{P}\Big(|\breve{\sigma}_{ij}-\sigma^*_{ij}| \lesssim \gamma^2 \sqrt{\frac{\delta\log d}{n}}\Big) \geq 1-2d^{\delta}.$
With no loss of generality, we can assume $2\delta \log d>e$, then $\gamma^2 \lesssim \sigma^2\log n$, then (\ref{2.7}) follows. It is not hard to see that (\ref{2.8}) follows from (\ref{2.7}) via a union bound.  \hfill $\square$

\vspace{3mm}
\noindent{\textbf{Proof of Theorem \ref{theorem2}}.} Since $\gamma$ has been specified with some $C_1$, from Theorem \ref{theorem1} we know there exists an absolute constant $D_1$ such that 
\begin{equation}
\mathbbm{P}\Big(|\breve{\sigma}_{ij}-\sigma^*_{ij}| \leq D_1\sigma^2 \log n\sqrt{\frac{\delta \log d}{n}}\Big) \geq 1-2d^{-\delta}.
    \label{A.5}
\end{equation}
Assume $C_2$ is sufficiently large such that $C_2 >D_1$. We first rule out $2d^{-\delta}$ probability and assume $|\breve{\sigma}_{ij}-\sigma^*_{ij}| \leq D_1\sigma^2 \log n\sqrt{\frac{\delta \log d}{n}}$. Recall that $\widehat{\sigma}_{ij}=\mathcal{T}_{\zeta}(\breve{\sigma}_{ij})$, we analyse two cases.

\noindent{\underline{\it Case 1.}} $|\breve{\sigma}_{ij}|<\zeta$, then by definition we have $\widehat{\sigma}_{ij}=0$, hence  $|\widehat{\sigma}_{ij}-\sigma^*_{ij}| =|\sigma^*_{ij}| \leq |\sigma^*_{ij}|$. Besides, by triangle inequality we have $|\sigma^*_{ij}|\leq |\sigma^*_{ij}-\breve{\sigma}_{ij}|+|\breve{\sigma}_{ij}|\leq (D_1+C_2)\sigma^2 \log n \sqrt{\frac{\delta \log d}{n}},$
hence we have
$$|\widehat{\sigma}_{ij}-\sigma^*_{ij}| \leq (D_1+C_2+1)\min\Big\{|\sigma^*_{ij}|,\sigma^2\log n \sqrt{\frac{\delta \log d}{n}}\Big\}.$$
\noindent{\underline{\it Case 2.}} $|\breve{\sigma}_{ij}|\geq \zeta$, then we have $\widehat{\sigma}_{ij} = \breve{\sigma}_{ij}$, hence $|\widehat{\sigma}_{ij}-\sigma^*_{ij}| = |\breve{\sigma}_{ij}-\sigma^*_{ij}| \leq D_1\sigma^2\log n \sqrt{\frac{\delta \log d}{n}}$. Moreover, since $C_2>D_1$, we have 
$|\sigma^*_{ij}|\geq |\breve{\sigma}_{ij}| - |\breve{\sigma}_{ij}- \sigma^*_{ij}| \geq (C_2-D_1)\sigma^2 \log n \sqrt{\frac{\delta \log d}{n}},$ which implies that $\sigma^2\log n \sqrt{\frac{\delta \log d}{n}} \leq \frac{1}{C_2-D_1}|\sigma^*_{ij}|$, hence we have 
$|\widehat{\sigma}_{ij}-\sigma^*_{ij}| \leq \frac{D_1}{C_2-D_1}|\sigma^*_{ij}|.$
By putting pieces together we obtain 
$$|\widehat{\sigma}_{ij}-\sigma^*_{ij}| \leq \Big(D_1+ \frac{D_1}{C_2-D_1}\Big)\min\Big\{|\sigma^*_{ij}|,\sigma^2\log n \sqrt{\frac{\delta \log d}{n}}\Big\}.$$
Combining two cases leads to (\ref{2.10}), hence the proof is concluded. \hfill $\square$

\vspace{3mm}
\noindent{\textbf{Proof of Theorem \ref{theorem3}}.} Since $\gamma$ and $\zeta$ are properly set with some $C_1,C_2$, by Theorem \ref{theorem2}, (\ref{2.10}) holds with some absolute constant $D_1$ hidden behind "$\lesssim$". For convenience we define \begin{equation}
    \mathscr{A}_{ij}= \Big\{|\widehat{\sigma}_{ij}-\sigma^*_{ij}|\leq D_1 \min\{|\sigma^*_{ij}|,\sigma^2\log n \sqrt{\frac{\delta \log d}{n}}\}\Big\}.
    \label{A.6}
\end{equation}
Let $\mathscr{A}_{ij}^c$ be its complement, then we have $\mathbbm{P}(\mathscr{A}_{ij}^c) \leq 2d^{-\delta}.$ For $d\times d$ symmetric matrix $A = [\alpha_1,...,\alpha_n]$ with columns $\alpha_j$, we have $\|A\|_{\mathrm{op}}\leq \sup_{j\in [d]}\|\alpha_j\|_1$. Thus, some algebra gives
\begin{equation}
    \begin{aligned}
    \label{74add_1}
   & \mathbbm{E}\|\bm{\widehat{\Sigma}-\Sigma^*}\|_{\mathrm{op}}^p  \leq \mathbbm{E}\Big[\sup_{j\in[d]}\sum_{i=1}^d|\widehat{\sigma}_{ij}-\sigma^*_{ij}|\Big]^p 
   \leq  \mathbbm{E}\sup_{j\in [d]}\Big[\sum_{i=1}^d|\widehat{\sigma}_{ij}-\sigma^*_{ij}|\mathbbm{1}(\mathscr{A}_{ij}) + \sum_{i=1}^d|\widehat{\sigma}_{ij}-\sigma^*_{ij}|\mathbbm{1}(\mathscr{A}_{ij}^c) \Big]^p \\
   &\leq  {2^p\mathbbm{E}\sup_{j\in[d]}\Big[\sum_{i=1}^d|\widehat{\sigma}_{ij}-\sigma^*_{ij}|\mathbbm{1}(\mathscr{A}_{ij})\Big]^p} +  {2^p\mathbbm{E}\sup_{j\in[d]}\Big[\sum_{i=1}^d|\widehat{\sigma}_{ij}-\sigma^*_{ij}|\mathbbm{1}(\mathscr{A}_{ij}^c)\Big]^p} := R_1+R_2,
    \end{aligned}
\end{equation}
\noindent{\underline{\it Bound of $R_1$.}} Let us first bound $R_1$. By (\ref{2.3}) and (\ref{A.6}) we have 
\begin{equation}
    \begin{aligned}
    &\sum_{i=1}^d|\widehat{\sigma}_{ij}-\sigma^*_{ij}|\mathbbm{1}(\mathscr{A}_{ij}) \leq \sum_{i=1}^d D_1\min\Big\{|\sigma^*_{ij}|,\sigma^2\log n \sqrt{\frac{\delta \log d}{n}}\Big\} \\
    \leq &\sum_{i=1}^d D_1|\sigma^*_{ij}|^q \Big(\sigma^2\log n \sqrt{\frac{\delta \log d}{n}}\Big)^{1-q} \leq D_1s\Big(\sigma^2\log n \sqrt{\frac{\delta \log d}{n}}\Big)^{1-q}.
       \label{713_1}
    \end{aligned}
\end{equation}
This further gives  \begin{equation}
R_1 \leq \Big(2D_1s\Big[\sigma^2\log n \sqrt{\frac{\delta\log d}{n}}\Big]^{1-q}\Big)^p.
    \nonumber
\end{equation}
\noindent{\underline{\it Bound of $R_2$.}} Recall   $\widehat{\sigma}_{ij}= \mathcal{T}_{\zeta}\breve{\sigma}_{ij}$, let $T_1 = \big[\sum_{i=1}^d|\widehat{\sigma}_{ij}-\sigma^*_{ij}|\mathbbm{1}(\mathscr{A}_{ij}^c)\big]^p$, we have
\begin{equation}
    \begin{aligned}
    &T_1  \leq   \Big[\sum_{i=1}^d  |\sigma^*_{ij}|\mathbbm{1}(\mathscr{A}_{ij}^c)\mathbbm{1}(|\breve{\sigma}_{ij}|<\zeta)+\sum_{i=1}^d  |\breve{\sigma}_{ij}-\mathbbm{E}\breve{\sigma}_{ij}|\mathbbm{1}(\mathscr{A}_{ij}^c)+\sum_{i=1}^d|\mathbbm{E}\breve{\sigma}_{ij}-\sigma^*_{ij}|\mathbbm{1}(\mathscr{A}_{ij}^c)\Big]^p \\
    &\leq  (3d)^{p-1}\Big[\sum_{i=1}^d |\sigma^*_{ij}|^p\mathbbm{1}(\mathscr{A}_{ij}^c)\mathbbm{1}(|\breve{\sigma}_{ij}|<\zeta)+ \sum_{i=1}^d|\breve{\sigma}_{ij}-\mathbbm{E}\breve{\sigma}_{ij}|^p\mathbbm{1}(\mathscr{A}_{ij}^c)+\sum_{i=1}^d |\mathbbm{E}\breve{\sigma}_{ij}-\sigma^*_{ij}|^p\mathbbm{1}(\mathscr{A}_{ij}^c)\Big].
    \nonumber
    \end{aligned}
\end{equation}
Combining with the form of $R_2$ yields
\begin{equation}
    \begin{aligned}
    &R_2  \leq 2^p\mathbbm{E}\sum_{j=1}^d\Big[\sum_{i=1}^d|\widehat{\sigma}_{ij}-\sigma^*_{ij}|\mathbbm{1}(\mathscr{A}_{ij}^c)\Big]^p 
    \leq  6^pd^{p-1}\Big({\mathbbm{E}\sum_{i,j}|\sigma^*_{ij}|^p\mathbbm{1}(\mathscr{A}_{ij}^c)\mathbbm{1}(|\breve{\sigma}_{ij}|<\zeta)} \\ &+   {\mathbbm{E}\sum_{i,j}|\breve{\sigma}_{ij}-\mathbbm{E}\breve{\sigma}_{ij}|^p\mathbbm{1}(\mathscr{A}_{ij}^c)} + {\mathbbm{E}\sum_{i,j}|\mathbbm{E}\breve{\sigma}_{ij}-\sigma^*_{ij}|^p\mathbbm{1}(\mathscr{A}_{ij}^c)} \Big):= 2^pd^{p-1}\big(R_{21}+R_{22}+R_{23}\big).
    \label{A.8}
    \end{aligned}
\end{equation}
Let us deal with $R_{21},R_{22},R_{23}$ separately.

\noindent{\underline{\it Bound of $R_{21}$.}} Suppose the event $\mathscr{A}_{ij}^c\cap \{|\breve{\sigma}_{ij}|<\zeta\}$ holds, then $\widehat{\sigma}_{ij}=0$, combining with (\ref{A.6}) we know $|\widehat{\sigma}_{ij}-\sigma^*_{ij}| = |\sigma^*_{ij}|>D_1 \min\{|\sigma^*_{ij}|,\sigma^2\log n \sqrt{\frac{\delta \log d}{n}}\}.$ Recall (\ref{A.4}), we assume $|\mathbbm{E}\breve{\sigma}_{ij}-\sigma^*_{ij}| \leq D_{1,0}\sigma^2\sqrt{\frac{\delta \log d}{n}}$
for some constant $D_{1,0}$. To avoid technical complication, we simply assume $D_1,C_2$ are sufficiently large and satisfy $D_1 \geq \max\{3C_2,3\}$, $C_2\geq\max\{ D_{1,0},10C_1^2\}$.

Combining with (\ref{2.9}) in Theorem \ref{theorem2}, we have $$|\sigma^*_{ij}|\geq D_1\sigma^2\log n \sqrt{\frac{\delta\log d}{n}} \geq 3\zeta > 3|\breve{\sigma}_{ij}| \geq 3|\sigma^*_{ij}|- 3|\sigma^*_{ij}-\breve{\sigma}_{ij}|,$$ which implies \begin{equation}
\label{713_2}
    |\sigma^*_{ij}-\breve{\sigma}_{ij}| \geq \frac{2}{3}|\sigma^*_{ij}|~~\mathrm{and}~~\zeta \leq \frac{1}{3}|\sigma^*_{ij}|.
\end{equation} 
Moreover, we have 
$$|\mathbbm{E}\breve{\sigma}_{ij}-\sigma^*_{ij}|\leq D_{1,0}\sigma^2 \sqrt{\frac{\delta \log d}{n}}\leq \zeta \leq \frac{1}{3}|\sigma^*_{ij}|.$$ Besides $|\sigma^*_{ij}|\geq 3\zeta$,  based on $  |\sigma^*_{ij}-\breve{\sigma}_{ij}|\geq \frac{2}{3}|\sigma^*_{ij}| $, we use   triangle inequality and obtain   \begin{equation}
\label{713_3}
    \frac{2}{3}|\sigma^*_{ij}|\leq |\sigma^*_{ij}-\breve{\sigma}_{ij}| \leq |\breve{\sigma}_{ij} -\mathbbm{E}\breve{\sigma}_{ij}|+|\mathbbm{E}\breve{\sigma}_{ij}-\sigma^*_{ij}|\leq |\breve{\sigma}_{ij} -\mathbbm{E}\breve{\sigma}_{ij}| + \frac{1}{3}|\sigma^*_{ij}|,
\end{equation}
which implies $|\breve{\sigma}_{ij} -\mathbbm{E}\breve{\sigma}_{ij}|\geq \frac{1}{3}|\sigma^*_{ij}|$. Therefore, we draw the conclusion that$$\mathscr{A}_{ij}^c\cap \{|\breve{\sigma}_{ij}|<\zeta\}\Longrightarrow \{|\sigma^*_{ij}|>3\zeta\}\cap \{|\breve{\sigma}_{ij} -\mathbbm{E}\breve{\sigma}_{ij}|\geq \frac{1}{3}|\sigma^*_{ij}|\}.$$
Now we can invoke Hoeffding's inequality (Proposition \ref{pro3}) and obtain 
\begin{equation}
    \nonumber
    \begin{aligned}
    &R_{21}= \sum_{i,j}|\sigma^*_{ij}|^p\mathbbm{E}\big[\mathbbm{1}(\mathscr{A}_{ij}^c\cap \{|\breve{\sigma}_{ij}|<\zeta\})\big]
     \\ &\leq \sum_{i,j}|\sigma^*_{ij}|^p\mathbbm{1}(|\sigma^*_{ij}|>3\zeta)\mathbbm{P}\Big(|\breve{\sigma}_{ij} -\mathbbm{E}\breve{\sigma}_{ij}|\geq \frac{1}{3}|\sigma^*_{ij}|\Big) \\
    &\leq2\sum_{i,j}|\sigma^*_{ij}|^p\mathbbm{1}(|\sigma^*_{ij}|>3\zeta)\exp\Big(-\frac{n|\sigma^*_{ij}|^2}{18\gamma^4}\Big) \\
    \end{aligned}
\end{equation}
 Moreover, some calculus can verify  $\sup_{y\geq 0}y^{\frac{p}{2}}\exp(-y/36)\leq (D_2)^p(\sqrt{p})^p$. Thus, we proceed as
\begin{equation}
    \begin{aligned}
    &R_{21}\leq2\sum_{i,j}|\sigma^*_{ij}|^p\mathbbm{1}(|\sigma^*_{ij}|>3\zeta)\exp\Big(-\frac{n|\sigma^*_{ij}|^2}{18\gamma^4}\Big) \\
    &=  2\Big(\frac{\gamma^2}{\sqrt{n}}\Big)^p\sum_{i,j}\Big(\Big[\frac{n|\sigma^*_{ij}|^2}{\gamma^4}\Big]^{\frac{p}{2}}\exp\Big[-\frac{n|\sigma^*_{ij}|^2}{36\gamma^4}\Big]\Big)\Big(\mathbbm{1}(|\sigma^*_{ij}|>3\zeta)\exp\Big[-\frac{n|\sigma^*_{ij}|^2}{36\gamma^4}\Big]\Big)\\
    &\leq 2\Big(\frac{\gamma^2}{\sqrt{n}}\Big)^p\Big(\sup_{y\geq 0}y^{\frac{p}{2}}\exp\Big[-\frac{y}{36}\Big]\Big)\Big(d^2\exp\Big[-\frac{n\zeta^2}{4\gamma^4}\Big]\Big),
    \nonumber
    \end{aligned}
\end{equation}
    Recall (\ref{2.6}), (\ref{2.9}) and that we assume $C_2 \geq 10C_1^2$, we have $\zeta \geq 10C_1^2\sigma^2 \log n\sqrt{\frac{\delta\log d}{n}}\geq 10\gamma^2\sqrt{\frac{\delta\log d}{n}},$ which delivers $d^2\exp(-\frac{n\zeta^2}{4\gamma^4})\leq d^{2-25\delta}.$ We now put pieces together and obtain 
\begin{equation}
R_{21}\leq d^{2-25\delta}\Big(  2D_2\gamma^2\sqrt{\frac{p}{n}}\Big)^p \leq   d^{2-25\delta}\Big(D_3\sigma^2 \log n\sqrt{\frac{\delta}{n}}\Big)^p.
    \nonumber
\end{equation}
\noindent{\underline{\it Bound of $R_{22}$.}} By Cauchy-Schwarz inequality we have 
$$R_{22}\leq \sum_{i,j}\sqrt{\mathbbm{E}|\breve{\sigma}_{ij}-\mathbbm{E}\breve{\sigma}_{ij}|^{2p}\mathbbm{P}(\mathscr{A}_{ij}^c)} \leq \sum_{i,j}d^{-\frac{\delta}{2}}\sqrt{2\mathbbm{E}|\breve{\sigma}_{ij}-\mathbbm{E}\breve{\sigma}_{ij}|^{2p}}.$$
Recall $\breve{\sigma}_{ij} = \frac{1}{n}\sum_{k=1}^n \frac{\gamma^2}{2}\big[\dot{X}_{k1,i}\dot{X}_{k2,j} + \dot{X}_{k2,i}\dot{X}_{k1,j}\big]$ with each summand lying between $[-\frac{\gamma^2}{n},\frac{\gamma^2}{n}]$, so by Hoeffding's Lemma (e.g., Lemma 1.8 in \cite{rigollet2015high}), $\breve{\sigma}_{ij}-\mathbbm{E}\breve{\sigma}_{ij}$ is the sum of $n$ independent random variable, and each variable has sub-Gaussian norm scaling $O\big(\frac{\gamma^2}{n}\big)$. Thus, Proposition \ref{pro2} gives $\|\breve{\sigma}_{ij}-\mathbbm{E}\breve{\sigma}_{ij} \|_{\psi_2}= O\big(\frac{\gamma^2}{\sqrt{n}}\big)$. Now we invoke   Proposition \ref{pro1}(b) to obtain
\begin{equation}
R_{22}\lesssim d^{2-\frac{\delta}{2}}\Big(D_{4,0}\gamma^2\sqrt{\frac{p}{n}}\Big)^p \leq d^{2-\frac{\delta}{2}}\Big(D_4\sigma^2\log n\sqrt{\frac{\delta}{n}}\Big)^p.
    \nonumber
\end{equation}
\noindent{\underline{\it Bound of $R_{23}.$}} Note that $|\mathbbm{E}\breve{\sigma}_{ij}-\sigma^*_{ij}|$ is constant, hence we use (\ref{A.4}) and obtain
\begin{equation}
    \begin{aligned}
    R_{23} \leq  |\mathbbm{E}\breve{\sigma}_{ij} &-\sigma^*_{ij}|^p\sum_{i,j} 2d^{-\delta}  \leq  2d^{2-\delta}\Big(D_5 \sigma^2  \sqrt{\frac{\delta\log d}{n}}\Big)^p.
    \nonumber
    \end{aligned}
\end{equation}

Now we are in a position to put everything together. By combining the upper bounds for $R_{2i}, i = 1,2,3$, we have $R_{21}+R_{22}+R_{23} \leq d^{2-\frac{\delta}{2}}\big(D_6\sigma^2\log n\sqrt{\frac{\delta \log d}{n}}\big)^p.$
Substitute it into (\ref{A.8}), recall $p=\frac{\delta}{4}$ and $\delta \geq 4$, we obtain 
$$R_{2}\leq d^{1-\frac{\delta }{4}}\big(6D_6\sigma^2\log n\sqrt{\frac{\delta \log d}{n}}\big)^p\leq \big(6D_6\sigma^2\log n\sqrt{\frac{\delta \log d}{n}}\big)^p.$$
This bound is dominated by the bound of $R_1$ when $\delta \log d(\log n)^2/n$ is sufficiently small (note that conventionally one assumes $s = \Omega(1)$). Thus, there exists absolute constant $D_7$ such that 
\begin{equation}
\mathbbm{E}\|\bm{\widehat{\Sigma}-\Sigma^*}\|_{\mathrm{op}}^p \leq \Big(D_7s\Big[\sigma^2\log n \sqrt{\frac{\delta\log d}{n}}\Big]^{1-q}\Big)^p,
    \nonumber
\end{equation}
which gives (\ref{2.12}). We further invoke Markov inequality:
\begin{equation}
    \begin{aligned}
   & \mathbbm{P}\Big( \|\bm{\widehat{\Sigma}-\Sigma^*}\|_{\mathrm{op}}\geq e^4D_7s\Big[\sigma^2\log n \sqrt{\frac{\delta\log d}{n}}\Big]^{1-q} \Big)\\
   = & \mathbbm{P}\Big( \|\bm{\widehat{\Sigma}-\Sigma^*}\|_{\mathrm{op}}^p\geq \Big(e^4 D_7s\Big[\sigma^2\log n \sqrt{\frac{\delta\log d}{n}}\Big]^{1-q}\Big)^p \Big) 
   \leq   \exp(-4p) =\exp(-\delta),
    \nonumber
    \end{aligned}
\end{equation} (\ref{2.13}) follows. Now the proof is concluded.\hfill $\square$
\subsection{Heavy-tailed Data} 
\noindent{\textbf{Proof of Theorem \ref{theorem4}}.} Since $\gamma > \eta \geq |\widetilde{X}_{k,i}|$, by using Corollary \ref{corollary1} we can "expect out" the independent dithering noises $\Gamma_{k1,i},\Gamma_{k2,j}$, 
\begin{equation}
    \begin{aligned}
    &\mathbbm{E}\breve{\sigma}_{ij} = \mathbbm{E}\big(\gamma^2\cdot\mathrm{sign}(\widetilde{X}_{k,i}+\Gamma_{k1,i})\mathrm{sign}(\widetilde{X}_{k,j}+\Gamma_{k2,j})\big) \\
    = & \mathbbm{E}_{\widetilde{X}_{k,i}\widetilde{X}_{k,j}}\Big(\mathbbm{E}_{\Gamma_{k1,i}}\big[\gamma\cdot \mathrm{sign}(\widetilde{X}_{k,i}+\Gamma_{k1,i})\big]\Big)\Big(\mathbbm{E}_{\Gamma_{k2,j}}\big[\gamma\cdot \mathrm{sign}(\widetilde{X}_{k,j}+\Gamma_{k2,j})\big]\Big)
    =\mathbbm{E} \widetilde{X}_{k,i}\widetilde{X}_{k,j} .
       \nonumber
    \end{aligned}
\end{equation}
Thus, by triangle inequality we have 
\begin{equation}
|\breve{\sigma}_{ij}-\sigma^*_{ij}| \leq {|\breve{\sigma}_{ij}-\mathbbm{E}\breve{\sigma}_{ij}|}  +{|\mathbbm{E}(X_{k,i}X_{k,j}- \widetilde{X}_{k,i}\widetilde{X}_{k,j})|}:=R_1+R_2.
    \label{A.12}
\end{equation}
\noindent{\underline{\it Bound of $R_1$}.} From $\breve{\sigma}_{ij}= \sum_{k=1}^n\frac{\gamma^2}{2n}(\dot{X}_{k1,i}\dot{X}_{k2,j}+\dot{X}_{k1,j}\dot{X}_{k2,i}) $ 
we know $\breve{\sigma}_{ij}$ is mean of n independent random variables lying in $[-\gamma^2,\gamma^2]$, then by Hoeffding's inequality (Proposition \ref{pro3}) and plug in the value of $\gamma$ (\ref{2.16}), we have
\begin{equation}
\label{718add1}
    \mathbbm{P}(R_1 \geq t)\leq 2\exp\Big(-\frac{nt^2}{2\gamma^4}\Big)=2\exp\Big(-\frac{t^2\sqrt{n\delta \log d}}{2C_4^4M}\Big), \ \forall t>0.
\end{equation}
Setting $t = \sqrt{2M}C_4^2 \left(\frac{\delta \log d}{n}\right)^{{1}/{4}}$   yields 
$\mathbbm{P}\big(R_1 \geq \sqrt{2}C_4^2\sqrt{M}\big[\frac{\delta \log d}{n}\big]^{{1}/{4}}\big)\leq 2d^{-\delta}.$

\noindent{\underline{\it Bound of $R_2$}.} Since the truncated version $\widetilde{X}_{k,i}\neq X_{k,i}$ only when $|X_{k,i}|>\eta$, so 
\begin{equation}
    \begin{aligned}
    &R_2 \leq   \mathbbm{E}\Big[|X_{k,i}X_{k,j}- \widetilde{X}_{k,i}\widetilde{X}_{k,j}|(\mathbbm{1}(\{|X_{k,i}|>\eta\}\cup\{|X_{k,j}|>\eta\}))\Big] \\
    &= {\mathbbm{E}\Big[|X_{k,i}X_{k,j}|\mathbbm{1}(|X_{k,i}|>\eta)\Big]}  + {\mathbbm{E}\Big[|X_{k,i}X_{k,j}|\mathbbm{1}(|X_{k,j}|>\eta)\Big]} := R_{21}+R_{22}.
    \nonumber
    \end{aligned}
\end{equation}
By Cauchy-Schwarz inequality, we bound $R_{21}$ by 
$R_{21}\leq \sqrt{\mathbbm{E}|X_{k,i}X_{k,j}|^2\mathbbm{P}(|X_{k,i}|>\eta)}$, moreover, we have 
$\mathbbm{E}|X_{k,i}X_{k,j}|^2\leq \mathbbm{E}(|X_{k,i}|^4+|X_{k,j}|^4)/{2}\leq M$. A direct application of Markov inequality yields that
$\mathbbm{P}(|X_{k,i}|>\eta)\leq \frac{\mathbbm{E}|X_{k,i}|^4}{\eta^4} = \frac{M}{\eta^4}.$
Plug in the above two inequalities and the value of $\eta$, we have $R_{21} \leq \frac{M}{\eta^2} = \frac{\sqrt{M}}{C_3^2}\big(\frac{\delta \log d }{n}\big)^{{1}/{4}}$. Since $R_{22}$ can be bounded likewise, it holds that
\begin{equation}
R_2=|\mathbbm{E}(X_{k,i}X_{k,j}-\widetilde{X}_{k,i}\widetilde{X}_{k,j})|\leq \frac{2}{C_3^2}\sqrt{M}\Big(\frac{\delta \log d}{n}\Big)^{\frac{1}{4}}.
    \label{A.14}
\end{equation}
Now we can put things together and obtain (\ref{2.17}). Moreover, (\ref{2.18}) follows from a union bound, hence the proof is concluded. \hfill $\square$ 

\vspace{3mm}
\noindent{\textbf{Proof of Theorem \ref{theorem5}}.} The proof is parallel to that of Theorem \ref{theorem2}. For some specified $C_3,C_4$, by Theorem \ref{theorem4} there exists an absolute constant $D_1$ such that 
\begin{equation}
    \mathbbm{P}\Big(|\breve{\sigma}_{ij}- \sigma^*_{ij}|\leq D_1\sqrt{M}\Big[\frac{\delta \log d}{n}\Big]^{\frac{1}{4}}\Big)\geq 1-2d^{-\delta}.
    \label{A.15}
\end{equation}
We assume $C_5>D_1$ and first rule out probability $2d^{-\delta}$ in (\ref{A.15}), so we can proceed the proof upon the event $|\breve{\sigma}_{ij}- \sigma^*_{ij}|\leq D_1\sqrt{M}\big[\frac{\delta \log d}{n}\big]^{{1}/{4}}$. According to the threshold $\zeta$ we discuss two cases.

\noindent{\underline{\it Case 1.}} $|\breve{\sigma}_{ij}|<\zeta$, then we have $\widehat{\sigma}_{ij}= 0$, thus, $|\widehat{\sigma}_{ij}-\sigma^*_{ij}| = |\sigma^*_{ij}| \leq |\sigma^*_{ij}|$. Moreover, triangle inequality gives 
$|\sigma^*_{ij}|\leq |\sigma_{ij}^* - \breve{\sigma}_{ij}|+|\breve{\sigma}_{ij}| \leq (D_1+C_5)\sqrt{M}\left(\frac{\delta \log d}{n}\right)^{\frac{1}{4}}, $ so we have $$|\widehat{\sigma}_{ij}-\sigma^*_{ij}| \leq (D_1+C_5+1)\min\Big\{|\sigma^*_{ij}|,\sqrt{M}\Big[\frac{\delta \log d}{n}\Big]^{\frac{1}{4}}\Big\}$$

\noindent{\underline{\it Case 2.}} $|\breve{\sigma}_{ij}|>\zeta$, then we have $\widehat{\sigma}_{ij}= \breve{\sigma}_{ij}$, which leads to $|\widehat{\sigma}_{ij}-\sigma^*_{ij}| = |\breve{\sigma}_{ij}-\sigma^*_{ij}| \leq D_1\sqrt{M}\big[\frac{\delta \log d}{n}\big]^{{1}/{4}}.$
Let us show it can also be bounded by $|\sigma^*_{ij}|$. A reverse triangle inequality gives $$|\sigma^*_{ij}|\geq |\breve{\sigma}_{ij}|-|\breve{\sigma}_{ij}-\sigma^*|>\zeta - |\breve{\sigma}_{ij}-\sigma^*|\geq (C_5 - D_1)\sqrt{M}\big[\frac{\delta \log d}{n}\big]^{{1}/{4}},$$
so we obtain $\sqrt{M}\big[\frac{\delta \log d}{n}\big]^{{1}/{4}}\leq \frac{1}{C_5-D_1}|\sigma^*_{ij}|$. Now we can draw the conclusion that
$$|\widehat{\sigma}_{ij}-\sigma^*_{ij}| \leq (D_1+\frac{D_1}{C_5-D_1})\min\Big\{|\sigma^*_{ij}|,\sqrt{M}\Big[\frac{\delta \log d}{n}\Big]^{\frac{1}{4}}\Big\}.$$
Combining two cases leads to (\ref{2.21}), so we complete the proof. \hfill $\square$

\vspace{3mm}
\noindent{\textbf{Proof of Theorem \ref{theorem6}}.} Since $\eta,\gamma,\zeta$ are specified with some $C_3,C_4,C_5$, by Theorem \ref{theorem5} there exists absolute constant $D_1$ such that (\ref{2.21}) holds. We define the event
\begin{equation}
\mathscr{A}_{ij}= \Big\{|\widehat{\sigma}_{ij}-\sigma^*_{ij}|\leq D_1 \min\{|\sigma^*_{ij}|,\sqrt{M}\big[\frac{\delta \log d}{n}\big]^{\frac{1}{4}}\}\Big\},
    \label{A.16}
\end{equation}
then we have $\mathbbm{P}(\mathscr{A}_{ij}^c) \leq 2d^{-\delta}$ (Here, $\mathscr{A}_{ij}^c$ denotes the complementary event). Now we can divide the operator norm error according to $\mathscr{A}_{ij}$ and $\mathscr{A}_{ij}^c$, it gives
\begin{equation}
    \begin{aligned}
   & \mathbbm{E}\|\bm{\widehat{\Sigma}-\Sigma^*}\|_{\mathrm{op}}^p \leq \mathbbm{E}\Big[\sup_{j\in[d]}\sum_{i=1}^d|\widehat{\sigma}_{ij}-\sigma^*_{ij}|\Big]^p \leq  \mathbbm{E}\sup_{j\in [d]}\Big[\sum_{i=1}^d|\widehat{\sigma}_{ij}-\sigma^*_{ij}|\mathbbm{1}(\mathscr{A}_{ij}) + \sum_{i=1}^d|\widehat{\sigma}_{ij}-\sigma^*_{ij}|\mathbbm{1}(\mathscr{A}_{ij}^c) \Big]^p \\
   &\leq{2^p\mathbbm{E}\sup_{j\in[d]}\Big[\sum_{i=1}^d|\widehat{\sigma}_{ij}-\sigma^*_{ij}|\mathbbm{1}(\mathscr{A}_{ij})\Big]^p}+ {2^p\mathbbm{E}\sup_{j\in[d]}\Big[\sum_{i=1}^d|\widehat{\sigma}_{ij}-\sigma^*_{ij}|\mathbbm{1}(\mathscr{A}_{ij}^c)\Big]^p}:=R_1+R_2.
   \nonumber
    \end{aligned}
\end{equation}
\noindent{\underline{\it Bound of $R_1$.}} By the sparsity (\ref{2.3}) and (\ref{A.16}), for any $j\in [d]$ we have 
\begin{equation}
    \begin{aligned}
    &\sum_{i=1}^d|\widehat{\sigma}_{ij}-\sigma^*_{ij}|\mathbbm{1}(\mathscr{A}_{ij}) \leq \sum_{i=1}^d D_1\min\Big\{|\sigma^*_{ij}|,\sqrt{M}\Big[\frac{\delta \log d}{n}\Big]^{\frac{1}{4}}\Big\} \\
    \leq &\sum_{i=1}^d D_1|\sigma^*_{ij}|^q \Big(\sqrt{M}\Big[\frac{\delta \log d}{n}\Big]^{\frac{1}{4}}\Big)^{1-q} \leq D_1s\Big(\sqrt{M}\Big[\frac{\delta \log d}{n}\Big]^{\frac{1}{4}}\Big)^{1-q}.
       \nonumber
    \end{aligned}
\end{equation}
This leads to \begin{equation}
R_1 \leq \Big(2D_1sM^{(1-q)/2}\big( \frac{\delta \log d}{n}\big)^{(1-q)/4}\Big)^p.
    \nonumber
\end{equation}
\noindent{\underline{\it Bound of $R_2$.}} Let $T_1 = \big[\sum_{i=1}^d|\widehat{\sigma}_{ij}-\sigma^*_{ij}|\mathbbm{1}(\mathscr{A}_{ij}^c)\big]^p$. Recall that $\widehat{\sigma}_{ij}= \mathcal{T}_{\zeta}\breve{\sigma}_{ij}$, under $\mathscr{A}_{ij}^c$ we divide the problem into $\{|\breve{\sigma}_{ij}|<\zeta\}$ and $\{|\breve{\sigma}_{ij}|\geq  \zeta\}$, then   triangle inequality  yields
\begin{equation}
    \begin{aligned}
   & T_1 \leq  \Big[\sum_{i=1}^d |\sigma^*_{ij}|\mathbbm{1}(\mathscr{A}_{ij}^c)\mathbbm{1}(|\breve{\sigma}_{ij}|<\zeta)+ \sum_{i=1}^d|\breve{\sigma}_{ij}-\mathbbm{E}\breve{\sigma}_{ij}|\mathbbm{1}(\mathscr{A}_{ij}^c)+\sum_{i=1}^d |\mathbbm{E}\breve{\sigma}_{ij}-\sigma^*_{ij}|\mathbbm{1}(\mathscr{A}_{ij}^c)\Big]^p \\
    &\leq(3d)^{p-1}\Big[\sum_{i=1}^d |\sigma^*_{ij}|^p\mathbbm{1}(\mathscr{A}_{ij}^c)\mathbbm{1}(|\breve{\sigma}_{ij}|<\zeta)+ \sum_{i=1}^d|\breve{\sigma}_{ij}-\mathbbm{E}\breve{\sigma}_{ij}|^p\mathbbm{1}(\mathscr{A}_{ij}^c)+\sum_{i=1}^d |\mathbbm{E}\breve{\sigma}_{ij}-\sigma^*_{ij}|^p\mathbbm{1}(\mathscr{A}_{ij}^c)\Big].
    \nonumber
    \end{aligned}
\end{equation}
Now we put it into the expression of $R_2$ and obtain 
\begin{equation}
    \begin{aligned}
    &R_2  \leq 2^p\mathbbm{E}\sum_{j=1}^d\Big[\sum_{i=1}^d|\widehat{\sigma}_{ij}-\sigma^*_{ij}|\mathbbm{1}(\mathscr{A}_{ij}^c)\Big]^p 
    \leq  6^pd^{p-1}\Big( {\mathbbm{E}\sum_{i,j}|\sigma^*_{ij}|^p\mathbbm{1}(\mathscr{A}_{ij}^c)\mathbbm{1}(|\breve{\sigma}_{ij}|<\zeta)}  \\ &+  {\mathbbm{E}\sum_{i,j}|\breve{\sigma}_{ij}-\mathbbm{E}\breve{\sigma}_{ij}|^p\mathbbm{1}(\mathscr{A}_{ij}^c)} + {\mathbbm{E}\sum_{i,j}|\mathbbm{E}\breve{\sigma}_{ij}-\sigma^*_{ij}|^p\mathbbm{1}(\mathscr{A}_{ij}^c)} \Big):=6^p d^{p-1}(R_{21}+R_{22}+R_{23}).
    \label{A.18}
    \end{aligned}
\end{equation}
\noindent{\underline{\it Bound of $R_{21}$.}} Suppose the event $\mathscr{A}_{ij}^c\cap \{|\breve{\sigma}_{ij}|<\zeta\}$ holds, then $\widehat{\sigma}_{ij}=0$, thus, (\ref{A.16}) delivers that $|\widehat{\sigma}_{ij}-\sigma^*_{ij}| = |\sigma^*_{ij}|>D_1 \min\{|\sigma^*_{ij}|,\sqrt{M}\big[\frac{\delta \log d}{n}\big]^{1/4}\}.$ With no loss of generality, we assume $D_1 \geq \max\{3C_5,3\}$, and $C_5\geq \max\{{2}/{C_3^2},4C_4^2\}$. Combining with (\ref{2.20}), Theorem \ref{theorem5}, we have $$|\sigma^*_{ij}|\geq D_1\sqrt{M}\big[\frac{\delta \log d}{n}\big]^{{1}/{4}}\geq 3\zeta > 3|\breve{\sigma}_{ij}| \geq 3|\sigma^*_{ij}|- 3|\sigma^*_{ij}-\breve{\sigma}_{ij}|,$$
which implies $|\sigma^*_{ij}-\breve{\sigma}_{ij}| \geq \frac{2}{3}|\sigma^*_{ij}|$ and $\zeta \leq \frac{1}{3}|\sigma^*_{ij}|$. Since $\eta <\gamma$, it always holds that $\mathbbm{E}\breve{\sigma}_{ij}-\sigma^*_{ij} = \mathbbm{E}(X_{k,i}X_{k,j}-\widetilde{X}_{k,i}\widetilde{X}_{k,j}).$ Combining (\ref{A.14}), (\ref{2.20}) gives
$$|\mathbbm{E}\breve{\sigma}_{ij}-\sigma^*_{ij}|\leq \frac{2}{C_3^2}\sqrt{M}\Big[\frac{\delta \log d}{n}\Big]^{{1}/{4}}\leq C_5\sqrt{M}\Big[\frac{\delta \log d}{n}\Big]^{{1}/{4}} \leq \zeta \leq  \frac{1}{3}|\sigma^*_{ij}|.$$
We upper bound $|\sigma^*_{ij}-\breve{\sigma}_{ij}|$ by triangle inequality and have $$\frac{2}{3}|\sigma^*_{ij}|\leq |\sigma^*_{ij}-\breve{\sigma}_{ij}| \leq |\breve{\sigma}_{ij} -\mathbbm{E}\breve{\sigma}_{ij}|+|\mathbbm{E}\breve{\sigma}_{ij}-\sigma^*_{ij}|\leq |\breve{\sigma}_{ij} -\mathbbm{E}\breve{\sigma}_{ij}| + \frac{1}{3}|\sigma^*_{ij}|,$$
which implies $|\breve{\sigma}_{ij} -\mathbbm{E}\breve{\sigma}_{ij}|\geq \frac{1}{3}|\sigma^*_{ij}|$. Therefore, $$\mathscr{A}_{ij}^c\cap \{|\breve{\sigma}_{ij}|<\zeta\}\Longrightarrow \{|\sigma^*_{ij}|>3\zeta\}\cap \{|\breve{\sigma}_{ij} -\mathbbm{E}\breve{\sigma}_{ij}|\geq \frac{1}{3}|\sigma^*_{ij}|\},$$
so we can bound $R_{21}$ via
\begin{equation}
    \begin{aligned}
    &R_{21}= \sum_{i,j}|\sigma^*_{ij}|^p\mathbbm{E}\Big[\mathbbm{1}(\mathscr{A}_{ij}^c\cap \{|\breve{\sigma}_{ij}|<\zeta\})\Big]\\
    &\leq \sum_{i,j}|\sigma^*_{ij}|^p\mathbbm{1}(|\sigma^*_{ij}|>3\zeta)\mathbbm{P}\Big(|\breve{\sigma}_{ij} -\mathbbm{E}\breve{\sigma}_{ij}|\geq \frac{1}{3}|\sigma^*_{ij}|\Big)  \leq 2\sum_{i,j}|\sigma^*_{ij}|^p\mathbbm{1}(|\sigma^*_{ij}|>3\zeta)\exp\Big(-\frac{n|\sigma^*_{ij}|^2}{18\gamma^4}\Big) \\
    &=  2\Big(\frac{\gamma^2}{\sqrt{n}}\Big)^p\sum_{i,j}\Big(\Big[\frac{n|\sigma^*_{ij}|^2}{\gamma^4}\Big]^{\frac{p}{2}}\exp\Big[-\frac{n|\sigma^*_{ij}|^2}{36\gamma^4}\Big]\Big)\Big(\mathbbm{1}(|\sigma^*_{ij}|>3\zeta)\exp\Big[-\frac{n|\sigma^*_{ij}|^2}{36\gamma^4}\Big]\Big)\\
    &\leq  2\Big(\frac{\gamma^2}{\sqrt{n}}\Big)^p\Big(\sup_{y\geq 0}y^{\frac{p}{2}}\exp\Big[-\frac{y}{36}\Big]\Big)\Big(d^2\exp\Big[-\frac{n\zeta^2}{4\gamma^4}\Big]\Big) \leq 2\Big(\frac{\gamma^2}{\sqrt{n}}\Big)^p\Big(\sup_{y\geq 0}y^{\frac{p}{2}}\exp\Big[-\frac{y}{36}\Big]\Big)d^{2-4\delta},
    \nonumber
    \end{aligned}
\end{equation}
where the second inequality is from Hoeffding's inequality (Proposition \ref{pro3}), while we plug in $\gamma,\zeta$ and use $C_5\geq 4C_4^2$ in the last line. Some calculus show $\sup_{y\geq 0}y^{{p}/{2}}\exp(-{y}/{36})\leq D_2^pp^{p/2}.$ Then we plug in the above inequality and the value of $\gamma$ (\ref{2.16}), for some $D_3$ we have
\begin{equation}
R_{21}\leq d^{2-4\delta}\Big(2D_2\gamma^2\sqrt{\frac{p}{n}}\Big)^p \leq  d^{2-4\delta}\Big(D_3\sqrt{M} \Big[\frac{\delta}{n\log d}\Big]^{\frac{1}{4}}\Big)^p
    \nonumber
\end{equation}
\noindent{\underline{\it Bound of $R_{22}$.}} This is the same as the corresponding part in the proof of Theorem \ref{theorem3}. In brief, we can show an upper bound of the same form, but with different value of $\gamma^2$ (given in (\ref{2.16})):  
\begin{equation}
R_{22} \leq d^{2-\frac{\delta}{2}}\Big(D_{4,0}\gamma^2\sqrt{\frac{p}{n}}\Big)^p\leq d^{2-\frac{\delta}{2}} \Big(D_4\sqrt{M}\Big[\frac{\delta}{n\log d}\Big]^{\frac{1}{4}}\Big)^p
    \nonumber
\end{equation}
\noindent{\underline{\it Bound of $R_{23}$.}} Note that $|\mathbbm{E}\breve{\sigma}_{ij}-\sigma^*_{ij}|=|\mathbbm{E}(\widetilde{X}_{k,i}\widetilde{X}_{k,j}-X_{k,i}X_{k,j})|$ is constant which has been bounded in the proof of Theorem \ref{theorem4}. In particular, (\ref{A.14}) gives $|\mathbbm{E}\breve{\sigma}_{ij}-\sigma^*_{ij}|\leq \frac{2}{C_3^2}\sqrt{M}\big[\frac{\delta \log d}{n}\big]^{{1}/{4}}.$
By combining with $\mathbbm{P}(\mathscr{A}_{ij}^c)\leq 2d^{-\delta}$, we bound $R_{23}$ via 
\begin{equation}
    \begin{aligned}
    R_{23} \leq  |\mathbbm{E}\breve{\sigma}_{ij} &-\sigma^*_{ij}|^p\sum_{i,j} 2d^{-\delta}\leq   d^{2-\delta}\Big(D_5\sqrt{M}\Big[\frac{\delta \log d}{n}\Big]^{\frac{1}{4}}\Big)^p,
    \nonumber
    \end{aligned}
\end{equation}
Now we are in a position to put things together. By combining the upper bounds for $R_{2i},i=1,2,3,$ we have 
$R_{21}+R_{22}+R_{23} \leq D_6^pd^{2-\frac{\delta}{2}} M^{p/2}\big[\frac{\delta \log d}{n}\big]^{{p}/{4}}.$ We further substitute it  into (\ref{A.18}), and recall $p=\frac{\delta}{4}$, $\delta \geq 4$, we obtain 
$$R_{2}\leq d^{1-\frac{\delta }{4}}\Big(6D_6\sqrt{M}\Big[\frac{\delta \log d}{n}\Big]^{\frac{1}{4}}\Big)^p\leq \Big(6D_6\sqrt{M}\Big[\frac{\delta \log d}{n}\Big]^{\frac{1}{4}}\Big)^p.$$
When ${\delta\log d}/{n}$ is small enough, this upper bound for $R_2$ is smaller than the obtained bound for $R_1$. Thus, we know there exists absolute constant $D_7$ such that 
$$\mathbbm{E}\|\bm{\widehat{\Sigma}-\Sigma^*}\|_{\mathrm{op}}^p \leq \Big(D_7sM^{(1-q)/2}\Big[\frac{\delta \log d}{n}\Big]^{{(1-q)}/{4}}\Big)^p,$$
(\ref{2.23}) follows. We further use Markov inequality:
\begin{equation}
    \begin{aligned}
   & \mathbbm{P}\Big( \|\bm{\widehat{\Sigma}-\Sigma^*}\|_{\mathrm{op}}\geq e^4D_7sM^{(1-q)/2}\Big[\frac{\delta \log d}{n}\Big]^{{(1-q)}/{4}} \Big)\\
   = & \mathbbm{P}\Big( \|\bm{\widehat{\Sigma}-\Sigma^*}\|_{\mathrm{op}}^p\geq \Big[e^4 D_7sM^{(1-q)/2}\Big[\frac{\delta \log d}{n}\Big]^{{(1-q)}/{4}} \Big]^p\Big) 
   \leq  \exp(-4p) =\exp(-\delta),
    \nonumber
    \end{aligned}
\end{equation} this displays (\ref{2.24}) and concludes the proof. \hfill $\square$
\section{Proofs: Sparse Linear Regression}
\noindent{\textbf{Proof of Lemma \ref{lemma2}}.} The proof is obtained by modifying and combining  Lemma 1 in \cite{negahban2012unified} and Theorem 1 in \cite{fan2021shrinkage}.

\noindent
\textbf{I.} From the detinition of $\bm{\widehat{\Theta}}$ (\ref{3.3}), we have 
\begin{equation}
\mathcal{L}(\bm{\widehat{\Theta}})-\mathcal{L}(\bm{\Theta^*})\leq \lambda\|\bm{\Theta^*}\|_{\mathrm{nu}}-\lambda \|\bm{\widehat{\Theta}}\|_{\mathrm{nu}}.
    \label{B.1}
\end{equation}
By (\ref{3.4}), some algebra delivers that
\begin{equation}
\begin{aligned}
   \mathcal{L}(\bm{\widehat{\Theta}})-&\mathcal{L}(\bm{\Theta^*})= \frac{1}{2}\mathrm{vec}(\bm{\widehat{\Delta}})^T\bm{Q}\mathrm{vec}(\bm{\widehat{\Delta}}) -\left<\bm{B},\bm{\widehat{\Delta}}\right>+\mathrm{vec}(\bm{\Theta^*})^T\bm{Q}\mathrm{vec}(\bm{\widehat{\Delta}}) \\
   = & \frac{1}{2}\mathrm{vec}(\bm{\widehat{\Delta}})^T\bm{Q}\mathrm{vec}(\bm{\widehat{\Delta}}) +\left<\mathrm{mat}(\bm{Q}\cdot\mathrm{vec}(\bm{\Theta^*}))-\bm{B},\bm{\widehat{\Delta}}\right>.
\end{aligned}
    \label{B.2}
\end{equation}
Since $Q$ is positive semi-definite, combining with $\left<\bm{A_1},\bm{A_2}\right>\leq \|\bm{A_1}\|_{\mathrm{op}}\|\bm{A_2}\|_{\mathrm{nu}}$, (\ref{3.5})
\begin{equation}
      \begin{aligned}
      \mathcal{L}(\bm{\widehat{\Theta}})-\mathcal{L}(\bm{\Theta^*}) \geq & -\|\mathrm{mat}(\bm{Q}\cdot\mathrm{vec}(\bm{\Theta^*}))-\bm{B}  \|_{\mathrm{op}} \|\bm{\widehat{\Delta}}\|_{\mathrm{nu}}
      \geq -\frac{\lambda}{2} \|\bm{\widehat{\Delta}}\|_{\mathrm{nu}} .
         \label{B.3}
      \end{aligned}
\end{equation}
\noindent{\textbf{II.}} Consider the SVD $\bm{\Theta^*} = \bm{U}\bm{\Sigma}\bm{V^T}= \begin{bmatrix} \bm{U_1} & \bm{U_2}
\end{bmatrix}\begin{bmatrix}\bm{\Sigma_{11}} & 0 \\ 0 & \bm{\Sigma_{22}}
\end{bmatrix}\begin{bmatrix} \bm{V_1^T} \\ \bm{V_2^T} 
\end{bmatrix},$
where $\bm{U_1} , \bm{V_1}\in \mathbb{R}^{d \times z},\bm{U_2},$ $\bm{V_2} \in \mathbb{R}^{d\times (d-z)}$ is a partition of singular vectors, $z\in \{0,1,...,d\}$ will be specified later. If $z\geq 1$ we consider two linear subspaces of $\mathbb{R}^{d\times d}$ defined as $\mathcal{M} = \{\bm{U_1A_1 V_1^*}: \bm{A_1}\in \mathbb{R}^{z\times z}\}$ and $$\overline{\mathcal{M}}= \Big\{\begin{bmatrix} \bm{U_1} & \bm{U_2}
\end{bmatrix}\begin{bmatrix}\bm{A_{1}} & \bm{A_2} \\ \bm{A_3} & 0
\end{bmatrix}\begin{bmatrix} \bm{V_1^T} \\ \bm{V_2^T} 
\end{bmatrix}: \bm{A_1}\in \mathbb{R}^{z\times z}, \bm{A_2}\in \mathbb{R}^{z\times (d-z)},\bm{A_3}\in \mathbb{R}^{(d-z)\times z}\Big\},$$
then let $\mathcal{P}_{\mathcal{M}}$ and $\mathcal{P}_{\overline{\mathcal{M}}}$ denote the projection onto $\mathcal{M}$ and $\overline{\mathcal{M}}$ respectively. Given a matrix $\bm{\Delta} \in \mathbb{R}^{d\times d}$, assume that 
$\bm{\Delta} =\begin{bmatrix} \bm{U_1} & \bm{U_2}
\end{bmatrix}\begin{bmatrix}\bm{\Delta_{11}} & \bm{\Delta_{12}} \\ \bm{\Delta_{21}} & \bm{\Delta_{22}}
\end{bmatrix}\begin{bmatrix} \bm{V_1^T} \\ \bm{V_2^T} 
\end{bmatrix} ,$ then $\mathcal{P}_{\mathcal{M}}$ and $\mathcal{P}_{\overline{\mathcal{M}}}$ have the explicit form $$\mathcal{P}_{\mathcal{M}} \bm{\Delta} = \bm{U_1\Delta_{11}V_1^T}\ \mathrm{and} \ \mathcal{P}_{\overline{\mathcal{M}}}\bm{\Delta}=\begin{bmatrix} \bm{U_1} & \bm{U_2}
\end{bmatrix}\begin{bmatrix}\bm{\Delta_{11}} & \bm{\Delta_{12}} \\ \bm{\Delta_{21}} & 0
\end{bmatrix}\begin{bmatrix} \bm{V_1^T} \\ \bm{V_2^T} 
\end{bmatrix} .$$
Besides, let $\mathcal{P}_{\mathcal{M}^\bot} \bm{\Delta}  =\bm{\Delta}-  \mathcal{P}_{\mathcal{M}}\bm{\Delta}$, $\mathcal{P}_{\overline{\mathcal{M}}^\bot} \bm{\Delta}  =\bm{\Delta}-  \mathcal{P}_{\overline{\mathcal{M}}}\bm{\Delta}$. Note that the nuclear norm is decomposable \cite{negahban2012unified} with respect to the pair of subspaces $(\mathcal{M},\overline{\mathcal{M}})$ since for any $\bm{\Delta_1,\Delta_2}\in \mathbb{R}^{d\times d}$, it holds that \begin{equation}
    \|\mathcal{P}_{\mathcal{M}}\bm{\Delta_1}+ \mathcal{P}_{\overline{\mathcal{M}}^\bot}\bm{\Delta_2}\|_{\mathrm{nu}} = \|\mathcal{P}_{\mathcal{M}}\bm{\Delta_1}\|_{\mathrm{nu}}+ \|\mathcal{P}_{\overline{\mathcal{M}}^\bot}\bm{\Delta_2}\|_{\mathrm{nu}}.
    \label{B.4}
\end{equation}
By using $\mathcal{P}_{\mathcal{M}}$ and $\mathcal{P}_{\overline{\mathcal{M}}}$, we have $\|\bm{\widehat{\Delta}}\|_{\mathrm{nu}}\leq \|\mathcal{P}_{\overline{\mathcal{M}}}\bm{\widehat{\Delta}}\|_{\mathrm{nu}} + \|\mathcal{P}_{\overline{\mathcal{M}}^\bot}\bm{\widehat{\Delta}}\|_{\mathrm{nu}}$, plug in (\ref{B.3}) and combine with (\ref{B.1}), we obtain 
\begin{equation}
\|\bm{\widehat{\Theta}}\|_{\mathrm{nu}}- \|\bm{\Theta^*}\|_{\mathrm{nu}}\leq \frac{1}{2}\Big[\|\mathcal{P}_{\overline{\mathcal{M}}}\bm{\widehat{\Delta}}\|_{\mathrm{nu}}+ \|\mathcal{P}_{\overline{\mathcal{M}}^\bot}\bm{\widehat{\Delta}}\|_{\mathrm{nu}}\Big].
    \label{B.5}
\end{equation}
In the special case $z=0$, we just let $\mathcal{P}_{\mathcal{M}} = \mathcal{P}_{\overline{\mathcal{M}}} = \bm{0}$, it can be easily verified that (\ref{B.4}), (\ref{B.5}) and what follow still hold.

\noindent{\textbf{III.}} In this part we derive (\ref{3.6}). We calculate that 
\begin{equation}
    \begin{aligned}
        &\|\bm{\widehat{\Theta}}\|_{\mathrm{nu}} - \|\bm{\Theta^*}\|_{\mathrm{nu}} = \|\mathcal{P}_{\mathcal{M}}\bm{\Theta^*}+\mathcal{P}_{\mathcal{M}^\perp}\bm{\Theta^*} +\mathcal{P}_{\overline{\mathcal{M}}}\bm{\widehat{\Delta}}+\mathcal{P}_{\overline{\mathcal{M}}^\perp}\bm{\widehat{\Delta}}\|_{\mathrm{nu}}-\|\mathcal{P}_{\mathcal{M}}\bm{\Theta^*}+\mathcal{P}_{\mathcal{M}^\perp}\bm{\Theta^*} \|_{\mathrm{nu}} \\
        &\geq  \|\mathcal{P}_{\mathcal{M}}\bm{\Theta^*}\|_{\mathrm{nu}} +\|\mathcal{P}_{\overline{\mathcal{M}}^\perp}\bm{\widehat{\Delta}}\|_{\mathrm{nu}}-\|\mathcal{P}_{\mathcal{M}^\perp}\bm{\Theta^*}\|_{\mathrm{nu}}-\|\mathcal{P}_{\overline{\mathcal{M}}}\bm{\widehat{\Delta}}\|_{\mathrm{nu}} -\|\mathcal{P}_{\mathcal{M}}\bm{\Theta^*}\|_{\mathrm{nu}} -\|\mathcal{P}_{\mathcal{M}^\perp}\bm{\Theta^*}\|_{\mathrm{nu}}\\
        &=\|\mathcal{P}_{\overline{\mathcal{M}}^\perp}\bm{\widehat{\Delta}}\|_{\mathrm{nu}}-2\|\mathcal{P}_{\mathcal{M}^\perp}\bm{\Theta^*}\|_{\mathrm{nu}}-\|\mathcal{P}_{\overline{\mathcal{M}}}\bm{\widehat{\Delta}}\|_{\mathrm{nu}},
        \label{B.6}
    \end{aligned}
\end{equation}
note that we use decomposability (\ref{B.4}) and triangle inequality in the third line. By combining (\ref{B.5}), (\ref{B.6}) we obtain $\|\mathcal{P}_{\overline{\mathcal{M}}^\perp}\bm{\widehat{\Delta}}\|_{\mathrm{nu}} \leq 3 \|\mathcal{P}_{\overline{\mathcal{M}}}\bm{\widehat{\Delta}}\|_{\mathrm{nu}} +4\|\mathcal{P}_{\mathcal{M}^\perp}\bm{\Theta^*}\|_{\mathrm{nu}}$, it holds that
\begin{equation}
    \|\bm{\widehat{\Delta}}\|_{\mathrm{nu}}\leq \|\mathcal{P}_{\overline{\mathcal{M}}^\perp}\bm{\widehat{\Delta}}\|_{\mathrm{nu}} +\|\mathcal{P}_{\overline{\mathcal{M}}}\bm{\widehat{\Delta}}\|_{\mathrm{nu}}\leq  4 (\|\mathcal{P}_{\overline{\mathcal{M}}}\bm{\widehat{\Delta}}\|_{\mathrm{nu}} +\|\mathcal{P}_{\mathcal{M}^\perp}\bm{\Theta^*}\|_{\mathrm{nu}}).
    \label{B.7}
\end{equation}
Assume the singular values of $\bm{\Theta^*}$ are $\sigma_1(\bm{\Theta^*})\geq ...\geq \sigma_d(\bm{\Theta^*})$.
Instead of choosing $z$ directly we choose a threshold $\tau >0$ and then let $z= \max\big\{\{0\}\cup \{w\in [d]: \sigma_w(\bm{\Theta^*}) \geq \tau\} \big\}.$
Since $\mathrm{rank}(\mathcal{P}_{\overline{\mathcal{M}}}\bm{\widehat{\Delta}})\leq 2z$, we have $\|\mathcal{P}_{\overline{\mathcal{M}}}\bm{\widehat{\Delta}}\|_{\mathrm{nu}} \leq \sqrt{2z}\|\bm{\widehat{\Delta}}\|_{\mathrm{F}}$. Moreover, by (\ref{3.2}) we have $$z\tau^q \leq \sum_{k=1}^z\sigma_k(\bm{\Theta^*})^q\leq \sum_{k=1}^d\sigma_k(\bm{\Theta^*})^q\leq r,$$ which implies $z\leq r\tau^{-q}$. Therefore, we have $\|\mathcal{P}_{\overline{\mathcal{M}}}\bm{\widehat{\Delta}}\|_{\mathrm{nu}} \leq \sqrt{2r}\tau^{-{q}/{2}}\|\bm{\widehat{\Delta}}\|_{\mathrm{F}}$. By simple algebra we can bound the last term in (\ref{B.7}) by
$$\|\mathcal{P}_{\mathcal{M}^\perp}\bm{\Theta^*}\|_{\mathrm{nu}} = \sum_{k=z+1}^d \sigma_k(\bm{\Theta^*}) = \sum_{k=z+1}^d  \sigma_k(\bm{\Theta^*})^q  \sigma_k(\bm{\Theta^*})^{1-q}\leq r\tau^{1-q}.$$
By putting pieces together, we obtain $$\|\bm{\widehat{\Delta}}\|_{\mathrm{nu}}\leq 4\Big(\sqrt{2r}\tau^{-\frac{q}{2}}\|\bm{\widehat{\Delta}}\|_{\mathrm{F}} + r\tau^{1-q}\Big),\ \forall \tau>0.$$
We only consider $\bm{\widehat{\Delta}}\neq 0$, then we choose $\tau = \Big(\frac{\|\bm{\widehat{\Delta}}\|_{\mathrm{F}}}{\sqrt{r}}\Big)^{{2}/{(2-q)}}$, then we obtain (\ref{3.6}).

\noindent{\textbf{IV.}} Assume we have RSC (\ref{3.7}), we derive the convergence rate. With RSC, from (\ref{B.2}) we have tighter estimation than (\ref{B.3}): 
$$\mathcal{L}(\bm{\widehat{\Theta}})-\mathcal{L}(\bm{\Theta^*}) \geq \frac{1}{2}\kappa \|\bm{\widehat{\Delta}}\|_{\mathrm{F}}^2-\frac{\lambda}{2} \|\bm{\widehat{\Delta}}\|_{\mathrm{nu}}.$$
On the other hand we have $\mathcal{L}(\bm{\widehat{\Theta}})-\mathcal{L}(\bm{\Theta^*}) \leq \lambda \|\bm{\widehat{\Delta}}\|_{\mathrm{nu}} $ from (\ref{B.1}). By combining them we obtain $ \|\bm{\widehat{\Delta}}\|_{\mathrm{nu}}\geq \frac{\kappa}{3\lambda} \|\bm{\widehat{\Delta}}\|_{\mathrm{F}}^2$. Then plug in (\ref{3.6}), the bound for Frobenius norm in (\ref{3.8}) follows. Again plug it into (\ref{3.6}) we obtain the bound for nuclear norm.  \hfill $\square$

\vspace{3mm}
\noindent{\textbf{Proof of Corollary \ref{corollary2}}.} (\ref{3.9}) can be recast as a trace regression $Y_k=\left<\bm{X_{k,tr}},\bm{\Theta_{tr}^*}\right>+\epsilon_k,$
where $\bm{X_{k,tr}}=\mathrm{diag}(X_{k})$, $\bm{\Theta_{tr}^*}= \mathrm{diag}(\Theta^*)$. Consider the convex set $\mathcal{S}=\{\bm{\Theta} \in \mathbb{R}^{d\times d}: \bm{\Theta} = \mathrm{diag}(\Theta), \|\Theta\|_{\max} \leq R\},$ let $\bm{B_{tr}}= \mathrm{diag}(B)$, and $\bm{Q_{tr}}\in \mathbb{R}^{d^2\times d^2}$ is the matrix whose submatrix constituted of the rows and columns with numbering in $\{1,d+2,2d+3,...,d^2\}$ is $\bm{Q}\in \mathbb{R}^{d\times d}$, and the rows and columns not in $\{1,d+2,2d+3,...,d^2\}$ are all zero. Obviously, $\bm{Q_{tr}}$ is positive semi-definite. It is not hard to see that $\widehat{\Theta}$ defined by (\ref{3.11}) is equivalent to finding the diagonal matrix $\bm{\widehat{\Theta}_{tr}}$ via
$$\bm{\widehat{\Theta}_{tr}}\in \mathop{\arg\min}\limits_{\bm{\Theta}\in \mathcal{S}} \ \mathcal{L}(\bm{\Theta}) + \lambda \|\bm{\Theta}\|_{\mathrm{nu}} ,$$
where the loss function is given by $\mathcal{L}(\bm{\Theta}) =  \frac{1}{2}\mathrm{vec}(\bm{\Theta})^T\bm{Q_{tr}}\mathrm{vec}(\bm{\Theta})-\left<\bm{B_{tr}},\bm{\Theta}\right>$, and then let $\widehat{\Theta} $ be the main diagonal of $\bm{\widehat{\Theta}_{tr}}$. Then all the results follow by using Lemma \ref{lemma2}. \hfill $\square$

\subsection{Sub-Gaussian Data}
\vspace{3mm}
\noindent{\textbf{Proof of Theorem \ref{theorem7}}.} To use Corollary \ref{corollary2} we only need to establish (\ref{3.12}), (\ref{3.13}). 

\vspace{1mm}

\noindent{\textbf{I.}} We first show that when $(\log n)^2\log d/n$ is sufficiently small, $\bm{\widehat{\Sigma}}$ is positive definite with high probability. By Assumption \ref{assumption3} and Theorem \ref{theorem3} we have
\begin{equation}
\mathbbm{P}\Big(\|\bm{\widehat{\Sigma}-\Sigma_{XX}}\|_{\mathrm{op}}\leq D_1 \sigma^2 \log n \sqrt{\frac{\delta \log d}{n}}\Big)\geq 1-\exp(-\delta).
    \label{B.8}
\end{equation}
Under sufficiently small $(\log n)^2\log d/n$ we have $\|\bm{\widehat{\Sigma}-\Sigma_{XX}}\|_{\mathrm{op}}\leq \kappa_0$ with probability higher than $1-\exp(-\delta)$. Use $\lambda_{\min}(\cdot)$ to denote the smallest eigenvalue for a symmetric matrix. Combining with $\lambda_{\min}(\bm{\Sigma}_{XX})\geq 2\kappa_0$ in Assumption \ref{assumption3}, we obtain
\begin{equation}
\label{add72_3}
    \lambda_{\min}(\bm{\widehat{\Sigma}}) \geq \lambda_{\min}(\bm{\Sigma}_{XX})-\|\bm{\widehat{\Sigma}}-\bm{\Sigma}_{XX}\|_{\mathrm{op}}\geq \kappa_0,
\end{equation}
which implies that $\bm{\widehat{\Sigma}}$ is positive definite, and (\ref{3.13}) holds.

\noindent{\textbf{II.}} It remains to bound $\|\bm{\widehat{\Sigma}}\Theta^*-\widehat{\Sigma}_{YX}\|_{\max}$ and show $(\ref{3.12})$ holds with high probability. Let $\Sigma_{YX}= \mathbbm{E}Y_kX_k$ and first note that $$ \Sigma_{YX}= \mathbbm{E}(Y_kX_k) = \mathbbm{E}(X_kX_k^T \Theta^* + \epsilon_kX_k) = \mathbbm{E}(X_kX_k^T)\Theta^* = \bm{\Sigma}_{XX}\Theta^*.$$
By repeating the proof of Theorem \ref{theorem1}, we have the element-wise error for $\widehat{\Sigma}_{YX}$\begin{equation}
\mathbbm{P}\Big(\|\widehat{\Sigma}_{YX}-\Sigma_{YX}\|_{\max}\leq D_2 \sigma^2 \log n \sqrt{\frac{\delta \log d}{n}}\Big)\geq 1-2d^{1-\delta}.
    \label{B.9}
\end{equation}
   We now   combine (\ref{B.8}) and (\ref{B.9}), it holds with probability higher than $1-2d^{1-\delta}-\exp(-\delta)$ that
\begin{equation}
\begin{aligned}
 & \|\bm{\widehat{\Sigma}}\Theta^*-\widehat{\Sigma}_{YX}\|_{\max} \leq \|\bm{\widehat{\Sigma}}\Theta^*-\bm{\Sigma}_{XX}\Theta^*\|_{\max} + \|\Sigma_{YX}-\widehat{\Sigma}_{YX}\|_{\max} \\
  \leq & \|\bm{\widehat{\Sigma}}-\bm{\Sigma}_{XX}\|_{\mathrm{op}}\|\Theta^*\|_{2} + \|\Sigma_{YX}-\widehat{\Sigma}_{YX}\|_{\max}
  \leq (D_1R+D_2)\sigma^2 \log n \sqrt{\frac{\delta \log d}{n}}. 
\end{aligned}
    \label{add72_1}
\end{equation}
Thus, we can choose sufficiently large $C_6$ in (\ref{3.20}) such that $C_6 \geq 2(D_1R + D_2)$, then (\ref{3.12}) holds with high probability. Now that (\ref{3.12}) and (\ref{3.13}) have been verified, Corollary \ref{corollary2} gives (\ref{3.14}). We further substitute (\ref{3.20}) into (\ref{3.14}) and conclude the proof. \hfill $\square$

\subsection{Heavy-tailed Data}
 \noindent{\textbf{Proof of Theorem \ref{theorem8}}.} The proof is parallel to Theorem \ref{theorem7}. By Assumption \ref{assumption3} and Theorem \ref{theorem6}, we have the probability tail for operator norm deviation
\begin{equation}
\mathbbm{P}\Big(\|\bm{\widehat{\Sigma}}-\bm{\Sigma}_{XX}\|_{\mathrm{op}}\leq D_1\sqrt{M}\Big[\frac{\delta\log d}{n}\Big]^{1/4}\Big)\geq 1-\exp(-\delta),
    \label{B.10}
\end{equation}
when ${\log d}/{n}$ is sufficiently small, we can assume $\|\bm{\widehat{\Sigma}}-  \bm{\Sigma}_{XX}\|_{\mathrm{op}} \leq \kappa_0  $ with probability higher than $1-\exp(-\delta)$. This, together with  $\lambda_{\min}(\bm{\Sigma}_{XX})\geq 2\kappa_0$ given in Assumption \ref{assumption3}, gives $\lambda_{\min}(\bm{\widehat{\Sigma}})\geq \kappa_0$ under the same probability. Thus, with high probability $\bm{\widehat{\Sigma}}$ is positive definite and (\ref{3.13}) holds.

It remains to establish (\ref{3.12}) and apply Corollary \ref{corollary2}. By repeating the proof of Theorem \ref{theorem4}, we can show the max-norm error for $\widehat{\Sigma}_{YX}$ to approximate $\Sigma_{YX}= \mathbbm{E} Y_kX_k$ as 
\begin{equation}
\mathbbm{P}\Big(\|\widehat{\Sigma}_{YX} - \Sigma_{YX}\|_{\max }  \leq D_2 \sqrt{M}\Big[\frac{\delta \log d}{n}\Big]^{{1}/{4}}\Big)\geq 1-2d^{1-\delta}.
    \label{B.11}
\end{equation}
Now we combine (\ref{B.10}) and (\ref{B.11}), with probability higher than $1-\exp(-\delta)-2d^{1-\delta}$ it yields
\begin{equation}
\begin{aligned}
  &\|\bm{\widehat{\Sigma}}\Theta^*-\widehat{\Sigma}_{YX}\|_{\max} \leq \|\bm{\widehat{\Sigma}}\Theta^*-\bm{\Sigma}_{XX}\Theta^*\|_{\max} + \|\Sigma_{YX}-\widehat{\Sigma}_{YX}\|_{\max} \\
  \leq & \|\bm{\widehat{\Sigma}}-\bm{\Sigma}_{XX}\|_{\mathrm{op}}\|\Theta^*\|_{2} + \|\Sigma_{YX}-\widehat{\Sigma}_{YX}\|_{\max} 
  \leq  (D_1R+D_2)\sqrt{M}\Big[\frac{\delta \log d}{n}\Big]^{{1}/{4}}. 
\end{aligned}
    \label{72add2}
\end{equation}
Thus, in (\ref{3.23}) we can choose sufficiently large $C_7$ such that $C_7 \geq 2(D_2R+D_1)$, then we verify $\lambda \geq 2 \|\bm{\widehat{\Sigma}}\Theta^*-\widehat{\Sigma}_{YX}\|_{\max}$. Now we can use (\ref{3.14}) in Corollary \ref{corollary2} and plug in (\ref{3.23}), the desired error bounds follow. \hfill $\square $

\subsection{1-bit Compressed Sensing}

\textbf{Proof of Theorem \ref{1bitcssg}.} We prove the error bound based on Corollary \ref{corollary2}. Evidently, we need to show setting $\lambda =C_8 \sqrt{\frac{\delta \log d \log n}{n}}$ with sufficiently large $C_8$ can guarantee $\lambda \geq 2\|\bm{\widehat{\Sigma}}_{XX}\Theta^*- \widehat{\Sigma}_{YX}\|_{\max}$. First we use triangle inequality to obtain 
\begin{equation}
\label{72p_1}
    \begin{aligned}
    &\|\bm{\widehat{\Sigma}}_{XX}\Theta^* - \widehat{\Sigma}_{YX}\|_{\max} = \big\|\frac{1}{n}\sum_{k=1}^nX_kX_k^T\Theta^* - \frac{1}{n}\sum_{k=1}^n \gamma\cdot \dot{Y}_kX_k\big\|_{\max}\\
    &= \big\|\frac{1}{n}\sum_{k=1}^n (Y_k-\gamma\cdot \dot{Y}_k)X_k-\frac{1}{n}\sum_{k=1}^n\epsilon_kX_k\big\|_{\max} \\
    &\leq \big\|\frac{1}{n}\sum_{k=1}^n (Y_k-\gamma\cdot \dot{Y}_k)X_k-\mathbbm{E}\big[(Y_k-\gamma\cdot \dot{Y}_k)X_k\big]\big\|_{\max} +\|\mathbbm{E}\big[(Y_k-\gamma\cdot \dot{Y}_k)X_k\big]\|_{\max}\\
    &+ \|\frac{1}{n}\sum_{k=1}^n\epsilon_kX_k\|_{\max} :=R_1+R_2+R_3,
    \end{aligned}
\end{equation}
where we use $Y_k = X_k^T\Theta^* + \epsilon_k$ in the second line.

   \noindent \underline{\it Bound of $R_1$.}  Note that $\|X_k\|_{\psi_2}\leq \sigma_1$, $\|\epsilon_k\|_{\psi_2}\leq \sigma_2$, $\|\Theta^*\|\leq R$ imply $\|Y_k\|_{\psi_2} = O(R\sigma_1+\sigma_2)= O(1)$, (since we assume $\sigma_1,\sigma_2,R$ are absolute constants). Thus, we have $\|Y_k- \gamma\cdot\dot{Y}_k\|_{\psi_2}= O(1+\gamma)$. Denote the $j$-th entry of $X_k$ by $X_{k,j}$, then $X_{k,j} \leq \sigma_1$. Hence, by (\ref{n1.2}) we obtain $\|(Y_k-\gamma\cdot\dot{Y}_k) X_{k,j}\|_{\psi_1}\leq \|Y_k -\gamma\cdot \dot{Y}_k\|_{\psi_2}\|X_{k,j}\|_{\psi_2} = O(1+\gamma)$. Therefore, by Bernstein's inequality   in Proposition \ref{pro4} we obtain \begin{equation}
       \label{713_7}
       \begin{aligned}
       \mathbbm{P}\Big(\big|\frac{1}{n}\sum_{k=1}^n(Y_k-\gamma\cdot\dot{Y}_k)X_{k,j}-&\mathbbm{E}((Y_k-\gamma\cdot\dot{Y}_k)X_{k,j})\big| \geq t\Big)\\&\leq  2\exp\big(-D_1n\min\{\frac{t^2}{(1+\gamma)^2},\frac{t}{1+\gamma}\}\big),~\forall t>0.
       \end{aligned}
   \end{equation}
   Moreover, we use $\gamma = C_8'\sqrt{\log n}$ and a union bound to obtain \begin{equation}
       \mathbbm{P}(R_1\geq t) \leq 2d\exp\big(-D_2n\min\{\frac{t^2}{ \log n},\frac{t}{\sqrt{ \log n}}\}\big).
       \nonumber
   \end{equation}
   Thus, setting $t = D_3\sqrt{\frac{\delta\log n\log d}{n}}$ with large $D_3$ yields $R_1 \leq D_3\sqrt{\frac{\delta \log n\log d}{n}}$ with probability at least $1-2d^{1-\delta}$.

    \noindent \underline{\it Bound of $R_2$.} By Lemma \ref{lemma1} when $|Y_k|\leq \gamma$ we have $\mathbbm{E}_{\Lambda_k}\big(\gamma \cdot\dot{Y}_k\big) = Y_k $. Use this fact and Cauchy-Schwarz inequality, we can first bound $R_2$ from above as 
    \begin{equation}
        \nonumber
        \begin{aligned}
        &R_2 = \big\|\mathbbm{E}\big(Y_k-\gamma\cdot\dot{Y}_k\big)X_k\big\|_{\max} = \big\| \mathbbm{E}\big(Y_k-\gamma\cdot\dot{Y}_k\big)X_k\mathbbm{1}\big(|Y_k|>\gamma\big)\big\|_{\max}\\
        &\leq  \max_{j\in[d]}~ \mathbbm{E}\big(|Y_kX_{k,j}|\mathbbm{1}(|Y_k|>\gamma)\big) \leq \max_{j\in[d]}~\sqrt{\mathbbm{E}\big[|Y_k|^2|X_{k,j}|^2\big]}\sqrt{\mathbbm{P}(|Y_k|>\gamma)}\\
       & \leq  \max_{j\in [d]} \sqrt{\frac{1}{2}\big(\mathbbm{E}|Y_k|^4+\mathbbm{E}|X_{k,j}|^4\big)\mathbbm{P}(|Y_k|>\gamma)} \lesssim  \exp(-D_4\gamma^2)=n^{-n^4(C_8')^2},
        \end{aligned}
    \end{equation}
    where the last inequality follows from $\|Y_k\|_{\psi_2},\|X_{k,j}\|_{\psi_2} = O(1)$, Proposition \ref{pro1} and the choice $\gamma= C_8'\sqrt{\log n}$. Thus, as long as $C_8'$ is sufficiently large, we have $R_2 = O\big(\sqrt{\frac{\delta \log n\log d}{n}}\big)$.

     \noindent \underline{\it Bound of $R_3$.} For $j\in[d]$, by (\ref{n1.2}) it is evident that $\|\epsilon_kX_{k,j}\|_{\psi_1}\leq \|\epsilon_k\|_{\psi_2}\|X_{k,j}\|_{\psi_2}\leq \sigma_1\sigma_2 = O(1)$. Thus,  Bernstein's inequality (Proposition \ref{pro4}) followed by a union bound gives \begin{equation}
         \label{713_6}
         \mathbbm{P}\big(R_3\geq t\big)\leq 2d\exp\big(-D_{4,1}n \min\{t^2,t\}\big),~\forall~t>0.
     \end{equation}
       We further set $t =   \sqrt{\frac{\delta \log d}{D_{4,1} n }}$ and obtain 
     $$ \mathbbm{P}\Big(R_3 \leq \sqrt{\frac{\delta \log d}{D_{4,1} n }}\Big)\geq 1- 2d^{1-\delta}.$$
     By (\ref{2.6}) we can assume $\sigma < \gamma \lesssim \sigma \sqrt{\log n} $. Thus, by (\ref{72p_1}) and the upper bounds for $R_1,R_2,R_3$, with probability higher than $1-4d^{2-\delta}$ we have $$\|\bm{\widehat{\Sigma}}_{XX}\Theta^* - \widehat{\Sigma}_{YX}\|_{\max}\lesssim    \sqrt{\frac{\delta \log n \log d}{n}}.$$
    Therefore, we can choose sufficiently large $C_8$ in (\ref{add72_6})  to guarantee $\lambda \geq 2\| \bm{\widehat{\Sigma}}_{XX}\Theta^* - \widehat{\Sigma}_{YX}\|_{\max}$ holds with high probability. By Corollary \ref{corollary2}, it already leads to (\ref{72add_4}), a relation that facilitates the following discussions.

     Now that (\ref{3.12}) has been verified, we turn to consider the RSC (\ref{3.13}). When $\frac{\delta \log d}{n}$ is sufficiently small, combining with $\lambda_{\min}(\bm{\Sigma}_{XX})\geq 2\kappa_0$,  Lemma 2(a) in \cite{fan2021shrinkage} gives 
     \begin{equation}
         \nonumber
         \mathbbm{P}\Big(\widehat{\Delta}^T\bm{\widehat{\Sigma}}_{XX} \widehat{\Delta} \geq \kappa_0\|\widehat{\Delta}\|_2^2 - \frac{D_5\delta \log d}{n}\|\widehat{\Delta}\|_1^2\Big)\geq 1- 3d^{1-\delta}.
     \end{equation}
       This event, together with (\ref{72add_4}), implies \begin{equation}
         \label{73add_4}
         \widehat{\Delta}^T\bm{\widehat{\Sigma}}_{XX}\widehat{\Delta} \geq \kappa_0\|\widehat{\Delta}\|_2^2 - D_6\cdot \frac{\delta \log d}{n}\cdot s^{\frac{2}{2-q}}\|\widehat{\Delta}\|_2^{\frac{4-4q}{2-q}}.
     \end{equation}
     We proceed the proof upon the condition (\ref{73add_4}) and divide it into the following two cases.
     
     \vspace{2mm}

     \noindent{\underline{\it Case 1.}} If $D_6\cdot \frac{\delta \log d}{n}\cdot s^{\frac{2}{2-q}}\|\widehat{\Delta}\|_2^{\frac{4-4q}{2-q}}\leq \frac{\kappa_0}{2}\|\widehat{\Delta}\|_2^2$, (\ref{73add_4}) gives the RSC (\ref{3.13}) with $\kappa = \frac{\kappa_0}{2}$. Thus, we can invoke (\ref{3.14}) in Corollary \ref{corollary2} and then plug in the value of $\lambda$ in (\ref{add72_6}). This displays the desired error bounds.
     
     \vspace{1mm}
     
     \noindent{\underline{\it Case 2.}}  Otherwise, it holds that \begin{equation}
     \label{73_add6}
         D_6\cdot \frac{\delta \log d}{n}\cdot s^{\frac{2}{2-q}}\|\widehat{\Delta}\|_2^{\frac{4-4q}{2-q}}\geq \frac{\kappa_0}{2}\|\widehat{\Delta}\|_2^2.
     \end{equation}  With no loss of generality, we assume $\widehat{\Delta}\neq 0$. Under the scaling that $\sqrt{s}\big(\sqrt{\frac{\delta \log d}{n}}\big)^{1-q/2}$ is sufficiently small we have $q \in (0,1)$ (Since when $q=0$, $D_6 \cdot \frac{\delta \log d}{n}\cdot s^{\frac{2}{2-q}} <\frac{\kappa_0}{2}$ together with (\ref{73_add6}) gives $\widehat{\Delta}=0$). Again use sufficiently small $\sqrt{s}\big(\sqrt{\frac{\delta \log d}{n}}\big)^{1-q/2}$, (\ref{73_add6}) delivers
     $$ \| \widehat{\Delta}\|_2 \lesssim \Big[\sqrt{s}\Big(\sqrt{\frac{\delta \log d}{n}}\Big)^{1-\frac{q}{2}}\Big]^{\frac{2}{q}}\leq \sqrt{s}\Big(\sqrt{\frac{\delta \log d}{n}}\Big)^{1-\frac{q}{2}}.$$
     This, together with (\ref{72add_4}), gives the upper bound for $\|\widehat{\Delta}\|_1$ as 
     $$ \|\widehat{\Delta}\|_1\lesssim s \Big(\sqrt{\frac{\delta \log d}{n}}\Big)^{1-q}.$$
     Thus, we conclude the proof. \hfill $\square$
     
     \vspace{2mm}
     \noindent
     \textbf{Proof of Theorem \ref{1bitcsht}.} The proof is still based on Corollary \ref{corollary2}. First let us verify the crucial relation  $\lambda \geq 2\| \bm{\widehat{\Sigma}}_{\tilde{X}\tilde{X}}\Theta^* - \widehat{\Sigma}_{YX}\|_{\max}$. Note that $ \bm{\Sigma}_{XX}\Theta^*= \mathbbm{E}Y_kX_k$, by triangle inequality we can divide it into three terms $R_i, ~1\leq i\leq 3$  
     \begin{equation}
         \label{73p_7}
         \begin{aligned}
         &\| \bm{\widehat{\Sigma}}_{\tilde{X}\tilde{X}}\Theta^* - \widehat{\Sigma}_{YX}\|_{\max} \leq \big\|\big(\bm{\widehat{\Sigma}}_{\tilde{X}\tilde{X}}- \bm{\Sigma}_{XX}\big)\Theta^*\big\|_{\max}+ \big\|\mathbbm{E}\big(\gamma\cdot\dot{Y}_k\widetilde{X}_k - Y_kX_k\big)\big\|_{\max} \\ &+ \big\|\frac{1}{n}\sum_{k=1}^n \gamma\cdot\dot{Y}_k\widetilde{X}_k - \mathbbm{E}\big(\gamma\cdot\dot{Y}_k\widetilde{X}_k\big)\big\|_{\max} : = R_1+R_2+R_3.
         \end{aligned}
     \end{equation}
    
    \noindent
    \underline{\it Bound of $R_1$.} We first decompose   $\|\bm{\widehat{\Sigma}}_{\tilde{X}\tilde{X}}-\bm{\Sigma}_{XX}\|_{\max}$ as $$\|\bm{\widehat{\Sigma}}_{\tilde{X}\tilde{X}}-\bm{\Sigma}_{XX}\|_{\max}\leq \big\| \bm{\widehat{\Sigma}}_{\tilde{X}\tilde{X}}-\mathbbm{E}\widetilde{X}_k\widetilde{X}_k^T\big\|_{\max} + \big\| \mathbbm{E}\big(X_kX_k^T-\widetilde{X}_k\widetilde{X}_k^T\big)\big\|_{\max}:= R_{11}+R_{12}.$$
    Let us deal with them element-wisely. For $R_{11}$ and any $(i,j)\in [d]\times [d]$, recall that the truncated covariate satisfies $|\widetilde{X}_{k,i}|\leq \eta_X$, combining with (\ref{3.22}) it gives \begin{equation}
        \nonumber
       \begin{cases}  \displaystyle
           \sum_{k=1}^n \mathbbm{E}\big(\widetilde{X}_{k,i}\widetilde{X}_{k,j}\big)^2\leq \sum_{k=1}^n\mathbbm{E} X_{k,i}^2X_{k,j}^2 \leq \sum_{k=1}^n \frac{1}{2}\big(\mathbbm{E}X_{k,i}^4 + \mathbbm{E}X_{k,j}^4 \big)\leq nM\\  \displaystyle
           \sum_{k=1}^n \mathbbm{E}(\widetilde{X}_{k,i}\widetilde{X}_{k,j})_+^q \leq \sum_{k=1}^n \mathbbm{E}|\widetilde{X}_{k,i}\widetilde{X}_{k,j}|^q\leq (\eta_X^2)^{q-2}\sum_{k=1}^n\mathbbm{E}(\widetilde{X}_{k,i}\widetilde{X}_{k,j})^2\leq nM\cdot (\eta_X^2)^{q-2},\forall q\geq 3
        \end{cases}.
    \end{equation}
    Thus, by the version of Bernstein's inequality given in Theorem 2.10 in \cite{boucheron2013concentration}, we obtain 
    \begin{equation}
        \nonumber
        \mathbbm{P}\Big(\big|\frac{1}{n}\sum_{k=1}^n\widetilde{X}_{k,i}\widetilde{X}_{k,j}-\mathbbm{E}\widetilde{X}_{k,i}\widetilde{X}_{k,j}\big|> \sqrt{\frac{2Mt}{n}}+\frac{\eta_X^2t}{n}\Big)\leq \exp(-t),~\forall~t>0. 
    \end{equation}
     Moreover, we can use an union bound and get 
     \begin{equation}
         \nonumber
         \mathbbm{P}\Big(R_{11}>\sqrt{\frac{2Mt}{n}}+\frac{\eta_X^2t}{n}\Big)\leq d^2\cdot\exp(-t),~\forall~t>0.
     \end{equation}
     Thus, we set $t = \delta \log d$ and plug in $\eta_X\asymp \big(\frac{n}{\log d}\big)^{1/4}$, then with probability at least $1-2d^{2-\delta}$ we have $R_{11}\lesssim \sqrt{\frac{\delta \log d}{n}}$. We now turn to $R_{12}$ and have the $(i,j)$-th entry bounded by 
     \begin{equation}
         \nonumber
         \begin{aligned}
        \big| \mathbbm{E} \big(X_{k,i}X_{k,i}- \widetilde{X}_{k,i}\widetilde{X}_{k,j}\big)\big| \leq \mathbbm{E}|X_{k,i}X_{k,j}|\big(\mathbbm{1}(|X_{k,i}|>\eta_X)+\mathbbm{1}(|X_{k,j}|>\eta_X)\big) . 
         \end{aligned}
     \end{equation}
     The two terms can be bounded  likewise, so we only deal with one of them by Cauchy-Schwarz inequality and (\ref{3.22}): \begin{equation}
         \begin{aligned}
         \nonumber
         & \mathbbm{E}|X_{k,i}X_{k,j}|\mathbbm{1}(|X_{k,i}|>\eta_X)\leq \sqrt{\mathbbm{E}|X_{k,i}X_{k,j}|^2}\sqrt{\mathbbm{P}(|X_{k,i}|>\eta_X)}\\ 
     &\leq  \sqrt{\frac{1}{2}\big(\mathbbm{E}|X_{k,i}|^4+\mathbbm{E}|X_{k,j}|^4\big)}\sqrt{\frac{\mathbbm{E}|X_{k,i}|^4}{\eta_X^4}}\leq\frac{M}{\eta_X^2}\lesssim \sqrt{\frac{\delta \log d}{n}}.
         \end{aligned}
     \end{equation}
     Therefore, with high probability we have $$R_1 \leq\|\bm{\widehat{\Sigma}}_{\tilde{X}\tilde{X}}-\bm{\Sigma}_{XX}\|_{\max}\|\Theta^*\|_1\lesssim (R_{11}+R_{12})\lesssim\sqrt{\frac{\delta \log d}{n}}.$$

     \noindent
    \underline{\it Bound of $R_2$.} We consider the $j$-th entry. Note that $\gamma>\eta_Y$, Lemma \ref{lemma1} gives 
    \begin{equation}
        \nonumber
        | \mathbbm{E}(\gamma\cdot\dot{Y}_k\widetilde{X}_{k,j}-Y_kX_{k,j})| = |\mathbbm{E}(\widetilde{Y}_k\widetilde{X}_{k,j}-Y_kX_{k,j})| \leq \mathbbm{E}\big(|Y_kX_{k,j}|\mathbbm{1}(|Y_k|>\eta_Y)+\mathbbm{1}(|X_{k,j}|>\eta_X)\big).
    \end{equation}
     By Cauchy-Schwarz inequality, (\ref{3.22}) and the value of $\eta_Y$, we obtain 
     $$\mathbbm{E}|Y_kX_{k,j}|\mathbbm{1}(|Y_k|>\eta_Y)\leq \sqrt{\mathbbm{E}|Y_kX_{k,j}|^2\cdot \mathbbm{P}(|Y_k|>\eta_Y)} \leq \frac{M}{\eta_Y^2}\lesssim \Big(\frac{\delta \log d}{n}\Big)^{\frac{1}{3}}.$$
     Similarly, it holds that $\mathbbm{E}|Y_kX_{k,j}|\mathbbm{1}(|X_{k,j}>\eta_X)\leq \frac{M}{\eta_X^2}\lesssim \sqrt{\frac{\delta \log d}{n}}$. Since this is valid for any $j\in [d]$, we obtain $R_2 \lesssim \big(\frac{\delta \log d}{n}\big)^{1/3}.$
     
     \vspace{1mm}
     
    \noindent
    \underline{\it Bound of $R_3$.} We consider the $j$-th entry first. Recall that $|\widetilde{X}_{k,j}|\leq \eta_X$, and by (\ref{3.22}) we know $\mathbbm{E}\widetilde{X}_{k,j}^2 \leq \mathbbm{E}|X_{k,j}|^2\leq \sqrt{M}$, thus we have 
    \begin{equation}
        \label{718_add2}
        \begin{cases}  \displaystyle
           \sum_{k=1}^n \mathbbm{E}\big(\gamma\cdot\dot{Y}_k \widetilde{X}_{k,j}\big)^2 = \gamma^2\sum_{k=1}^n\mathbbm{E}\widetilde{X}_{k,j}^2 \leq n\sqrt{M}\gamma^2 \\  \displaystyle
           \sum_{k=1}^n \mathbbm{E}(\gamma\cdot\dot{Y}_k \widetilde{X}_{k,j})_+^q \leq \gamma^q \sum_{k=1}^n\mathbbm{E} |\widetilde{X}_{k,j}|^q \leq n\sqrt{M}\gamma^2 (\gamma\cdot\eta_X)^{q-2},~\forall~q\geq 3.
        \end{cases}
    \end{equation}
     Now, we can invoke the Bernstein's inequality given in Theorem 2.10 in \cite{boucheron2013concentration} and obtain 
     \begin{equation}
         \nonumber
         \mathbbm{P}\Big(\big|\frac{1}{n}\sum_{k=1}^n\gamma\cdot\dot{Y}_k\widetilde{X}_{k,j}-\mathbbm{E}\gamma\cdot\dot{Y}_k\widetilde{X}_{k,j}\big|>\gamma\sqrt{\frac{2\sqrt{M}t}{n}}+ \frac{\gamma\cdot\eta_Xt}{n}\Big)\leq \exp(-t),~\forall~t>0.  
     \end{equation}
     Thus, for some absolute constant hidden behind "$\gtrsim$", a union bound gives 
     \begin{equation}
         \label{718_add3}
         \mathbbm{P} \Big(R_3\gtrsim \gamma\sqrt{\frac{t}{n}}+\frac{\gamma\cdot \eta_X\cdot t}{n}\Big) \leq d\cdot\exp(-t),~\forall~t>0.
     \end{equation}
     We set $t = \delta \log d$ and plug in our choices $\eta_X\asymp \big(\frac{n}{\delta \log d}\big)^{1/4}$ and $\gamma \asymp \big(\frac{n}{\delta \log d}\big)^{1/6}$, it yields that $R_3\lesssim \big(\frac{\delta \log d}{n}\big)^{1/3}$ holds with probability at least $1-d^{1-\delta}$.

     Now combining the upper bounds for $R_i,1\leq i\leq 3$ and (\ref{73p_7}), we can choose $\lambda =C_{12} \big(\frac{\delta \log d}{n}\big)^{1/3}$ with sufficiently large $C_{12}$ to guarantee $\lambda \geq 2\| \bm{\widehat{\Sigma}}_{\tilde{X}\tilde{X}}\Theta^* - \widehat{\Sigma}_{YX}\|_{\max}$. Note that $ \bm{\Sigma}_{XX}\Theta^*= \mathbbm{E}Y_kX_k$. By Corollary \ref{corollary2} under the same probability we have (\ref{72add_4}), i.e., $\|\widehat{\Delta}\|_1\leq 10s^{\frac{1}{2-q}}\|\widehat{\Delta}\|_2^{\frac{2-2q}{2-q}}.$

     To invoke Corollary \ref{corollary2} we still need to establish the RSC (\ref{3.13}). Note that   our choice of the truncation parameter $\eta_X$ is the same as \cite{fan2021shrinkage}, so we can use  Lemma 2(b) therein\footnote{This result is presented with the probability term reversed in different versions of \cite{fan2021shrinkage}, but the proof therein is find and can yield what we need here.}. Combining with $\lambda_{\min}(\bm{\Sigma}_{XX})\geq 2\kappa_0$ and  (\ref{72add_4}), it gives 
     \begin{equation}
     \nonumber
         \mathbbm{P}\Big(\widehat{\Delta}^T\bm{\widehat{\Sigma}}_{\tilde{X}\tilde{X}}\widehat{\Delta} \geq 2\kappa_0\|\widehat{\Delta}\|_2^2 - D_1 \sqrt{\frac{\delta \log d}{n}} s^{\frac{2}{2-q}}\|\widehat{\Delta}\|_2^{\frac{4-4q}{2-q}}  \Big) \geq 1- d^{2-\sqrt{\delta}}.
     \end{equation}
     We assume the above event holds, and divide the discussion into two cases. 
     
     \vspace{1mm}

     \noindent{\underline{\it Cases 1.}} If $D_1 \sqrt{\frac{\delta \log d}{n}} s^{\frac{2}{2-q}}\|\widehat{\Delta}\|_2^{\frac{4-4q}{2-q}} \leq \kappa_0 \|\widehat{\Delta}\|_2^2$, we have $\widehat{\Delta}^T\bm{\widehat{\Sigma}}_{\tilde{X}\tilde{X}}\widehat{\Delta} \geq \kappa_0\|\widehat{\Delta}\|_2^2$, thus confirming the RSC (\ref{3.13}). Therefore, we can use   (\ref{3.14}) in Corollary \ref{corollary2} and plug in $\lambda \asymp \big(\frac{\delta \log d}{n}\big)^{1/3}$ to yield the error bound    $  \|\widehat{\Delta}\|_2 \lesssim \sqrt{s}\Big(\frac{\delta \log d}{n}\Big)^{(1-\frac{q}{2})/3}:=B_1$.

     \noindent{\underline{\it Cases 2.}} Otherwise, we assume \begin{equation}
         \label{75_1}
         D_1 \sqrt{\frac{\delta \log d}{n}} s^{\frac{2}{2-q}}\|\widehat{\Delta}\|_2^{\frac{4-4q}{2-q}} > \kappa_0 \|\widehat{\Delta}\|_2^2.
     \end{equation} With no loss of generality we assume $\widehat{\Delta}\neq 0$. If $q=0$, \textcolor{black}{under the scaling that $s\big(\sqrt{\frac{\delta \log d}{n}}\big)^{1-q/2}$ is sufficiently small,} (\ref{75_1}) can imply $\widehat{\Delta} = 0$. Thus, we assume $q\in (0,1)$ without losing generality, then (\ref{75_1}) gives
     $$ \|\widehat{\Delta}\|_2  \lesssim \Big[s \Big(\sqrt{\frac{\delta \log d}{n}}\Big)^{1-q/2}\Big]^{1/q} = s^{\frac{1}{q}}\Big(\frac{\delta \log d}{n}\Big)^{\frac{1}{2q}-\frac{1}{4}}:=B_2 .$$
     
     Therefore, we obtain $\|\widehat{\Delta}\|_2  \lesssim \max\{B_1,B_2\} = B_1 \max\big\{ 1, \frac{B_2}{B_1}\big\}$. Because we have assumed the additional $s\big(\frac{\delta \log d}{n}\big)^{\frac{1}{2}-\frac{q}{3}}=O(1)$ for $q\in (0,1)$, it leads to      
     \begin{equation}
         \begin{aligned}
         \nonumber
         \frac{B_2}{B_1} = \Big(s\big(\frac{\delta \log d}{n}\big)^{\frac{1}{2}-\frac{q}{3}}\Big)^{\frac{1}{q}-\frac{1}{2}}=O(1).
         \end{aligned}
     \end{equation}
     Therefore, we arrive at the desired upper bound $\|\widehat{\Delta}\|_2 \lesssim \sqrt{s}\Big(\frac{\delta \log d}{n}\Big)^{{(1-{q}/{2})}/{3}}.$ Combining with (\ref{72add_4}), the bound for $\|\widehat{\Delta}\|_1$  follows. \hfill $\square$

     \vspace{2mm}
     \noindent
     \textbf{Proof of Theorem \ref{lower2}.} We simply write $\mathscr{K}(s,R)$ as $\mathscr{K}$ in this proof, and we use the shorthand $\Sigma_s$ to denote the set of $s$-sparse vectors in $\mathbb{R}^d$.

     We first use the sparse Varshamov-Gilbert (e.g., \cite[Lemma 4.14]{rigollet2015high}) to construct a packing set $\mathscr{K}_0:=\{\Theta^{(1)}_0,...,\Theta^{(N)}_0\}\subset \Sigma_s$ such that \begin{itemize}
         \item for any  $i\in [N]$,   ${\Theta}^{(i)}_0$ has $s$ non-zero entries that equal 1;
         
         \item $\log N \geq \frac{s}{8}\log \frac{d}{2s}$;
         
         \item for $i\neq j$, $\Theta^{(i)}_0$ and $\Theta^{(j)}_0$ contain at least $\frac{s}{2}$ different entries. 
     \end{itemize}
     Then we let $\hat{\alpha} = D_1\gamma \sqrt{\frac{ \log \frac{d}{2s}}{nK_u}}$ for some constant $D_1$, and recall that we choose $\gamma\asymp \big(\frac{n}{\log d}\big)^{1/6}$.  We consider the  set of parameters $\mathscr{K}_1=\hat{\alpha}\mathscr{K}_0=\{{\Theta}^{(i)}= \hat{\alpha}{\Theta}^{(i)}_0:i\in [N]\}$. Because we assume $n\gtrsim K_u^{-1}\big(\frac{s}{R}\big)^3\log\frac{d}{2s}$, we have $\|\Theta^{(i)}\|_1=s\hat{\alpha} \leq R$, and so $\mathscr{K}_1 \subset \mathscr{K}$.

      For $a\in \mathbb{R}$ we let $\mathsf{T}_\eta(a) = \sign(a)\min\{|a|,\eta\}$. It suffices to consider the noiseless case $\epsilon_k = 0$, and hence for underlying matrix $\Theta$ we have the observations$$\mathscr{P}(\Theta)=\{\dot{Y}_k=\sign(\mathsf{T}_{\eta_Y}(X_k^T\Theta)+\Lambda_k),k=1,2,...,n\}.$$
      Note that for any $i\neq j$   $$\frac{s}{2}\hat{\alpha}^2\leq \|\Theta^{(i)}-\Theta^{(j)}\|_2^2\leq 2s \hat{\alpha}^2.$$
      By reduction to hypothesis testing and Fano's inequality (e.g., \cite[Section 4]{rigollet2015high}), we have \begin{equation}\label{fano2}
          \inf_{\widehat{\Theta}}\sup_{\Theta\in \mathscr{K}}\mathbbm{P}_{\Theta}\Big(\|\widehat{\Theta}-\Theta\|_2^2>\frac{s}{8}\hat{\alpha}^2\Big) \geq 1-\frac{\frac{1}{N^2}\sum_{i,j=1}^N\mathsf{KL}(\mathscr{P}(\Theta^{(i)}),\mathscr{P}(\Theta^{(j)}))+\log 2}{\log N}.
      \end{equation}
     Then we estimate $\mathsf{KL}(\mathscr{P}(\Theta^{(i)}),\mathscr{P}(\Theta^{(j)}))$. Let $\Theta^{(i)}$ be the underlying parameter, then because $\eta_Y<\frac{8}{9}\gamma$,     the corresponding $\dot{Y}_k$ follows a (symmetrized) Bernoulli distribution with success probability $$\mathbbm{P}(\dot{Y}_k = 1) = \mathbbm{P}\big(\Lambda_k > -\mathsf{T}_{\eta_Y}(X_k^T\Theta)\big) = \frac{\gamma+\mathsf{T}_{\eta_Y}(X_k^T\Theta)}{2\gamma}\in\big[\frac{1}{18},\frac{17}{18}\big].$$
     We use $\mathsf{KL}(p,q)$ to denote the KL divergence between Bernoulli distribution with success probability $p$ and $q$, then \cite[Lemma A.4]{davenport20141} provides $\mathsf{KL}(p,q)\leq \frac{(p-q)^2}{q(1-q)}$. Because $\{\dot{Y}_k:k=1,...,n\}$ are independent, we have \begin{equation}
         \begin{aligned}\nonumber
             &\mathsf{KL}(\mathscr{P}(\Theta^{(i)}),\mathscr{P}(\Theta^{(j)})) = \sum_{k=1}^n \mathsf{KL}\Big(\frac{1}{2}+\frac{\mathsf{T}_{\eta_Y}(X_k^T\Theta^{(i)})}{2\gamma},\frac{1}{2}+\frac{\mathsf{T}_{\eta_Y}(X_k^T\Theta^{(j)})}{2\gamma}\Big)\\
             &\leq \sum_{k=1}^n \frac{D_2}{\gamma^2}\Big(\mathsf{T}_{\eta_Y}(X_k^T\Theta^{(i)})-\mathsf{T}_{\eta_Y}(X_k^T\Theta^{(j)})\Big)^2\leq \frac{D_2}{\gamma^2}\sum_{k=1}^n \Big(X_k^T(\Theta^{(i)}-\Theta^{(j)})\Big)^2\\&\leq \frac{D_2K_un}{\gamma^2}\big\|\Theta^{(i)}-\Theta^{(j)}\big\|_2^2\leq 2D_2s \frac{K_un\hat{\alpha}^2}{\gamma^2}=2D_2D_1^2s\log \frac{d}{2s}.
         \end{aligned}
     \end{equation} 
     Note that in the first inequality we use $\mathsf{KL}(p,q)\leq \frac{(p-q)^2}{q(1-q)}$, and because $q$ is bounded away from 0 and 1, $\frac{1}{q(1-q)}$ is bounded by absolute constant, then in the following inequalities we use the assumption $\sum_{k=1}^n|X_k^TV|^2\leq n K_u \|V\|^2_2$ for $V\in \Sigma_{2s}$, finally we plug in our choice of $\hat{\alpha}$. Since the estimate is valid for any $i,j$, and $\log N\geq \frac{s}{8}\log \frac{d}{2s}$, so we can set $D_1$ sufficiently small so that under relatively large $s$, the right hand side of (\ref{fano2}) is greater than $\frac{3}{4}$. Also, we perform some algebra to arrive at $$\sqrt{s}\hat{\alpha} \asymp \Big(\frac{\log \frac{d}{2s}}{\log d}\Big)^{\frac{1}{6}}\sqrt{\frac{s}{K_u}}\Big(\frac{\log \frac{d}{2s}}{n}\Big)^{1/3}.$$
     Putting this into (\ref{fano2}) completes the proof. \hfill $\square$

\section{Proofs: Low-rank Matrix Completion}\label{appendixC}
\subsection{Sub-Gaussian Data}
\noindent{\textbf{Proof of Lemma \ref{lemma3}}.} \textbf{I.} We first prove several facts that would be frequently used later.

\noindent{\underline{\it Fact 1:}} $\mathbbm{E}\bm{X_k^TX_k} = \mathbbm{E}\bm{X_kX_k^T} = \bm{I_d}/d$.

\noindent
Since $X_k$ and $X_k^T$ follow the same distribution, we only calculate $\mathbbm{E}\bm{X_k^TX_k}$. Equivalent to (\ref{4.2}) we can assume $\bm{X_k}= e_{k(i)}e_{k(j)}^T$ where $(k(i),k(j)) \sim \mathrm{uni}([d]\times [d])$. Then we calculate that
\begin{equation}
    \begin{aligned}
    \mathbbm{E}\bm{X_k^TX_k} &= \mathbbm{E}_{k(i),k(j)} e_{k(j)}e_{k(i)}^Te_{k(i)}e_{k(j)}^T 
    =  \mathbbm{E}_{k(j)}e_{k(j)}e_{k(j)}^T \\&=  \sum_{k(j)=1}^d d^{-1} e_{k(j)}e_{k(j)}^T  =  \bm{I_d}/d.
      \nonumber
    \end{aligned}
\end{equation}
\noindent{\underline{\it Fact 2:}} Given random matrix $\bm{A} \in \mathbb{R}^{d\times d}$, then $\|\mathbbm{E}\bm{A}\|_{\mathrm{op}}\leq \mathbbm{E}\|\bm{A}\|_{\mathrm{op}}$. 
\noindent
Let $\mathcal{S}=\{x\in\mathbb{R}^d:\|x\|_2=1\}$, by using $\|\bm{B}\|_{\mathrm{op}}=\sup_{U,V\in\mathcal{S}}U^T\bm{B}V$, we have 
\begin{equation}
    \begin{aligned}
    \|\mathbbm{E}\bm{A}\|_{\mathrm{op}} = \sup_{U,V\in\mathcal{S}}&\mathbbm{E}[U^T\bm{A}V] \leq \mathbbm{E}[ \sup_{U,V\in\mathcal{S}}U^T\bm{A}V] = \mathbbm{E}\|\bm{A}\|_{\mathrm{op}}
    \nonumber
    \end{aligned}
\end{equation}
\noindent{\underline{\it Fact 3:}} Given random matrix $\bm{A} \in \mathbb{R}^{d\times d}$, then $\|\mathbbm{E}(\bm{A}-\mathbbm{E}\bm{A})^T(\bm{A}-\mathbbm{E}\bm{A})\|_{\mathrm{op}}\leq \|\mathbbm{E}\bm{A^TA}\|_{\mathrm{op}}$, $\|\mathbbm{E}(\bm{A}-\mathbbm{E}\bm{A})(\bm{A}-\mathbbm{E}\bm{A})^T\|_{\mathrm{op}}\leq \|\mathbbm{E}\bm{AA^T}\|_{\mathrm{op}}$.

\noindent
We only show the first inequality, the second follows likewise. By calculation we have
$$\|\mathbbm{E}(\bm{A}-\mathbbm{E}\bm{A})^T(\bm{A}-\mathbbm{E}\bm{A})\|_{\mathrm{op}} =\|\mathbbm{E} \bm{A^TA}- \mathbbm{E}\bm{A^T}\mathbbm{E}\bm{A} \|_{\mathrm{op}}\leq \|\mathbbm{E} \bm{A^TA}\|_{\mathrm{op}},$$
where we use the positive semi-definiteness of $\mathbbm{E}\bm{A^T}\mathbbm{E}\bm{A}$ and $\mathbbm{E} \bm{A^TA}- \mathbbm{E}\bm{A^T}\mathbbm{E}\bm{A}$.

\noindent{\textbf{II.}} We now start the proof. We first note that $$\bm{\Sigma}_{Y\bm{X}}= \mathbbm{E}(Y_k\bm{X_k})= \mathbbm{E}(\left<\bm{X_k,\Theta^*}\right>\bm{X_k} + \epsilon_k\bm{X_k})= \mathbbm{E}(\left<\bm{X_k,\Theta^*}\right>\bm{X_k}),$$
so by using triangle inequality we obtain 
\begin{equation}
\begin{aligned}
&\Big\|\frac{1}{n}\sum_{k=1}^n\big[\left<\bm{X_k,\Theta^*}\right>-\gamma\cdot \dot{Y}_k\big]\bm{X_k}\Big\|_{\mathrm{op}}  \leq   {\Big\|{\frac{1}{n}\sum_{k=1}^n\big[\gamma \cdot\dot{Y}_k \bm{X_k}-\mathbbm{E}(\gamma\cdot \dot{Y}_k \bm{X_k})\big]}\Big\|_{\mathrm{op}}} +\\&{\Big\|\mathbbm{E}\big[(\gamma\cdot \dot{Y}_k-Y_k) \bm{X_k}\big]\Big\|_{\mathrm{op}}}  + {\Big\|{\frac{1}{n}\sum_{k=1}^n\big[\left<\bm{X_k,\Theta^*}\right>\bm{X_k}-\mathbbm{E}(\left<\bm{X_k,\Theta^*}\right>\bm{X_k})\big]}\Big\|_{\mathrm{op}}}:=R_1+R_2+R_3 .
\end{aligned}
\label{713_4}
\end{equation}

\vspace{1mm}

\noindent{\underline{\it Bound of $R_1$.}} We intend to use matrix Bernstein inequality (See Theorem 6.1.1 in \cite{tropp2015introduction}) to bound $R_1$. Consider a finite seqnence of independent, zero-mean random matrices $\big\{\bm{S_k}:=\gamma \cdot \dot{Y}_k \bm{X_k}-\mathbbm{E}(\gamma\cdot \dot{Y}_k \bm{X_k}):k\in [n]\big\},$
and by Fact 2 we have $$\|\bm{S_k}\|_{\mathrm{op}}\leq \|\gamma \cdot\dot{Y}_k \bm{X_k}\|_{\mathrm{op}} + \|\mathbbm{E}(\gamma \cdot\dot{Y}_k \bm{X_k})\|_{\mathrm{op}}\leq \gamma + \mathbbm{E}\|\gamma\cdot \dot{Y}_k \bm{X_k}\|_{\mathrm{op}} \leq  2\gamma.$$  Then we bound $\max\{\|n\cdot\mathbbm{E}\bm{S_kS_k^T}\|_{\mathrm{op}},\|n\cdot\mathbbm{E}\bm{S_k^TS_k}\|_{\mathrm{op}}\}$. By using Fact 1 and Fact 3. we have 
$
\|\mathbbm{E}\bm{S_kS_k^T}\|_{\mathrm{op}} \leq \|\mathbbm{E}\gamma^2\cdot\bm{X_kX_k^T}\|_{\mathrm{op}} \leq \gamma^2/d,
$
and similarly it holds that $\|\mathbbm{E}\bm{S_k^TS_k}\|_{\mathrm{op}} \leq \gamma^2/d$. Thus, we have $$ \nu\big(\sum_{k=1}^n\bm{S_k}\big) := \max\{\|n\cdot\mathbbm{E}\bm{S_kS_k^T}\|_{\mathrm{op}},\|n\cdot\mathbbm{E}\bm{S_k^TS_k}\|_{\mathrm{op}}\}\leq \frac{n\gamma^2}{d}.$$
By using matrix Bernstein inequality, for any $t>0$ we have 
\begin{equation}
\mathbbm{P}(R_1\geq t)\leq 2d\exp\Big(-\frac{nt^2}{2\gamma[\gamma/d + {2t}/{3}]}\Big).
    \label{713_5}
\end{equation}
We let $t= 2\gamma\sqrt{\frac{\delta \log (2d)}{nd}}$, when $\frac{\delta d \log (2d)}{n}<9/16$ it holds that 
\begin{equation}
    \begin{aligned}
    \mathbbm{P}\Big(R_1 \geq  2\gamma\sqrt{\frac{\delta \log (2d)}{nd}}\Big) \leq  2d\exp\Big(-\frac{ndt^2}{4\gamma^2}\Big)
    \leq (2d)^{1-\delta}.
    \label{C.2}
    \end{aligned}
\end{equation}

\noindent{\underline{\it Bound of $R_2$.}} We first bound the max norm error $\|\mathbbm{E}[(\gamma\cdot\dot{Y}_k-Y_k) \bm{X_k}]\|_{\max}$. Let $X_{k,ij}$ denotes the $(i,j)$-th entry of $\bm{X_k}$. Consider specific $(i,j)\in [d]\times [d]$, then the distribution of $X_{k,ij}$ is given by $\mathbbm{P}(X_{k,ij} =1) = d^{-2}$, otherwise $X_{k,ij} =0$. Also, we let $\Theta^*_{ij}$ be the (i,j)-th entry of $\bm{\Theta^*}$. When $|Y_k |< \gamma$ by Lemma \ref{lemma1} we have $\mathbbm{E}_{\Lambda_k}\big(\gamma\cdot\dot{Y}_k\big)= Y_k.$ Furthermore, it holds that
\begin{equation}
\begin{aligned}
&|\mathbbm{E} (\gamma\cdot \dot{Y}_k - Y_k)X_{k,ij} |\leq  \mathbbm{E}|Y_k|X_{k,ij}\mathbbm{1} (|Y_k|\geq \gamma)
=  \mathbbm{E}\big(\mathbbm{E}\big[|Y_k|X_{k,ij}\mathbbm{1} (|Y_k|\geq \gamma) \big| X_{k,ij}\big]\big) \\
&=  d^{-2} \mathbbm{E}\big[|Y_k|X_{k,ij}\mathbbm{1} (|Y_k|\geq \gamma) \big| X_{k,ij} = 1\big]= d^{-2}\mathbbm{E}\big[|\Theta^*_{ij}+ \epsilon_k|\mathbbm{1}(|\Theta^*_{ij}+ \epsilon_k|\geq \gamma)\big].
\end{aligned}
    \nonumber
\end{equation}
By (\ref{4.5}) we have $|\Theta^*_{ij}|\leq \alpha^*$, recall that $\gamma \geq 2\alpha^*$, so   $|\Theta^*_{ij}+ \epsilon_k|\geq \gamma$ implies   $$|\epsilon_k|\geq |\Theta^*_{ij}+ \epsilon_k| - |\Theta^*_{ij}|\geq \gamma - \alpha^* \geq \frac{\gamma}{2},$$
which implies $|\epsilon_k|\geq \alpha^*$. Moreover, we obtain 
$$|\Theta^*_{ij}+ \epsilon_k|\leq |\Theta^*_{ij}|+ |\epsilon_k|\leq \alpha^* +  |\epsilon_k|\leq 2|\epsilon_k|.$$
Thus, we apply Cauchy-Schwarz inequality, (\ref{4.11}), Proposition \ref{pro1}, it gives
\begin{equation}
\begin{aligned}
&|\mathbbm{E} (\gamma\cdot \dot{Y}_k - Y_k)X_{k,ij} |\leq   d^{-2}\mathbbm{E}\big[|\Theta^*_{ij}+ \epsilon_k|\mathbbm{1}(|\Theta^*_{ij}+ \epsilon_k|\geq \gamma)\big]  \leq  2d^{-2}\mathbbm{E}\big[|\epsilon_k|\mathbbm{1}(|\epsilon_k|\geq \frac{\gamma}{2})\big] \\
&\leq 2d^{-2}\sqrt{\mathbbm{E}\epsilon_k^2}\sqrt{\mathbbm{P}\big(|\epsilon_k|\geq \frac{\gamma}{2}\big)}\lesssim   d^{-2}\sigma \exp\Big(-\frac{D_1\gamma^2}{\sigma^2}\Big) \lesssim  d^{-1}\sigma \sqrt{\frac{\delta  \log (2d)}{nd}},
\end{aligned}
    \nonumber
\end{equation}
where the last "$\lesssim$" follows from $\gamma$ given in (\ref{4.11}) with sufficiently large $C_{13}$. Since the estimation holds for any $(i,j)$, we have $\|\mathbbm{E}[(\gamma \dot{Y}_k-Y_k) \bm{X_k}]\|_{\max} \leq  d^{-1}\sigma \sqrt{\frac{\delta  \log (2d)}{nd}}$. By the relation between $\|.\|_{\mathrm{op}}$ and $\|.\|_{\max}$, it further gives
\begin{equation}
\begin{aligned}
& R_2  \leq d\cdot\|\mathbbm{E}[(\gamma\cdot \dot{Y}_k-Y_k) \bm{X_k}]\|_{\max} \lesssim \sigma \sqrt{\frac{\delta  \log (2d)}{nd}}.
\end{aligned}
    \nonumber
\end{equation}
\noindent{\underline{\it Bound of $R_3$.}} Similar to $R_1$ we use matrix Bernstein inequality. We consider the finite independent, zero-mean random matrix sequence
$$\big\{ \bm{W_k}: =\left<\bm{X_k,\Theta^*}\right>\bm{X_k}-\mathbbm{E}(\left<\bm{X_k,\Theta^*}\right>\bm{X_k}):  k \in [n] \big\}.$$
Note that $|\left<\bm{X_k,\Theta^*}\right>|\leq \alpha^*$, so $\|\left<\bm{X_k,\Theta^*}\right>\bm{X_k}\|_{\mathrm{op}}\leq \alpha^*$, by Fact 2 we have 
$$\|\bm{W_k}\|_{\mathrm{op}}\leq \|\left<\bm{X_k,\Theta^*}\right>\bm{X_k}\|_{\mathrm{op}} + \|\mathbbm{E}\left<\bm{X_k,\Theta^*}\right>\bm{X_k}\|_{\mathrm{op}}\leq 2\alpha^*\leq 2 \gamma .$$
By using Fact 1 and Fact 3, we obtain 
$\|\mathbbm{E}\bm{W_kW_k^T}\|_{\mathrm{op}}\leq \|\mathbbm{E}\bm{\left<\bm{X_k,\Theta^*}\right>^2X_kX_k^T}\|_{\mathrm{op}}\leq \frac{(\alpha^*)^2}{d}.$
Likewise we have $\|\mathbbm{E}\bm{W_k^TW_k}\|_{\mathrm{op}}\leq \frac{(\alpha^*)^2}{d}$, so we derive the bound 
$$\nu\big(\sum_{k=1}^n\bm{W_k}\big):=\max\{\|n\cdot\mathbbm{E}\bm{W_kW_k^T}\|_{\mathrm{op}},\|n\cdot\mathbbm{E}\bm{W_k^TW_k}\|_{\mathrm{op}}\}\leq \frac{n (\alpha^*)^2}{d}\leq \frac{n\gamma^2}{d}.$$
Parallel to $R_1$, by using Matrix Bernstein inequality and set $t = 2\gamma\sqrt{\frac{\delta  \log (2d)}{nd}}$,
\begin{equation}
    \begin{aligned}
    \mathbbm{P}\Big(R_3 \geq  2\gamma\sqrt{\frac{\delta \log (2d)}{nd}}\Big) \leq  (2d)^{1-\delta}.
    \label{C.5}
    \end{aligned}
\end{equation}
\noindent
 We combine the obtained upper bounds for $R_1,R_2,R_3$  and draw the conclusion that with probability higher than $1-2d^{1-\delta}$, we have
$$\Big\|\frac{1}{n}\sum_{k=1}^n\big[\left<\bm{X_k,\Theta^*}\right>-\gamma \cdot\dot{Y}_k\big]\bm{X_k}\Big\|_{\mathrm{op}} \lesssim \gamma \sqrt{\frac{\delta \log d}{nd}}.$$
Now we can use $\gamma \leq C_{13} \max\{\alpha^*,\sigma\}\sqrt{\log n }$ to conclude the proof. \hfill $\square$

\vspace{3mm}

\noindent{\textbf{Proof of Lemma \ref{lemma31}}.} \textbf{I.} We first decompose the complementary event of (\ref{4.14}) which can be stated as $\mathscr{B}=\{\exists\ \bm{\Theta_0}\in \mathcal{C}(\psi), \text{s.t. } \mathcal{F}_{\mathscr{X}}(\bm{\Theta_0})\leq \kappa d^{-2}\|\bm{\Theta_0}\|_{\mathrm{F}}^2-T_0\}$. Note that $\mathbbm{E}\mathcal{F}_{\mathscr{X}}(\bm{\Theta}) = d^{-2}\|\bm{\Theta}\|_{\mathrm{F}}^2$, so $\mathscr{B}$ implies the following event
\begin{equation}
\{\exists\ \bm{\Theta_0}\in \mathcal{C}(\psi), \text{s.t. }|\mathcal{F}_{\mathscr{X}}(\bm{\Theta_0})-\mathbbm{E}\mathcal{F}_{\mathscr{X}}(\bm{\Theta_0})|\geq (1-\kappa)d^{-2}\|\bm{\Theta_0}\|_{\mathrm{F}}^2+T_0\}.
    \label{add1}
\end{equation}
Let $D_0 = (\alpha^*d)^2(\psi\delta\log(2d)/n)^{1/2}$, then by (\ref{4.13}) we have $\|\bm{\Theta_0}\|_{\mathrm{F}}^2\geq D_0$, so by a specific $\beta>1$ (that will be selected later), there exists positive integer $l$ such that $\|\bm{\Theta_0}\|_{\mathrm{F}}^2\in [\beta^{l-1}D_0,\beta^lD_0)$. We further consider $\mathcal{C}(\psi,l) = \mathcal{C}(\psi)\cap\{\bm{\Theta}:\|\bm{\Theta}\|_{\mathrm{F}}^2\in[\beta^{l-1}D_0,\beta^lD_0)\}$, and define a term $$\mathcal{Z}_{\mathscr{X}}(l) = \sup_{\bm{\Theta}\in \mathcal{C}(\psi,l)}|\mathcal{F}_{\mathscr{X}}(\bm{\Theta})-\mathbbm{E}\mathcal{F}_{\mathscr{X}}(\bm{\Theta})|,$$
then we know the event defined in (\ref{add1}) implies the event 
\begin{equation}
\mathscr{B}_{l} = \{\mathcal{Z}_{\mathscr{X}}(l)\geq (1-\kappa)d^{-2}\beta^{l-1}D_0+T_0\}.
    \label{add2}
\end{equation}
By taking the union bound over $l\in\mathbb{N}^{*}$ we obtain $\mathbbm{P}(\mathscr{B})\leq \sum_{l=1}^{\infty}\mathbbm{P}(\mathscr{B}_l).$

\noindent
\textbf{II.} It suffices to bound $\mathbbm{P}(\mathscr{B}_l)$. We first bound the deviation $|\mathcal{Z}_{\mathscr{X}}(l)-\mathbbm{E}\mathcal{Z}_{\mathscr{X}}(l)|$. We consider $\widetilde{\mathscr{X}}=(\widetilde{\bm{X}}_1,\bm{X_2},...,\bm{X_n})$ where only the first component may be different from $\mathscr{X}$
\begin{equation}
    \begin{aligned}
    &\sup_{\mathscr{X},\widetilde{\mathscr{X}}}|\mathcal{Z}_{\mathscr{X}}(l)-\mathcal{Z}_{\widetilde{\mathscr{X}}}(l)|  
    =  \sup_{\mathscr{X},\widetilde{\mathscr{X}}} \Big|\sup_{\bm{\Theta}\in \mathcal{C}(\psi, l)} |\mathcal{F}_{\mathscr{X}}(\bm{\Theta})-\mathbbm{E}\mathcal{F}_{\mathscr{X}}(\bm{\Theta})|-\sup_{\bm{\Theta}\in \mathcal{C}(\psi, l)} |\mathcal{F}_{\widetilde{\mathscr{X}}}(\bm{\Theta})-\mathbbm{E}\mathcal{F}_{\widetilde{\mathscr{X}}}(\bm{\Theta})| \Big| \\
    & \leq  \sup_{\mathscr{X},\widetilde{\mathscr{X}}}\Big|\sup_{\bm{\Theta}\in \mathcal{C}(\psi, l)} |\mathcal{F}_{\mathscr{X}}(\bm{\Theta})-\mathcal{F}_{\widetilde{\mathscr{X}}}(\bm{\Theta}) | \Big| 
    =  \sup_{\bm{X_1},\bm{\widetilde{X}_1}} \sup_{\bm{\Theta}\in \mathcal{C}(\psi, l)} \frac{1}{n}\Big|\big< \bm{X_1},\bm{\Theta}\big>|^2 -|\big< \bm{\widetilde{X}_1},\bm{\Theta}\big>|^2\Big| 
    \leq  \frac{4(\alpha^*)^2}{n}.
    \nonumber
    \end{aligned}
\end{equation}
Note that n components of $\mathscr{X}$ are symmetrical, by bounded different inequality (e.g., Corollary 2.21, \cite{wainwright2019high}), for any $t>0$ we have
\begin{equation}
\mathbbm{P}\Big(\mathcal{Z}_{\mathscr{X}}(l)-\mathbbm{E}\mathcal{Z}_{\mathscr{X}}(l)\geq t\Big)\leq \exp\Big(-\frac{nt^2}{8(\alpha^*)^4}\Big). 
    \label{add3}
\end{equation}
It remains to bound $\mathbbm{E}\mathcal{Z}_{\mathscr{X}}(l)$. Let $\mathscr{E}=(\varepsilon_1,...,\varepsilon_n)$ be i.i.d. Rademacher random variables satisfying $\mathbbm{P}(\varepsilon_k=1)=\mathbbm{P}(\varepsilon_k=-1)=1/2$, then by symmetrization of expectations (e.g., Theorem 16.1, \cite{van2016estimation}), Talagrand's inequality (e.g., Theorem 16.2, \cite{van2016estimation}), the second constraint in (\ref{4.13}), it yields that
\begin{equation}
    \begin{aligned}
    &\mathbbm{E}\mathcal{Z}_{\mathscr{X}}(l) = \mathbbm{E}\sup_{\bm{\Theta}\in \mathcal{C}(\psi,l)}|\mathcal{F}_{\mathscr{X}}(\bm{\Theta})-\mathbbm{E}\mathcal{F}_{\mathscr{X}}(\bm{\Theta})| 
    = \mathbbm{E}\sup_{\bm{\Theta}\in \mathcal{C}(\psi,l)} \Big|\frac{1}{n}\sum_{k=1}^n\big\{\big<\bm{X_k},\bm{\Theta}\big>^2-\mathbbm{E}\big<\bm{X_k},\bm{\Theta}\big>^2\big\}\Big| \\
    &\leq 2\mathbbm{E}_{\mathscr{X}}\mathbbm{E}_{\mathscr{E}} \sup_{\bm{\Theta}\in \mathcal{C}(\psi,l)}\Big|\frac{1}{n}\sum_{k=1}^n \varepsilon_k \big<\bm{X_k},\bm{\Theta}\big>^2\Big| \leq 16\alpha^* \mathbbm{E} \sup_{\bm{\Theta}\in \mathcal{C}(\psi,l)} \Big|\big<\frac{1}{n}\sum_{k=1}^n\varepsilon_k\bm{X_k},\bm{\Theta}\big>\Big| \\
    &\leq 16\alpha^* \mathbbm{E}\Big\|\frac{1}{n}\sum_{k=1}^n\varepsilon_k\bm{X_k}\Big\|_{\mathrm{op}}\sup_{\bm{\Theta}\in \mathcal{C}(\psi,l)}\|\bm{\Theta}\|_{\mathrm{nu}}\leq 160\alpha^*r^{\frac{1}{2-q}}\{\beta^lD_0\}^{\frac{1-q}{2-q}}\mathbbm{E}\Big\|\frac{1}{n}\sum_{k=1}^n\varepsilon_k\bm{X_k}\Big\|_{\mathrm{op}}.
    \label{add4}
    \end{aligned}
\end{equation}
Assume $d\log(2d)/n<1/16$, by matrix bernstein inequality (Theorem 6.1.1, \cite{tropp2015introduction}) it holds that $\mathbbm{E}\Big\|\frac{1}{n}\sum_{k=1}^n\varepsilon_k\bm{X_k}\Big\|_{\mathrm{op}}\leq \frac{3}{2}\sqrt{\frac{\log(2d)}{nd}}$. We then plug it in (\ref{add4}), some algebra yields 
\begin{equation}
\mathbbm{E}\mathcal{Z}_{\mathscr{X}}(l)\leq \big\{(2-q)T_0\big\}^{\frac{1}{2-q}}\big\{d^{-2}\beta^lD_0\big\}^{\frac{1-q}{2-q}}\leq \frac{1-q}{2-q}\frac{\beta^lD_0}{d^2}+T_0
    \label{add5}
\end{equation}
By combining with (\ref{add2}), (\ref{add3}) and let $\kappa_1 = \frac{1-\kappa}{\beta}-\frac{1-q}{2-q}$ (here we assume $\kappa_1\in(0,1)$ since we can choose $\kappa$ sufficiently close to 0, $\beta$ sufficiently close to 1), we have
\begin{equation}
\mathbbm{P}(\mathscr{B}_l)\leq \mathbbm{P}\Big(\mathcal{Z}_{\mathscr{X}}(l)-\mathbbm{E}\mathcal{Z}_{\mathscr{X}}(l)\geq \kappa_1 \frac{\beta^lD_0}{d^2}\Big)\leq \exp\Big(-\frac{n\kappa_1^2\beta^{2l}D_0^2}{8(\alpha^*d)^4}\Big).
    \label{add6}
\end{equation}
We further plug in $D_0$ and use $\beta^{2l}\geq 2l\log \beta$, it yields that
$$\mathbbm{P}(\mathscr{B})\leq \sum_{l=1}^\infty\mathbbm{P}(\mathscr{B}_l)\leq \sum_{l=1}^\infty \big[(2d)^{-\frac{\psi\delta\kappa_1^2\log \beta}{4}}\big]^l \leq d^{-\delta}, $$
the last inequality holds since we can let $\psi$ be large such that $\psi \geq 4(\kappa_1^2 \log \beta)^{-1}$. \hfill $\square$

\vspace{3mm}

\noindent{\textbf{Proof of Theorem \ref{theorem9}}.} 
\noindent
\textbf{I.} By Lemma \ref{lemma3} we can choose sufficiently large $C_{14}$ in (\ref{4.16}) to ensure (\ref{4.7}) holds with probability higher than $1-2d^{1-\delta}$, then (\ref{4.8}) holds with high probability. From Lemma \ref{lemma31} we can further rule out probability $d^{-\delta}$ to ensure (\ref{4.14}) holds.

By (\ref{4.5}) and (\ref{4.6}) we have $\|\bm{\widehat{\Delta}}\|_{\max} \leq \| \bm{\widehat{\Theta}}\|_{\max}+\|\bm{\Theta^*}\|_{\max}\leq 2\alpha^*$. Thus, the estimation error  satisfies the first constraint of $\mathcal{C}(\psi)$. Since (\ref{4.8}) displays the second constraint in $\mathcal{C}(\psi)$, whether  $\bm{\widehat{\Delta}}\in\mathcal{C}(\psi)\in\mathcal{C}(\psi) $ holds only depends on the third constraint, and let us discuss as follows:

\noindent{\underline{\it Case 1}.} $\bm{\widehat{\Delta}}\notin\mathcal{C}(\psi)$. Note that it can only violate the third constraint of $\mathcal{C}(\psi)$, so we know that $\|\bm{\widehat{\Delta}}\|_{\mathrm{F}}^2 \leq (\alpha^*d)^2\sqrt{\frac{\psi\delta\log(2d)}{n}}$. Under the assumption $r\gtrsim d^q$, $n\lesssim d^2\log(2d)$, it holds that
\begin{equation}
\frac{\|\bm{\widehat{\Delta}}\|^2_{\mathrm{F}}}{d^2}\lesssim (\alpha^*)^2\sqrt{\frac{\delta \log(2d)}{n}} \lesssim (\alpha^*)^2\frac{\delta d\log d}{n}\lesssim rd^{-q}\Big((\alpha^*)^2\frac{\delta d\log d}{n}\Big)^{1-q/2}.
    \label{add7}
\end{equation}

\noindent{\underline{\it Case 2}.} 
$\bm{\widehat{\Delta}}\in\mathcal{C}(\psi)$. By (\ref{4.14}) we know $\mathcal{F}_{\mathscr{X}}(\bm{\widehat{\Delta}})\geq \kappa d^{-2}\|\bm{\widehat{\Delta}}\|_{\mathrm{F}}^2-T_0$. If $T_0 \geq \frac{1}{2}\kappa d^{-2}\|\bm{\widehat{\Delta}}\|_{\mathrm{F}}^2$, then we plug in $T_0$ and obtain 
\begin{equation}
\frac{\|\bm{\widehat{\Delta}}\|_{\mathrm{F}}^2}{d^2} \lesssim rd^{-q}\Big((\alpha^*)^2\frac{\delta d\log d}{n}\Big)^{1-q/2}.
    \label{add8}
\end{equation}
If $T_0\leq \frac{1}{2}\kappa d^{-2}\|\bm{\widehat{\Delta}}\|_{\mathrm{F}}^2$, then we have $\mathcal{F}_{\mathscr{X}}(\bm{\widehat{\Delta}})\geq \frac{1}{2}\kappa d^{-2}\|\bm{\widehat{\Delta}}\|_{\mathrm{F}}^2$. Note that this displays the RSC in (\ref{4.9}), so we now use Corollary \ref{corollary3} and obtain 
\begin{equation}
\frac{\|\bm{\widehat{\Delta}}\|_{\mathrm{F}}^2}{d^2}\lesssim rd^{-q}\Big(\max\{(\alpha^*)^2,\sigma^2\}\log d \log n \frac{\delta d }{n}\Big)^{1-q/2}. 
    \label{add9}
\end{equation}
Now we can see that in all cases considered above, the bound of $\|\bm{\widehat{\Delta}}\|_{\mathrm{F}}^2/d^2$ in (\ref{5.17}) holds. Then a direct application of (\ref{4.8}) delivers the bound of $\|\bm{\widehat{\Delta}} \|_{\mathrm{nu}}/d$ in (\ref{5.17}). To conclude, (\ref{5.17}) holds with probability higher than $1-3d^{1-\delta}$. \hfill $\square$

\subsection{Heavy-tailed Data}
\noindent{\textbf{Proof of Lemma \ref{lemma4}}.} From (\ref{4.1}) we have $\mathbbm{E}(Y_k\bm{X_k}) = \mathbbm{E}(\left<\bm{X_k,\Theta^*}\right>\bm{X_k})$, and since $\eta < \gamma$ by Lemma \ref{lemma1} we know $\mathbbm{E}_{\Lambda_k}\big(\gamma \cdot\dot{Y}_k \big)= \widetilde{Y}_k$, hence we have
\begin{equation}
\begin{aligned}
&\Big\|\frac{1}{n}\sum_{k=1}^n\Big[\left<\bm{X_k,\Theta^*}\right>-\gamma\cdot \dot{Y}_k\Big]\bm{X_k}\Big\|_{\mathrm{op}}  \leq  {\Big\|\frac{1}{n}{\sum_{k=1}^n\big[\gamma\cdot \dot{Y}_k \bm{X_k}-\mathbbm{E}(\gamma\cdot \dot{Y}_k \bm{X_k})\big]}\Big\|_{\mathrm{op}}} +\\&{\Big\|\mathbbm{E}\big[(\widetilde{Y}_k-Y_k) \bm{X_k}\big]\Big\|_{\mathrm{op}}}  + {\Big\|\frac{1}{n}{\sum_{k=1}^n\big[\left<\bm{X_k,\Theta^*}\right>\bm{X_k}-\mathbbm{E}(\left<\bm{X_k,\Theta^*}\right>\bm{X_k})\big]}\Big\|_{\mathrm{op}}}:=R_1+R_2+R_3.
\end{aligned}
\nonumber
\end{equation}

\vspace{2mm}

\noindent
\underline{\it Bound of $R_1,R_3$.} We use matrix Bernstein inequality (Theorem 6.1.1, \cite{tropp2015introduction}), and the arguments are exactly the same as the corresponding parts in the proof of Lemma \ref{lemma3}. As a result, one can still invoke Matrix Bernstein to show (\ref{C.2}) and (\ref{C.5}), but only with different value of $\gamma$. To obtain the explicit form of the bounds, we further plug in  $\gamma$ in (\ref{5.19}), with probability higher than $1-2d^{1-\delta}$ it gives 
\begin{equation}
  \max\{R_1,R_3\}\lesssim  \max\{\alpha^*,\sqrt{M}\}\left(\frac{\delta  \log d}{nd^3}\right)^{{1}/{4}} .
    \nonumber
\end{equation}

\vspace{1mm}

\noindent
\underline{\it Bound of $R_2$.} Let $X_{k,ij}$ be the $(i,j)$-th entry of $\bm{X_k} $, where $(i,j)\in [d]\times [d]$ is fixed, we first bound the element-wise error $|\mathbbm{E}(\widetilde{Y}_k - Y_k)X_{k,ij} |$. Recall the definition of truncation, Lemma \ref{lemma1} gives 
$$|\mathbbm{E}\big[(\widetilde{Y}_k - Y_k)X_{k,ij}\big] |= |\mathbbm{E}\big[(\widetilde{Y}_k - Y_k)X_{k,ij}\mathbbm{1}(|Y_k|> \eta)\big]|\leq \mathbbm{E} |Y_k|X_{k,ij}\mathbbm{1}(|Y_k|> \eta).$$
Note that $X_{k,ij}$ can only be $1$ or $0$, and $\mathbbm{P}(X_{k,ij} = 1) = d^{-2}$. Let $\Theta^*_{ij}$ be the $(i,j)$-th entry of $\bm{\Theta^*}$, we further compute it via law of total expectation, then use Cauchy-Schwarz inequality and Marcov's inequality, finally plug in $\eta$ finally. These steps deliver
\begin{equation}
\begin{aligned}
&\mathbbm{E} |Y_k|X_{k,ij}\mathbbm{1}(|Y_k|> \eta)=  \mathbbm{E}\Big(\mathbbm{E}\Big[|Y_k|X_{k,ij}\mathbbm{1}(|Y_k|> \eta) \Big| X_{k,ij}\Big]\Big) \\
&=  d^{-2} \mathbbm{E} \Big[|Y_k|X_{k,ij}\mathbbm{1}(|Y_k|> \eta) \Big| X_{k,ij}=1\Big]
=  d^{-2}\mathbbm{E} \Big[|\Theta^*_{ij}+ \epsilon_k|\mathbbm{1}(|\Theta^*_{ij}+ \epsilon_k| \geq \eta)\Big]\\
& \leq  d^{-2}\sqrt{\mathbbm{E}|\Theta^*_{ij}+ \epsilon_k|^2\mathbbm{P}(|\Theta^*_{ij}+ \epsilon_k|\geq \eta)} \leq d^{-2}\eta^{-1}\mathbbm{E}|\Theta_{ij}^*+\epsilon_k|^2 \\
&\leq 2d^{-2}\eta^{-1}[(\Theta^*_{ij})^2+\mathbbm{E}\epsilon_k^2] \leq 4(dC_{15})^{-1}\max\{\alpha^*,\sqrt{M}\}\Big(\frac{\delta  \log d}{nd^3}\Big)^{1/4}
 \end{aligned}
    \nonumber
\end{equation}
Since the above analysis works for all $(i,j) \in [d]\times [d]$, this is also an upper bound for the max norm, which delivers a bound for operator norm 
\begin{equation}
R_2  \leq d \Big\|\mathbbm{E}\big[(\widetilde{Y}_k-Y_k) \bm{X_k}\big]\Big\|_{\max}\lesssim \max\{\alpha^*,\sqrt{M}\}\Big(\frac{\delta  \log d}{nd^3}\Big)^{{1}/{4}}.
    \label{C.11}
\end{equation}
The result follows from the upper bounds for $R_1,R_2,R_3$. \hfill $\square$

\vspace{3mm}

\noindent{\textbf{Proof of Theorem \ref{theorem10}}.} By Lemma \ref{lemma4} we can choose sufficiently large $C_{17}$ in (\ref{5.21}) to ensure (\ref{4.7}) holds with probability higher than $1-2d^{1-\delta}$, then it further implies (\ref{4.8}), meaning that $\bm{\widehat{\Delta}}$ satisfies the second constraint of $\mathcal{C}(\psi)$. From Lemma \ref{lemma31} we can further rule out probability $d^{-\delta}$ so that (\ref{4.14}) holds. Evidently we have $\|\bm{\widehat{\Delta}}\|_{\max}\leq 2\alpha^*$. Thus, with probability higher than $1-3d^{1-\delta}$, $\bm{\widehat{\Delta}}$ satisfies the first two constraints of $\mathcal{C}(\psi)$, and (\ref{4.14}) holds. Based on these  conditions we further discuss as follows:

\noindent{\underline{\it Case 1.}} $\bm{\widehat{\Delta}}\notin \mathcal{C}(\psi)$, then by exactly the same analysis in proof of Theorem \ref{theorem9}, we obtain (\ref{add7}).

\noindent{\underline{\it Case 2.}} $\bm{\widehat{\Delta}}\in \mathcal{C}(\psi)$, then we have $\mathcal{F}_{\mathscr{X}}(\bm{\widehat{\Delta}}) \geq \kappa d^{-2}\|\bm{\widehat{\Delta}}\|_{\mathrm{F}}^2 - T_0$. If $T_0 \geq \frac{1}{2}\kappa d^{-2}\|\bm{\widehat{\Delta}}\|_{\mathrm{F}}^2$, then (\ref{add8}) holds. Otherwise, we have the restricted strong convexity $\mathcal{F}_{\mathscr{X}}(\bm{\widehat{\Delta}})\geq \frac{1}{2}\kappa d^{-2}\|\bm{\widehat{\Delta}}\|^2_{\mathrm{F}}$. We then apply Corollary \ref{corollary3} and plug in $\lambda$, it holds that 
\begin{equation}
    \|\bm{\widehat{\Delta}}\|_{\mathrm{F}}^2/d^2 \lesssim rd^{-q}\big(\max\{(\alpha^*)^2,M\} \sqrt{\frac{\delta d \log d}{n}}\big)^{1-{q}/{2}}.
    \label{C.18}
\end{equation}
It is not hard to see that the right hand side of (\ref{C.18}) dominates the bound in (\ref{add7}) and (\ref{add8}), so the bound for $\|\bm{\widehat{\Delta}}\|_{\mathrm{F}}^2/d^2$ in (\ref{5.22}) holds. The bound for $\|\bm{\widehat{\Delta}}\|_{\mathrm{nu}}/d $ follows from a direct application of (\ref{4.8}). \hfill $\square$

\vspace{2mm}

\noindent{\textbf{Proof of Theorem \ref{lower1}}.} We simply write $\mathscr{K}(r,\alpha^*)$ as $\mathscr{K}$. Recall our choice of dithering scale $$\gamma \asymp (\alpha^*+\sqrt{M})\Big(\frac{n}{d\log d}\Big)^{1/4}.$$
We   define $\hat{\alpha} = \min\big\{\frac{1}{4}\alpha^*, D_1\gamma\sqrt{\frac{rd}{n}}\big\}$, then by invoking \cite[Lemma A.3]{davenport20141} (set $\gamma$ therein to 1), we can find $\mathscr{K}_1= \{\bm{\Theta}^{(1)},\bm{\Theta}^{(2)},...,\bm{\Theta}^{(N)}\}\subset \mathscr{K}$ such that: 
\begin{itemize}
    \item $N\geq \exp\big(\frac{rd}{16}\big)$;
    \item  $\mathscr{K}_1\subset \{-\hat{\alpha},\hat{\alpha}\}^{d\times d}$, i.e., entries of $\bm{\Theta}\in \mathscr{K}_1$ are either $\hat{\alpha}$ or $-\hat{\alpha}$; 

    \item for different $\bm{\Theta}^{(i)},\bm{\Theta}^{(j)}\in \mathscr{K}_1$, it holds that $\|\bm{\Theta}^{(i)}-\bm{\Theta}^{(j)}\|_F^2 >\frac{1}{2}(\hat{\alpha}d)^2$.
\end{itemize}
It is sufficient to deal with the noiseless case $\epsilon_k=0$, i.e., when the underlying matrix is $\bm{\Theta}$, we let $\mathsf{T}_\eta(a)=\sign(a)\min\{|a|,\eta\}$ be the truncation operator and have the observations 
$$\mathscr{P}(\bm{\Theta})=\{\dot{Y}_k = \sign\big(\mathsf{T}_\eta(\big<\bm{X}_k,\bm{\Theta}\big>)+\Lambda_k\big),k=1,2,...,n\}.$$
By reduction to hypothesis testing and Fano's inequality \cite[Section 4]{rigollet2015high}, we have  
\begin{equation}
    \label{fano1}
    \inf_{\bm{\widehat{\Theta}}}\sup_{\bm{\Theta}}~\mathbbm{P}_{\bm{\Theta}}\big(\|\bm{\widehat{\Theta}}-\bm{\Theta}\|_F^2>\frac{1}{8}(\hat{\alpha}d)^2\big)\geq 1-\frac{\frac{1}{N^2}\sum_{i,j=1}^{N}\mathsf{KL}(\mathscr{P}(\bm{\Theta^{(i)}}),\mathscr{P}(\bm{\Theta}^{(j)}))+\log 2}{\log N}.
\end{equation}
We now estimate $\mathsf{KL}(\mathscr{P}(\bm{\Theta^{(i)}}),\mathscr{P}(\bm{\Theta}^{(j)}))$. Let $\bm{\Theta}^{(i)}$ be the underlying parameter, because $\big<\bm{X}_k,\bm{\Theta}^{(i)}\big> = \pm \hat{\alpha}$, and also $\eta >\hat{\alpha}$ holds trivially, so $\dot{Y}_k= \sign\big(\big<\bm{X}_k,\bm{\Theta}^{(i)}\big>+\Lambda_k\big)$ follows a (symmetrized) Bernoulli distribution with success probability $$\mathbbm{P}\big(\dot{Y}_k=1\big)= \mathbbm{P}\big(\Lambda_k>-\big<\bm{X}_k,\bm{\Theta}^{(i)}\big>\big)= \frac{\gamma \mp \hat{\alpha} }{2\gamma} = \frac{1}{2}\mp\frac{\hat{\alpha}}{2\gamma}.$$
Moreover, because $\mathscr{P}(\bm{\Theta}^{(i)})$ consists of n i.i.d. observations,  
$$\mathsf{KL}(\mathscr{P}(\bm{\Theta^{(i)}}),\mathscr{P}(\bm{\Theta}^{(j)}))\leq n\cdot \mathsf{KL}\Big(\frac{1}{2}+\frac{\hat{\alpha}}{2\gamma},\frac{1}{2}-\frac{\hat{\alpha}}{2\gamma}\Big),$$
where we use $\mathsf{KL}(p,q)$ to denote the KL divergence between Bernoulli distribution with success probability $p$ and $q$. Because $|\frac{\hat{\alpha}}{2\gamma}|\leq \frac{1}{4}\frac{\alpha^*}{2\gamma}\leq \frac{1}{8}$ (recall that we assume $\gamma>\eta>\alpha^*$), by using \cite[Lemma A.4]{davenport20141} we obtain \begin{equation}
\begin{aligned}
    \mathsf{KL}(\mathscr{P}(\bm{\Theta^{(i)}}),\mathscr{P}(\bm{\Theta}^{(j)}))\leq \frac{n(\hat{\alpha}/\gamma)^2}{(\frac{1}{2}+\frac{\hat{\alpha}}{2\gamma})(\frac{1}{2}-\frac{\hat{\alpha}}{2\gamma})}\leq \frac{16n\hat{\alpha}^2}{3\gamma^2}\leq \frac{16D_1^2rd}{3}<\frac{rd}{150},
\end{aligned}
\end{equation}
note that the last two inequalities holds because $\hat{\alpha}\leq D_1\gamma\sqrt{\frac{rd}{n}}$ and we can select  sufficiently small $D_1$. Now we put this into   (\ref{fano1}). Because $\log N \geq \frac{rd}{16}$, with slightly large $rd$ to overcome $\log 2$, the probability term on the right hand side of (\ref{fano1}) is greater than $\frac{3}{4}$, while the event on the left hand side is just 
$$\|\bm{\widehat{\Theta}}-\bm{\Theta}\|_F^2/d^2>\frac{1}{8}\hat{\alpha}^2=\min\Big\{\frac{(\alpha^*)^2}{128},D_2\big((\alpha^*)^2+M\big)r\sqrt{\frac{d}{n\log d}}\Big\} ,$$
where we plug in the value of $\gamma$. The proof is complete. \hfill $\square$

\section{Comparisons with Related Work}
\label{appendxD}

\subsection{1-bit Compressed Sensing}

In this part we compare our Theorems \ref{1bitcssg}-\ref{1bitcsht} with existing results of 1-bit CS. 

The traditional setting of 1-bit CS, where one aims to recovery a sparse $d$-dimensional signal $\Theta^*$ based on measurement $\dot{Y}_k=\mathrm{sign}(X_k^T\Theta^*)$ with some $X_k$, was first introduced in \cite{boufounos20081} and widely studied in subsequent works (e.g., \cite{jacques2013robust,plan2012robust,plan2013one}). By projection-based method \cite{plan2017high} or K-Lasso \cite{plan2016generalized}, similar results were obtained for a model with more general observation (that involves possibly unknown nonlinearity) and signal structure. Nevertheless, all these results are restricted to Gaussian sensing vectors that can be unrealistic in practice\footnote{More precisely, \cite{plan2016generalized} handles $X_k\sim \mathcal{N}(0,\bm{\Sigma})$ with unknown $\bm{\Sigma}$ while other several papers above assume $X_k\sim\mathcal{N}(0,\bm{I_d})$.}. There does exist one work, \cite{ai2014one}, presents result for $X_k$ with i.i.d. sub-Gaussian entries. However, the result in \cite{ai2014one} is still overly restrictive and impractical, see the discussions in \cite{dirksen2021non}.

To overcome the   restriction of Gaussian sensing vector (and also some other limitations), it was recently realized that introducing dithering noise can help. With dithering noise $\Lambda_k$, the measurement  now becomes $\dot{Y}_k = \mathrm{sign}(X_k^T\Theta^* + \Lambda_k)$. In this setting, specifically, we can recover the signal with norm information \cite{knudson2016one}, achieve exponentially-decaying error rate (This requires adaptive dithering) \cite{baraniuk2017exponential}, and perhaps more prominently, accommodate non-Gaussian $X_k$ \cite{thrampoulidis2020generalized,dirksen2018robust,dirksen2021non}. In what follows, we will focus on comparing Theorems \ref{1bitcssg}-\ref{1bitcsht} with  the most relevant works \cite{thrampoulidis2020generalized,dirksen2021non} that adopt uniform dithering noise. The comparisons will be conducted on exactly sparse $\Theta^*$ since \cite{dirksen2021non,thrampoulidis2020generalized} do not adopt the formulation $\sum_{k=1}^n|\theta_k^*|^q \leq s,~q\in (0,1)$ for approximately sparse $\Theta^*$. For other developments on 1-bit CS (or more generally, quantized compressed sensing), we refer  readers  to the survey papers \cite{boufounos2015quantization,dirksen2019quantized}.

Dirksen and Mendelson   \cite{dirksen2021non} first   essentially extend 1-bit CS  to non-Gaussian $X_k$. Their methodology is   based on random hyperplane tessellation. Specifically, under sub-Gaussian or even heavy-tailed $X_k$\footnote{In \cite{dirksen2021non}, heavy-tailed $X_k$ is assumed to satisfy $\mathbbm{E}\big(|v^TX_k|^2\big) \leq L\big(\mathbbm{E}|v^TX_k|\big)^2 $ for any $v\in \mathbb{R}^d$.} with uniform dithering noise, they show a relatively small number of random hyperplanes (that depends on the complexity of $\Theta^*$) leads to $\rho$-uniform tessellation on the signal set of interest. 
Moreover, they apply the new hyperplane tessellation results to 1-bit CS and propose two reconstruction optimization problems 
\begin{equation}
    \label{79_2}
    \begin{cases}
    \displaystyle  \mathrm{(a):} ~\widehat{\Theta} \in \mathop{\arg\min}\limits_{\Theta \in \mathbb{R}^d} \ \sum_{k=1}^n\mathbbm{1}(\mathrm{sign}(X_k^T\Theta + \Lambda_k)\neq Y_k),~~~\mathrm{s.t.}~~\|\Theta\|_0 \leq s,~\|\Theta\|_2 \leq 1
    \\
     \displaystyle  \mathrm{(b):}~\widehat{\Theta} \in \mathop{\arg\min}\limits_{\Theta \in \mathbb{R}^d} \    \frac{1}{2\lambda} \|\Theta\|_2^2  -\frac{1}{2n}\sum_{k=1}^n Y_kX_k^T\Theta ,~~~\mathrm{s.t.}~~\|\Theta\|_1 \leq s,~\|\Theta\|_2 \leq 1
    \end{cases}.
\end{equation}
Although (a) is shown to possess uniform recovery guarantee with fast rate in both sub-Gaussian and heavy-tailed $X_k$, it is essentially intractable due to the $\ell_0$ constraint and the 0-1 objective function. Also, a secondary drawback is that, the information of $\Lambda_k$ is needed in problem (a), which induces undesired memory or transmission costs. For these reasons, (a) is mainly of theoretical interest. Then, (b) is proposed as a convex relaxation  to remedy the downside of (a). Under sub-Gaussian $X_k$ and $\epsilon_k$, the error rate of (b) for $s$-sparse $\Theta^*$ was shown to be $\tilde{O}\big( \sqrt[\leftroot{-3}\uproot{3}4]{\frac{s}{n}} \big)$  
, and this is inferior to the near optimal rate $\tilde{O}\big(\sqrt{\frac{s}{n}}\big)$ provided by our Theorem \ref{1bitcssg}. Indeed, their error rate for sub-Gaussian data is even worse  than our rate for heavy-tailed data, i.e., $\tilde{O}\big(\sqrt[\leftroot{-3}\uproot{3}3]{\frac{s^2}{n}} \big)$ given in Theorem \ref{1bitcsht}, while the guarantee of (b) under heavy-tailed $X_k$ has not yet been established. On the other hand, the advantages of \cite{dirksen2021non} is that their guarantee is uniform, as an outcome of the hyperplane tessellation method. They also obtain a partial extension to structured random measurement matrix in the companion work \cite{dirksen2018robust}. These two aspects are  left as future research directions of our theories.



A result directly comparable to our Theorem \ref{1bitcssg} is due to Thrampoulidis and Rawat \cite{thrampoulidis2020generalized}. Under almost the same setting 
they assume $\Theta^*\in \mathcal{T}$ and consider a constrained Lasso 
\begin{equation}
    \widehat{\Theta} \in \mathop{\arg\min}\limits_{\Theta \in \mathcal{T}} \ \frac{1}{2n}\sum_{k=1}^n \big(X_k^T\Theta - \gamma\cdot \dot{Y_k}\big)^2.
    \label{79_1}
\end{equation}
This is analogous to our convex programming problem (\ref{add72_5}): up to constant, the objective of (\ref{79_1}) equals the loss function (i.e., the first two terms) in (\ref{add72_5});   the only difference is that, the structure of $\Theta^*$, specifically sparsity, is incorporated into (\ref{79_1}) via the constraint, but appears   in (\ref{add72_5}) as a regularizer. The relevant result is  in  \cite[Theorem IV.1]{thrampoulidis2020generalized}. Interestingly, when specialized to exactly $s$-sparse $\Theta^*$, they choose $\mathcal{T} = \{ \Theta:\|\Theta\|_1 \leq \|\Theta^*\|_1\}$ and show $\ell_2$ norm error rate $  \tilde{O}\big(\sqrt{\frac{s   }{n}}\big)$ that   coincides with Theorem \ref{1bitcssg}. Despite these similarities, our Theorem \ref{1bitcssg} exhibits several obvious improvements. Firstly, we consider pre-quantization noise $\epsilon_k$ while they only study noiseless case. Secondly, we assume zero-mean $X_k$ satisfies $\lambda_{\min}(\bm{\Sigma}_{XX})= \Omega(1)$, but \cite{thrampoulidis2020generalized} requires symmetric $X_k$ to satisfy a nondegeneracy condition formulated as $\inf_{\|v\|_2 = 1}\mathbbm{E}|v^TX_k| = \Omega(1)$, which is more restrictive. Thirdly, their guarantee is valid with probability at least $0.99$, while our probability term $1-O(d^{2-\delta})$ is finer. In addition, we comment that  a  pre-estimation of $\|\Theta^*\|_1$ is needed to specify $\mathcal{T}$ in (\ref{79_1}), while our unconstrained program (\ref{add72_5}) is free of this issue and hence    more practically appealing. On the other hand, their advantage  is    the more general assumption on signal structure (i.e., $\Theta^*\in T$). 

We also compare with two less related works \cite{knudson2016one,baraniuk2017exponential}, where the authors study $\dot{Y}_k = \mathrm{sign}(X_k^T \Theta^* + \Lambda_k)$ with Gaussian dithering noise $\Lambda_k$. Convex programming problems are  proposed in \cite{knudson2016one,baraniuk2017exponential} to recover $s$-sparse $\Theta^*$, but their results are only valid for standard Gaussian $X_k$. Specifically, the theoretical rate in   \cite[Theorem 4]{knudson2016one} reads $\tilde{O}\big( \sqrt[\leftroot{-3}\uproot{3}5]{\frac{s}{n}} \big)$, while \cite[Theorem 2]{baraniuk2017exponential} provides an error rate $\tilde{O}\big(\sqrt[\leftroot{-3}\uproot{3}4]{\frac{s}{n}}\big)$. Note that both are slower than the rates presented in our Theorem \ref{1bitcssg} (sub-Gaussian $X_k$), Theorem \ref{1bitcsht} (heavy-tailed $X_k$).

Therefore, our 1-bit CS results improve on the prior ones in terms of generality of sensing vectors and random noise (that can be sub-Gaussian or heavy-tailed), convergence rate. In particular,   while   \cite[Theorem 1.11]{dirksen2021non}  involves an intractable program and a   parameter hard to control (see $E(T_r)$ therein), our Theorem \ref{1bitcsht} provides the first convex program with rigorous recovery guarantee for 1-bit CS under heavy-tailed sensing vector.


\subsection{1-bit Matrix Completion}

In this part, we compare our Theorems \ref{theorem9}-\ref{theorem10} with existing results on 1-bit matrix completion (1-bit MC), a problem first proposed and studied in \cite{davenport20141,cai2013max}. 
Unlike 1-bit CS where it is still possible to recover the direction ${\Theta^*}/\|\Theta^*\|_2$ from the directly quantized measurement $\dot{Y}_k = \mathrm{sign}(X_k^T\Theta^*)$, due to the nature of the covariate in matrix completion (i.e., $\bm{X_k}= e_{i(k)}e_{j(k)}^T$), 1-bit MC could be extremely ill-posed if we only observe $\dot{Y}_k = \mathrm{sign}\big(\big<\bm{X_k}, \bm{\Theta^*}\big>\big)$. This issue happens even when $\bm{\Theta^*}$ is rank-1, see the discussion in \cite{davenport20141}. Thus, dithering noise (denoted by $\Lambda_k$) is  indispensable for the well-posedness of 1-bit MC, hence the   observation  becomes $$\dot{Y}_k = \mathrm{sign}\big(\big<\bm{X_k},\bm{\Theta}^*\big> + \Lambda_k\big).$$ Existing works consider dithering noise with rather general distribution, but particularly focus on the Logistic model and Probit model\footnote{The Probit model corresponds to Gaussian dithering noise $\Lambda_k \sim \mathcal{N}(0,\sigma^2)$.}.

Let us give a brief review of existing results. In the first study of 1-bit MC \cite{davenport20141}, Davenport et al. proposed to recover $\bm{\Theta^*}$ via negative log-likelihood minimization (put $d_1=d_2 =d$)   \begin{equation}
    \label{710_1}
      \bm{\widehat{\Theta} }\in \mathop{\arg\min}\limits_{\bm{\Theta} \in \mathbb{R}^{d \times d }}\   \mathcal{L}_{\mathrm{NLL}}(\bm{\Theta}), ~~~\mathrm{s.t.}~\|\bm{\Theta}\|_{\max} \leq \alpha^*, ~\|\bm{\Theta}\|_{\mathrm{nu}} \leq \alpha^*d\sqrt{r}~.
\end{equation}
In (\ref{710_1}), the first constraint is commonly used in matrix completion (see the interpretation at the beginning of Section \ref{section4}), the second constraint relaxes $\mathrm{rank}(\bm{\Theta})\leq r$ via the relation $$\|\bm{\Theta}\|_{\mathrm{nu}}\leq \sqrt{\mathrm{rank} ( {\bm{\Theta}})}\|\bm{\Theta}\|_{\mathrm{F}}\leq d\|\bm{\Theta}\|_{\max}\sqrt{\mathrm{rank} ( {\bm{\Theta}}) } ,$$ and the loss function $\mathcal{L}_{\mathrm{NLL}}(\bm{\Theta})$ is  
\begin{equation}
    \label{710_2}
     \begin{aligned}
     \mathcal{L}_{\mathrm{NLL}}(\bm{\Theta})=- \frac{1}{n}\sum_{k=1}^n \Big[\mathbbm{1}(\dot{Y}_k &= 1) \log \mathbbm{P}(\big<\bm{X_k},\bm{\Theta}\big> +\Lambda_k \geq 0  ) \\ &+ \mathbbm{1}(\dot{Y}_k = -1)\log \mathbbm{P}(\big<\bm{X_k},\bm{\Theta}\big>  + \Lambda_k <0 )\Big].
     \end{aligned}
\end{equation}
Some developments can be seen in subsequent works, to name a few, \cite{cai2013max} used another surrogate of matrix rank rather than the nuclear norm\footnote{This rank surrogate is called max-norm but totally different from $\|.\|_{\max}$ in our work. To avoid confusion, we refer readers to   \cite{cai2013max} for the details.}, \cite{klopp2015adaptive,bhaskar2016probabilistic,lafond2014probabilistic} extended 1-bit MC to  finite alphabets, \cite{ni2016optimal,bhaskar2016probabilistic,bhaskar2015probabilistic} imposed (exactly) low-rank constraint without relaxation, \cite{lafond2014probabilistic,klopp2015adaptive} adopted  nuclear norm penalty to avoid the pre-estimation of $\|\bm{\Theta}^*\|_{\mathrm{nu}}$ needed in (\ref{710_1}).

We emphasize that all above works are restricted to a noiseless setting; by saying this, we do not regard $\Lambda_k$ as a detrimental noise since the dithering is indeed beneficial to the recovery.  We are aware of only two recent papers \cite{shen2019robust,gao2018low} that deal with the noisy setting. In \cite{gao2018low}, Gao et al. considered a deterministic sparse pattern $\bm{S^*}$ mixing with the desired  low-rank structure $\bm{\Theta^*}$. Specifically, this more general "low-rank plus sparse" model can be formulated as 
\begin{equation}
    \label{711_1}
    \dot{Y}_k = \mathrm{sign}\big(\big<\bm{X_k},\bm{\Theta^*}+\bm{S^*}\big>+\Lambda_k\big),~~\mathrm{where}~\|\mathrm{vec}(\bm{S^*}) \|_0\leq s.
\end{equation}
In \cite{shen2019robust}, Shen et al. studied 1-bit MC with post-quantization noise in a form of sign flipping, which can be described by
\begin{equation}
    \label{711_2}
    \dot{Y}_k = \delta_{k}\cdot \mathrm{sign}\big(\big<\bm{X_k},\bm{\Theta^*}\big>+\Lambda_k\big),~~\mathrm{where}~\mathbbm{P}(\delta_k = -1) = \tau_0,~\mathbbm{P}(\delta_k = 1) = 1-\tau_0.
\end{equation}
 Evidently, for (\ref{711_1}) or (\ref{711_2}), as done in \cite{shen2019robust,gao2018low}, the recovery can still be based on negative log-likelihood minimization. However, if we consider pre-quantization noise $\epsilon_k$ with unknown distribution (this is a natural and well-studied situation in other statistical estimation problems), i.e., \begin{equation}
    \label{710_3}
    \dot{Y}_k = \mathrm{sign}\big(\big<\bm{X_k},\bm{\Theta^*}\big>+\epsilon_k+\Lambda_k\big),
\end{equation}
recovery based on likelihood no longer works due to lack of knowledge on $\mathcal{L}_{\mathrm{NLL}}(\bm{\Theta})$. Therefore, before our work, it was an open question whether 1-bit MC under unknown pre-quantization random noise is possible.

Our Theorems \ref{theorem9}-\ref{theorem10} provide an affirmative answer to this question. Particularly, under uniformly distributed $\Lambda_k$, sub-Gaussian or even heavy-tailed $\epsilon_k$, we formulate 1-bit CS as a convex programming problem and establish theoretical guarantee. Unlike the likelihood approach, we now use a generalized quadratic loss  
\begin{equation}
    \label{710_4}
    \mathcal{L}(\bm{\Theta}) = \frac{1}{2n}\sum_{k=1}^n \big(\big<\bm{X_k},\bm{\Theta}\big> - \gamma \cdot \dot{Y}_k\big)^2.
\end{equation}
For the core idea behind, while maximum likelihood estimation is a   standard estimation strategy, the inspiration of (\ref{710_4}) is drawn from Lemma \ref{lemma1}, i.e., $\gamma\cdot \dot{Y}_k$ can serve as a surrogate of the full observation $Y_k = \big<\bm{X_k},\bm{\Theta^*}\big>+\epsilon_k$.

At first glance, one may feel that (\ref{710_4}) is a bit coarse compared to negative log-likelihood, but   under uniform dither $\Lambda_k$ and sub-Gaussian $\epsilon_k$ our estimator achieves near minimax rate (Theorem \ref{theorem9}). For comparison, we go back to the noiseless (i.e., $\epsilon_k =0 $ in (\ref{710_3})) and exactly low-rank   (i.e., $q=0$ in (\ref{4.3})) case. In this case, Theorem \ref{theorem9} gives a bound $\tilde{O}\big((\alpha^*)^2\frac{rd }{n}\big)$ for mean squared error. This is faster than $\tilde{O}\big((\alpha^*)^2\sqrt{\frac{rd}{n}}\big)$ obtained in two pioneering works \cite{davenport20141,cai2013max}, and similar to the more recent paper \cite{lafond2014probabilistic}.

To conclude, we present the first result for 1-bit MC with unknown pre-quantization random noise, which can either be sub-Gaussian or heavy-tailed. In addition, by some extra technicalities, we believe our method can be extended to both deterministic sparse corruption in \cite{gao2018low} and sign flipping noise in \cite{shen2019robust}. 

\section{Details and Algorithms in Experiments}
\label{appendixE}
\subsection{Sparse Covariance Matrix Estimation}
\subsubsection{Detailed Simulation} To generate the $d\times d$ underlying covariance matrix $\bm{\Sigma^*}$ that satisfies Assumption \ref{assumption1} with $q = 0$ and sparsity $s$, we first construct  
$$\bm{\Sigma^*_0} = \begin{pmatrix} \bm{\Sigma_1^*} & \bm{0} \\ \bm{0} & \bm{I_{d-3s}} \end{pmatrix},$$
where $\bm{\Sigma^*_1} = \mathrm{diag}(\bm{\Sigma^*_2},\bm{\Sigma^*_2},\bm{\Sigma^*_2}) \in\mathbb{R}^{3s \times 3s}$, and   $\bm{\Sigma^*_2} = [\sigma^*_{2,ij}]\in\mathbb{R}^{s\times s}$ are defined as $\sigma^*_{2,ii} = 1$ for $i\in [s]$, $\sigma^*_{2,12} = \sigma^*_{2,21} = 0.99 - (s-2)\cdot 0.03$, $\sigma^*_{2,ij} = 0.03$ for all other entries. By normalizing the operator norm, we set $$\bm{\Sigma^*} = \frac{\bm{\Sigma_0^*}}{\|\bm{\Sigma_0^*}\|_{\mathrm{op}}}.$$
We i.i.d. draw sub-Gaussian $X_k \sim\mathcal{N}(\bm{0},\bm{\Sigma^*})$, and draw heavy-tailed $X_k$ from Student's t distribution via the Matlab function "mvtrnd($\cdot$)" with $\nu = 6$. Then, we apply the 1-bit quantization scheme with parameters slightly tuned to be well-functioning, to obtain the binary data $\big\{\dot{X}_{kj}:k\in [n],j=1,2\big\}$. Now, we can directly construct   the 1-bit estimator $\bm{\widehat{\Sigma}}$ defined in (\ref{2.4}), (\ref{2.19}), and track the experimental recovery error. In our results, each experiment is obtained as the mean value of  15 independent runs.

\subsection{Sparse Linear Regression}
\subsubsection{Detailed Simulation} We conduct numerical experiments of 1-bit QC-CS (Theorems \ref{theorem7}-\ref{theorem8}) and 1-bit CS (Theorems \ref{1bitcssg}-\ref{1bitcsht}). We consider isotropic covariate (i.e., $\mathbbm{E}X_kX_k^T = \bm{I_d}$), which    admits   Assumption \ref{assumption3} required for 1-bit QC-CS. For the covariate, sub-Gaussian $X_k$ are generated from Gaussian distribution, while entries of heavy-tailed $X_k$ are i.i.d. drawn from $\sqrt{\frac{2}{3}}\cdot t(\nu = 6)$. Here, $ t(\nu = 6)$  represents Student's t distribution with  $6$ degrees of freedom, and $\sqrt{\frac{2}{3}}$ aims to normalize the variance. We set the first $s$ entries of   $\Theta^*$ to be $\frac{1}{\sqrt{s}}$, while other entries are $0$, hence $\Theta^*$ is (exactly) $s$-sparse. {Sub-Gaussian and heavy-tailed noise $\epsilon_k$ are respectively drawn from $\mathcal{N}(0,\sqrt{\frac{3}{5}})$ and $0.3\cdot t(\nu = 6)$.} All these parameters specify the model, so we can generate the full data $\{(X_k,Y_k):k\in [n]\}$ for a specific $(n,d,s)$.

 Then we apply the 1-bit quantization scheme to quantize $\{(X_k,Y_k):k\in [n]\}$ to $\{(\dot{X}_{k1},\dot{X}_{k2},\dot{Y}_{k}):k\in [n]\}$ in 1-bit QC-CS, or $\{(X_k,\dot{Y}_k):k\in [n]\}$ in 1-bit CS. All parameters are properly set according to the Theorems, and we stress that  the truncation and dithering parameters  for $X_k$, $Y_k$ are different. For instance, in sub-Gaussian 1-bit QC-CS we use dithering noise $\Lambda_k \sim \mathrm{uni}\big([-\gamma_Y,\gamma_Y\big)$, $\Gamma_{kj}\sim \mathrm{uni}\big([-\gamma_X,\gamma_X]^d\big)$ with $\gamma_X\neq \gamma_Y$. After the data quantization, we can solve the proposed convex programming problems to obtain the estimator $\widehat{\Theta}$. We track the $\ell_2$ norm error $\|\widehat{\Theta}-\Theta^*\|_2$ and report the mean value of 15 independent runs.

\subsubsection{Algorithm} Note that the convex programming problems (\ref{3.18}), (\ref{add72_7}) and (\ref{add72_10}) share the common formulation of   
\begin{equation}
    \label{711_3}
    \widehat{\Theta} \in \mathop{\arg\min}\limits_{\Theta\in\mathbb{R}^d} \ \frac{1}{2}\Theta^T \bm{\widehat{\Sigma}_1} \Theta - \widehat{\Sigma}_2^T\Theta + \lambda \|\Theta\|_{1},
\end{equation}
where $\bm{\widehat{\Sigma}_1}$ is positive semi-definite, $\widehat{\Sigma}_2\in \mathbb{R}^d$. Here, we use alternating direction method of multipliers (ADMM) to solve (\ref{711_3}), and the convergence of our algorithm is guaranteed since the variable is divided into two blocks \cite{gabay1976dual}. For more details of ADMM, we refer readers to the survey paper \cite{boyd2011distributed}.

We now invoke the framework of ADMM and show the iterative formula.  
Divide $\Theta\in \mathbb{R}^d$ into $M,Z\in\mathbb{R}^d$, (\ref{3.18}) is equivalent to 
$$\mathop{\arg\min}\limits_{M,Z\in\mathbb{R}^d}\ \frac{1}{2}M^T\bm{\widehat{\Sigma}_1}M - \widehat{\Sigma}_2^TM + \lambda\|Z\|_1 ,\ \text{s.t. }M=Z.$$
By introducing the multiplier $\Upsilon\in\mathbb{R}^d$, the   augmented Lagrangian function reads
$$\frac{1}{2}M^T\bm{\widehat{\Sigma}_1}M-\widehat{\Sigma}^T_2M+  \lambda\|Z\|_1 +  \Upsilon^T(M-Z)+\frac{\rho}{2}\| M-Z\|_2^2.$$
Minimizing $(M,Z)$ alternatively and updating $\Upsilon$ via gradient ascent   give the iteration formulas 
\begin{equation}
\label{711_4}
    \begin{cases}
M_{t+1} = (\bm{\widehat{\Sigma}_1}+\rho\cdot \bm{I_d})^{-1}(\widehat{\Sigma}_2+\rho \cdot Z_t -\Upsilon_t) \\
Z_{t+1} =  \mathcal{S}_{\lambda / \rho} (M_{t+1}+\rho^{-1}\cdot\Upsilon_t) \\
\Upsilon_{t+1} = \Upsilon_t + \rho\cdot(M_{t+1}-Z_{t+1})
\end{cases}
\end{equation}
that updates $(M_t,Z_t,\Upsilon_t)$ to $(M_{t+1},Z_{t+1},\Upsilon_{t+1})$. In (\ref{711_4}), we define $\mathcal{S}_{\beta}(x) = \sign(x)\max\{0,|x|- \beta\}$ if $x\in\mathbb{R}$, and then let $\mathcal{S}_{\beta}(\cdot)$ element-wisely operate on vectors. This is known as the soft thresholding operator.

\subsection{Low-rank Matrix Completion}
\subsubsection{Detailed Simulation}  We simulate low-rank matrix completion with exactly low-rank matrix $\bm{\Theta^*}$. The $d\times d$ rank $r$ underlying matrix  $\bm{\Theta^*}$ is generated by the formulation $\bm{\Theta^*} = \frac{\bm{\Theta_l \Theta_r}}{\|\bm{\Theta_l \Theta_r}\|_{\mathrm{F}}}$, where entries of $\bm{\Theta_l} \in\mathbb{R}^{d\times r}$ and $\bm{\Theta_r}\in\mathbb{R}^{r\times d}$ are i.i.d. drawn from $\mathcal{N}(0,1)$. Furthermore, $\bm{\Theta^*}$ with different $(d,r)$ are controlled to possess comparable spikiness $\alpha(\bm{\Theta^*})= \frac{d\| \bm{\Theta^*}\|_{\max}}{\|\bm{\Theta^*}\|_{\mathrm{F}}}$. While   the covariate is specified to be $\bm{X_k} = e_{k(i)}e_{k(j)}^T$ with $(k(i),k(j))\sim \mathrm{uni}([d]\times [d])$ (\ref{4.2}), we test both sub-Gaussian noise and heavy-tailed noise. Specifically, sub-Gaussian or heavy-tailed $\epsilon_k$ are i.i.d. copies of $\mathcal{N}(0,\frac{1}{400})$ or $\frac{1}{250}\cdot \big(\frac{1}{\sqrt{3}}t(\nu = 3)\big)$, respectively. Here, $\frac{1}{\sqrt{3}}t(\nu = 3)$ is the Student's t distribution with $3$ degrees of freedom and variance rescaled to $1$. Following these parameters, the full data $\{(\bm{X_k},Y_k):k\in [n]\}$ are obtained from the model $Y_k = \big<\bm{X_k},\bm{\Theta^*}\big>+\epsilon_k$. The responses are   processed by the 1-bit quantization scheme and quantized to 1-bit $\dot{Y}_k$,  then solving the convex programming problem (\ref{4.6})   gives the estimator $\bm{\widehat{\Theta}}$. In (\ref{4.6}), we set $\alpha^* = \|\bm{\Theta^*}\|_{\max}$ and properly tune $\lambda$ so that it balances the data fidelity and low-rank structure. We track the Frobenius norm error $\|\bm{\widehat{\Theta}}-\bm{\Theta^*}\|_{\mathrm{F}}$ and report the mean value of $15$ independent trials.

\subsubsection{Algorithm}

We similarly apply ADMM to solve (\ref{4.6}), and first  separate variable $\bm{\Theta}$ to be two blocks $\bm{M},\bm{Z}\in\mathbb{R}^{d\times d}$. Define $\mathbbm{1}'(E)$ to be the indicator function widely used in optimization, i.e., $\mathbbm{1}'(E)=0$ if $E$ happens, $\mathbbm{1}'(E) = \infty$ otherwise. Then, we can move the max-norm constraint to objective and obtain the equivalent program
$$\mathop{\arg\min}\limits_{\bm{M},\bm{Z}\in\mathbb{R}^{d\times d}}\ \frac{1}{2n}\sum_{k=1}^n\big(\big<\bm{X_k},\bm{M}\big>-\gamma\cdot\dot{Y}_k\big)^2+\mathbbm{1}'(\| \bm{M}\|_{\max}\leq \alpha^*)+\lambda\|\bm{Z}\|_{\mathrm{nu}},\ \text{s.t. }\bm{M}=\bm{Z}.$$
Let $\bm{\Upsilon}\in\mathbb{R}^{d\times d}$ be the multiplier, we have the augmented Lagrangian function  
$$\frac{1}{2n}\sum_{k=1}^n\big(\big<\bm{X_k},\bm{M}\big>-\gamma\cdot\dot{Y}_k\big)^2+\mathbbm{1}'(\| \bm{M}\|_{\max}\leq \alpha^*)+\lambda\|\bm{Z}\|_{\mathrm{nu}}+\big<\bm{\Upsilon},\bm{M}-\bm{Z}\big>+\frac{\rho}{2}\|\bm{M}-\bm{Z}\|^2_{\mathrm{F}}.$$
Some additional notations are necessary before presenting the algorithms. Let $\mathcal{I}_{ij}=\{k\in[n]:\bm{X_k}=e_ie_j^T\}$, then we define $\bm{J_1} = [\bm{J_1}(i,j)],\bm{J_2}=[\bm{J_2}(i,j)]\in\mathbb{R}^{d\times d}$  as $$\bm{J_1}(i,j)= \sum_{k\in\mathcal{I}_{ij}}\gamma\cdot \dot{Y}_k\ , \   \ \bm{J_2}(i,j)= \sum_{k\in\mathcal{I}_{ij}}1= |\mathcal{I}_{ij}|.$$ We define $\mathcal{P}_{\Omega}(\cdot)$ to be the projection onto $\Omega \subset \mathbb{R}^{d\times d}$ under Frobenius norm. Let $\mathbf{1}$ be the all-ones matrix with self-evident size, and $\oslash$ represents the element-wise division between two matrices of the same size. Furthermore, we introduce the soft thresholding operator $\mathcal{S}_{\beta}(\cdot)$ for a matrix $\bm{A}$ that admits singular value decomposition $\bm{A}= \bm{U\Sigma V^*}$, where the singular values of $\bm{A}$ are arranged in the diagonal matrix $\bm{\Sigma}$. Based on $\mathcal{S}_\beta(x)=\sign(x)\max\{0,|x|- \beta\}$ for $x\in\mathbb{R}$, we define $\mathcal{S}_{\beta}(\bm{A})= \bm{U}\mathcal{S}_{\beta}(\bm{\Sigma})\bm{V^*}$ and let $\mathcal{S}_{\beta}(\cdot)$ element-wisely operates on the diagonal matrix $\bm{\Sigma}$. Now, one can derive the ADMM iteration formulas as
\begin{equation}
\begin{cases}
   \bm{M_{t+1}}=\mathcal{P}_{\|\bm{M}\|_{\max}\leq \alpha^*}[(n\rho\cdot\bm{Z_t}+\bm{J_1}-n\bm{\Upsilon_t})\oslash(n\rho\cdot\mathbf{1}+\bm{J_2})]\\
   \bm{Z_{t+1}} = \mathcal{S}_{\lambda /\rho}(\rho^{-1}\cdot\bm{\Upsilon_t}+\bm{M_{t+1}})\\
   \bm{\Upsilon_{t+1}}=\bm{\Upsilon_{t}}+\rho\cdot(\bm{M_{t+1}}-\bm{Z_{t+1}})
\end{cases}.
    \label{711_5}
\end{equation}
\end{document}